\documentclass[conference]{IEEEtran}

\ifCLASSINFOpdf
\else
\fi

\usepackage{tikz}
\usepackage{multirow}
\usepackage{amsmath}
\usepackage{amssymb}
\usepackage{placeins}
\usepackage{nicefrac}
\usepackage{xspace}
\usepackage{caption}
\usepackage{numprint}
\usepackage{subcaption}
\usepackage{graphicx}
\usepackage{algpseudocode}
\usepackage{amsmath}
\usepackage{algorithm}
\usepackage{graphicx}
\usepackage{pgfplots}
\pgfplotsset{compat=1.17}
\usepackage{url}
\usepackage{subcaption}
\usepackage{colortbl}
\usepackage{rotating}
\usepackage{pgfplots}
\usepackage{xcolor}
\usepackage{color,soul}
\usepackage{booktabs}
\usepackage{multirow}
\usepackage{siunitx} 
\usepackage[table]{xcolor} 

\usepackage{adjustbox}
\usepackage{wasysym}
\newcommand{\ournameNoSpace}{\mbox{LightSplit}}
\newcommand{\ourname}{\ournameNoSpace\xspace}
\newcommand{\ournameGen}{\ournameNoSpace's\xspace}
\newcommand{\paperTitle}{\ourname: Practical Privacy-Preserving Split Learning via Orthogonal Projections}

\newcommand{\sect}{Sect.~}

\newcommand{\etal}{\emph{et~al.}\xspace}

\newcommand{\ournameFixed}{\textsc{LightSplit-F}\xspace}
\newcommand{\ournameLearned}{\textsc{LightSplit-L}\xspace}

\newcommand{\scaleTable}[1]{\scalebox{0.84}{#1}}

\definecolor{bestgreen}{HTML}{E8F5E9}
\definecolor{rawgray}{HTML}{F5F5F5}
\definecolor{lsheat}{HTML}{1F5C8B}

\definecolor{DarkGreen}{RGB}{10,100,32}

\newcommand{\CR}{\mathrm{CR}}

\urldef{\coderepo}\url{https://github.com/Mert-Cihangiroglu/LightSplit}

\newcommand{\topksparse}{top-$K$ sparse\xspace}

\begin{document}
\title{\paperTitle\ (Full Version)$^{\dagger}$\thanks{$^{\dagger}$Please cite the version of this paper published at the Network and Distributed System Security (NDSS) Symposium 2027~\cite{cihangiroglu2027lightsplitndss}.}}

\author{
\IEEEauthorblockN{Mert Cihangiroglu}
\IEEEauthorblockA{University of Pavia\\
mert.cihangiroglu01@universitadipavia.it}
\and
\IEEEauthorblockN{Alessandro Pegoraro}
\IEEEauthorblockA{Technical University of Darmstadt\\
alessandro.pegoraro@tu-darmstadt.de}
\and
\IEEEauthorblockN{Phillip Rieger$^*$}
\IEEEauthorblockA{Technical University of Graz\\
phillip.rieger@tugraz.at}
\and \hspace{3.6cm}
\IEEEauthorblockN{Antonino Nocera}
\IEEEauthorblockA{\hspace{3.6cm}University of Pavia\\
\hspace{3.6cm}antonino.nocera@unipv.it}
\and\hspace{.7cm}
\IEEEauthorblockN{Ahmad-Reza Sadeghi}
\IEEEauthorblockA{\hspace{.7cm}Technical University of Darmstadt\\
\hspace{.7cm}ahmad.sadeghi@trust.tu-darmstadt.de}
}

\IEEEoverridecommandlockouts
\makeatletter\def\@IEEEpubidpullup{6.5\baselineskip}\makeatother

\maketitle

\begin{abstract}
Split learning (SL) enables collaborative training by partitioning a neural network across clients and a central server. However, the cut-layer interface introduces a key challenge: high-dimensional activations incur substantial communication overhead while exposing representations vulnerable to reconstruction attacks. Existing approaches typically address efficiency or privacy in isolation, relying on additional mechanisms such as sparsification, quantization, or noise injection.

We propose \textsc{\ourname}, which limits information exposure and reduces communication overhead by applying a lightweight fixed orthogonal random projection at the cut layer. Based on Shannon’s information theory, this projection acts as an information bottleneck that restricts instance-specific information and suppresses exploitable per-sample signals. By transmitting low-dimensional projections instead of raw activations, the server operates on lifted representations without requiring architectural modifications, ensuring compatibility with existing SL architectures. By avoiding additional trainable components on the client, the method remains lightweight and suitable for edge devices while preserving end-to-end differentiability via exact gradient propagation. As the projection is non-invertible, part of the original representation is irreversibly discarded at the client, \ourname reduces the information available for reconstruction and limits information exposure.

We evaluate \ourname on seven benchmarks spanning image classification, tabular data, and language-model fine-tuning, at compression ratios up to $32\times$ and under IID and non-IID client data. At up to $32\times$ it retains at least $94\%$ of baseline accuracy on CIFAR-10, FashionMNIST, GTSRB, and UCI HAR, $86\%$ on CIFAR-100, and $91\%$ on average under non-IID client data; at $8\times$ it retains $92\%$ of ImageNet-1k top-1 and raises GPT-2 perplexity by at most $1.15\times$. Against state-of-the-art reconstruction attacks \ourname resists inversion; the strongest, co-trained attack loses up to $87.3\%$ of its reconstruction SSIM, and subject re-identification \mbox{on UCI HAR drops by more than half.}

\end{abstract}

\IEEEpeerreviewmaketitle

\section{Introduction}
\label{sec:intro}
Split learning (SL) enables computationally restricted devices to perform training and inference of deep neural networks (DNNs) without sharing their potentially sensitive data. Instead of outsourcing the entire process to a powerful server, the DNN is split into multiple fragments, the head and the backbone. The first layers of the DNN process sensitive input data, form the head, and are located and processed on the local device. The activations of the final layer of the head, which are known as smashed data, are then sent to the server, and the server completes the forward and backward passes through the remaining layers, which form the backbone~\cite{vepakomma2018split,gupta2018distributed}. Sharing the server located backbone enables clients to benefit from each others' data without sharing them. Unlike Federated Learning (FL) \cite{mcmahan2023communicationefficientlearningdeepnetworks}, in which the client is required to provide sufficient resources to hold a full model copy, SL is also suitable for edge scenarios where devices possess limited memory and computing resources.

\noindent\textbf{Communication Overhead and Privacy Attacks.} The convenience comes at two costs. First, the smashed data must be exchanged at every training step, creating a communication bottleneck that scales linearly with the activation dimensionality. For a standard ResNet-18~\cite{he2015deepresiduallearningimage}, this amounts to over $150$\,GB of network traffic for a training run on a modest CIFAR-10 with five clients (Table~\ref{tab:comm_cost}). Second, the smashed data is an information-rich intermediate representation that an honest-but-curious server can exploit. 
Indeed, many novel works have shown that SL is susceptible to multiple security threats, such as input reconstruction~\cite{ erdogan2022unsplit, xu2024stealthy, zhu2025passiveinferenceattackssplit}, label leakage~\cite{li2022labelleak}, and backdoor manipulation~\cite{yu2023backdoor}. In particular, server-based reconstruction attacks pose a critical risk (notably SDAR~\cite{zhu2025passiveinferenceattackssplit}), demonstrating that a passive server can recover client inputs from uncompressed, smashed data with high fidelity, undermining the privacy premise of SL. In this work, we focus on mitigating such threats and protecting client data under these adversarial conditions.

\def\thefootnote{*}\footnotetext{Work was partially done while being affiliated with TU Darmstadt.}
\textbf{}
\renewcommand*{\thefootnote}{\arabic{footnote}}

\noindent\textbf{Existing Literature.}
Prior work in the cut layer addresses privacy or communication efficiency in isolation.
Privacy-focused defenses~\cite{vepakomma2020nopeek,mireshghallah2020shredderlearningnoisedistributions,singh2021disco} perturb activations to resist reconstruction, but transmit them at full dimensionality and leave the communication bottleneck untouched.
Communication efficiency focused methods reduce payload via compression, but each comes with its own limitations: top-$k$ sparsification and its variants~\cite{randtopk2023,ZhouQLCY24} reveal which positions are most active per input, learned codecs~\cite{bottlenet2020,frankensplit2024} require trainable encoders and external pretraining on the client, and batch-wise compression~\cite{hsieh2022c3slcircularconvolutionbasedbatchwise} reduces transmissions per round but not the per-sample payload.
Cryptographic mechanisms~\cite{li2023fedvs,gilad2016cryptonets} provide formal confidentiality at orders-of-magnitude computational overhead, which is unsuitable for resource-constrained clients.
To our knowledge, no existing method simultaneously reduces the per-sample payload, avoids client-side overhead, and limits the information exposed to the server under a semi-honest threat model.

\noindent\textbf{Goals and Contributions.}
To address both privacy leakage and communication overhead in split learning, we introduce \textsc{\ourname}.\footnote{Our implementation is available at \coderepo.} The method projects cut-layer activations into a lower-dimensional subspace using a lightweight orthogonal transformation inspired by the Johnson-Lindenstrauss (JL) lemma~\cite{johnson1984extensions}. This projection reduces the amount of transmitted data while constraining instance-specific information that can be exploited by reconstruction attacks. To further limit residual leakage, the training process reduces intra-class variation in the transmitted representations, encouraging samples of the same class to form compact clusters. As a result, the representation behaves more like a class prototype than an instance-specific descriptor, making individual samples harder to distinguish. Together, these components allow \textsc{\ourname} to limit information exposure while maintaining model utility and to seamlessly integrate into existing SL architectures without requiring modifications to the underlying network. The fixed projection and the regularization jointly allow the client model to align its representations with the projected subspace, enabling co-adaptation to the reduced interface without increasing the complexity of the bottleneck. Our contributions include:

\begin{itemize}
\item  We introduce \textsc{\ourname}, a split learning framework that jointly addresses communication overhead and privacy leakage. By projecting smashed data into a fixed, randomly chosen orthonormal subspace, \ourname discards information in the orthogonal complement, limiting the information available for reconstruction attacks and thereby mitigating privacy leakage (\sect\ref{sec:method-hlo}).

\item
We design an irreversible information bottleneck through the projection that discards components orthogonal to the projected subspace, limits information accessible to a curious server and degrades reconstruction attacks by making the inverse mapping underdetermined (\sect\ref{sec:method-projection}).

\item
We design a Within-Class Compaction (WCC) regularizer that preserves model utility by encouraging the client model to align its representations with the surviving subspace. Thus, \ourname enables the client to retain task-relevant information and maintain competitive accuracy even under non-IID data distributions (\sect\ref{sec:method-wcc}).

\item
\textsc{\ourname} introduces negligible overhead by employing a fixed orthonormal projection on the client, requiring only a single matrix multiplication without additional learnable parameters. On the server side, it supports two lift-back modes allowing us to trade off flexibility and computational efficiency, enabling operation without any additional learnable parameters (\sect\ref{sec:method-liftback}).
\item
As a consequence of suppressing task-irrelevant variation while preserving class-discriminative features, the projections emphasize class-level structure. As a desirable side effect, we observe that this improves separability in the projected space for some scenarios and has the potential to benefit analysis tasks that seek to detect model poisoning attacks (\sect\ref{subsec:backdoor}). 

\item
We perform a comprehensive evaluation across multiple architectures and datasets, demonstrating that \textsc{\ourname} reduces reconstruction capability of state-of-the-art attacks up to $7$--$8\times$ (SDAR attacker's reconstruction SSIM falls from $0.83$ to $0.11$ at $16\times$, App.~Table~\mbox{\ref{tab:final-sdar-usl-all-split-points}}) and reduces communication overhead up to $32{\times}$ while retaining $86$--$100\%$ of the accuracy (\sect\ref{sec:eval}).

\end{itemize}

\begin{table}[t]
\centering
\small
\caption{Aggregate split-point communication cost over $100$ epochs for $N\!=\!5$ clients on CIFAR-10.}%
\label{tab:comm_cost}
\scaleTable{%
\begin{tabular}{rrrr}
\toprule
\textbf{Compression} $\rho$ & \textbf{Payload} $k$ & \textbf{Floats/sample} & \textbf{Total (GB)} \\
\midrule
(Raw)\hspace{0.25cm} $1\times$   & $4{,}096$ & $8{,}192$ & $152.6$ \\
$8\times$        & $512$     & $1{,}024$ & $19.1$  \\
$16\times$       & $256$     & $512$     & $9.5$   \\
$32\times$       & $128$     & $256$     & $4.8$   \\
\bottomrule
\end{tabular}%
}
\end{table}

\section{Background}
\label{sec:background}

\subsection{Split Learning}
\label{sec:bg:sl}

\begin{figure}[t]
  \centering
  \includegraphics[width=0.76\columnwidth]{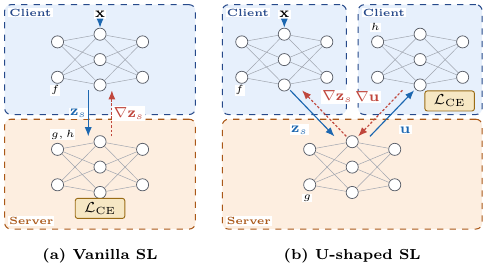}
  \caption{\textbf{Vanilla and U-shaped split-learning topologies.}}
  \label{fig:sl-topologies}
\end{figure}

In split learning (SL)~\cite{gupta2018distributed}, a DNN is partitioned at a designated \emph{cut layer} between a client holding privacy-sensitive data and a server with strong computational resources.
In the simplest configuration (vanilla), the client holds the first layers $f$, and the server holds the remainder $g$, including the classifier. Therefore, the server has access to both the intermediate activations and the training labels, as visualized in Fig.~\ref{fig:sl-topologies}a.
In the \emph{U-shaped} configuration~\cite{vepakomma2018split}, the model is split into three parts $H = h \circ g \circ f$: the client holds both the head $f$ (early layers) and the tail $h$, while the server holds only the intermediate backbone $g$, as visualized in Fig.~\ref{fig:sl-topologies}b.
We adopt U-shaped SL throughout this work, as it represents a more realistic and challenging setting: labels are private by design, and the server must reconstruct without access to them.

During each training step, the client feeds its private input $\mathbf{x}$ through $f$ to produce the intermediate activation tensor, termed \emph{smashed data}, and transmits it to the server.
The server runs the backbone $g$ and returns its output to the client, where the tail $h$ produces the final prediction and evaluates the task loss.
Gradients flow back through the same path: the client calculates the loss and backpropagates through $h$, passes the gradient to the server for backpropagation through $g$, and receives $\nabla_{\mathbf{z}}\mathcal{L}$ to update $f$.
Because both raw data and labels remain on the client, the server observes only the intermediate activations and the corresponding gradients, but never inputs, labels, or the final loss. In multi-client settings~\cite{thapa2022splitfed}, communication scales with both the cut-layer width $d$ and the number of participating clients~\cite{splitwireless2023,ecofed2024}: every round requires transmitting a $d$-dimensional activation \emph{and} a $d$-dimensional gradient per client per batch, as illustrated in Tab.~\ref{tab:comm_cost}.
Note that SL differs from federated learning~\cite{mcmahan2023communicationefficientlearningdeepnetworks} in that there is no local model copy or periodic aggregation. All clients contribute to the same shared backbone through sequential gradient updates. However, there is a collaborative version of SL, where clients exchange the weights and biases of the head and tail, enabling other clients to benefit from their training.

\subsection{Random Projections and Johnson-Lindenstrauss Lemma}
\label{sec:bg-rp}

The Johnson-Lindenstrauss (JL) lemma~\cite{johnson1984extensions} states that a random linear map $\mathbf{R} \in \mathbb{R}^{d \times k}$ with orthonormal columns preserves pairwise Euclidean distances among a finite point set up to a $(1\pm\varepsilon)$ factor with high probability, provided $k = \Omega(\varepsilon^{-2}\log n)$.
For any two activations $\mathbf{h}_i, \mathbf{h}_j \in \mathbb{R}^d$, the projected pair satisfies:
\begin{equation}
    (1{-}\varepsilon)\|\mathbf{h}_i - \mathbf{h}_j\|^2 \leq \|\mathbf{R}^\top\mathbf{h}_i - \mathbf{R}^\top\mathbf{h}_j\|^2 \leq (1{+}\varepsilon)\|\mathbf{h}_i - \mathbf{h}_j\|^2.
\end{equation}
Applied to cut-layer activations, this implies that the geometric structure underlying class separability, inter-class distances and intra-class clustering, is preserved under $\mathbf{R}$ without any task-adaptive optimization.

Dimensionality reduction does not automatically guarantee privacy.
Blocki et al.~\cite{blocki2012jlprivacy} and Kenthapadi et al.~\cite{kenthapadi2013privacyjl} show that formal privacy guarantees for JL-style transforms require added noise and single-shot release, conditions that do not hold under repeated or auxiliary-informed observation, precisely the conditions of iterative SL training.
Bora et al.~\cite{bora2017compressed} prove that signals near the range of a generative model can be recovered from as few as $O(k_{\text{gen}} \log d)$ measurements, far fewer than the ambient dimension.
As we will elaborate in \sect\ref{sec:method-wcc}, \ourname builds on the JL-lemma but complements it with a within-class regularizer to guide training towards class prototypes and to remove instance descriptions that may leak information.
The absence of formal privacy guarantees of the JL-lemma motivates our decision to evaluate privacy \emph{empirically} under adaptive attackers rather than relying on dimensionality reduction as an implicit guarantee.

\section{Problem Setting}
\label{sec:problem}
\subsection{System Model}
\label{sec:problem-system}

We study the U-shaped SL (see \sect\ref{sec:bg:sl}) where $n$ clients share a server-side backbone $g$, each client holds head $f$ and tail $h$, and the cut-layer activation $\mathbf{z} = f(\mathbf{x}) \in \mathbb{R}^d$ is the sole object transmitted\footnote{While we use U-Shaped SL as running example, we show in App.~\ref{app:sdar} that the proposed algorithm can also operate in vanilla-SL configurations}.
Every training step requires sending a $d$-dimensional activation \emph{and} receiving a $d$-dimensional gradient per sample, so aggregate bandwidth scales linearly with $d$ and the number of clients.
Our goal is to replace the transmitted $\mathbf{z}$ with a $k$-dimensional summary ($k \ll d$) that preserves task utility while restructuring what the server observes.

\subsection{Threat Model}
\label{sec:problem-threat}

In the following, we consider an adversary that aims to infer information about the clients' samples, i.e., reconstruct them.

We adopt the \emph{honest-but-curious} (semi-honest) threat model: the server faithfully executes the training protocol but may attempt to infer clients' private data from the information it legitimately receives.
During training, the server observes the compressed activations $\tilde{\mathbf{z}}_i \in \mathbb{R}^k$ (not the original $\mathbf{z}_i \in \mathbb{R}^d$), the gradients $\nabla \tilde{\mathbf{z}}_i$ returned to clients, and its own backbone parameters $\theta_g$, which evolve throughout training.

\subsection{Requirements and Challenges}
\label{sec:problem-requirements}
An approach that effectively addresses and is practical needs to fulfill the following requirements:

\noindent\textbf{R1 - Attack Prevention:}
The method must prevent or significantly deter reconstruction and inference attacks of curious servers, under the assumed threat model.

\noindent\textbf{R2 - Model Utility Preservation:} In order to be practical, the approach must maintain model performance comparable to the uncompressed baseline, avoiding degradation in task accuracy or convergence stability.

\noindent\textbf{R3 - Communication Efficiency:}
The method must reduce the dimensionality of transmitted representations, thereby lowering communication overhead at the cut layer, which dominates the overall training cost.

Projecting cut-layer activations and addressing privacy leakage while maintaining utility fundamentally differs from standard model compression or gradient quantization. Therefore, methods that meet these requirements must overcome several research challenges.

\noindent\textbf{C1 - Shared backbone constrains unilateral compression:} 
A first challenge arises from the shared server-side backbone $g$, which prevents clients from independently applying arbitrary compression to their smashed data. Since $g$ is trained on the received representations, uncoordinated compression induces distribution shifts, producing out-of-distribution inputs that destabilize the backbone across clients. Thus, compression cannot be a purely local design choice but must follow a globally consistent protocol. A challenge is therefore how to ensure compatibility across clients and preserve a shared representation space while maintaining compression efficiency, a constraint that becomes even more critical in SL variants with individual client heads.

\noindent\textbf{C2 - Compression and privacy are entangled:}
A second challenge arises from the inherent tension between compression and privacy. Na\"{i}ve compression approaches aim to preserve as much information as possible, which directly conflicts with the goal of limiting information leakage. In particular, retaining the most informative components often preserves exactly the signal exploited by reconstruction attacks. For instance, Top-$K$ sparsification~\cite{randtopk2023,ZhouQLCY24} reveals \emph{which} neurons fire most for each input, effectively exposing a per-sample feature saliency map. Similarly, learned codecs~\cite{bottlenet2020,frankensplit2024} are explicitly optimized for high-fidelity reconstruction of activations, which directly opposes the objective of limiting what the server can recover about the input. A challenge is therefore how to design compression mechanisms that retain task-relevant structure, specifically inter-class geometry, while discarding the per-sample variation that enables reconstruction.

\noindent\textbf{C3 - Client overhead undermines the SL premise:}
A third challenge arises from the resource constraints of clients in SL. Any compression mechanism that introduces additional trainable components increases the very cost that SL is designed to minimize. At the same time, fixed, non-trainable compression must produce task-compatible representations without adapting to the data distribution. A challenge is therefore, how to design lightweight compression mechanisms that preserve utility without increasing client-side complexity.

\section{\ourname}
\label{sec:approach}
To improve privacy and communication efficiency in SL and address the aforementioned challenges, we introduce \ourname, which projects smashed data into a lower-dimensional subspace to reduce communication and limit information exposure.

\subsection{Design Rationale}
\label{sec:method-hlo}
\begin{figure}[t]
  \centering
  \includegraphics[width=0.85\columnwidth,trim={0cm 9.65cm 17cm 0},clip]{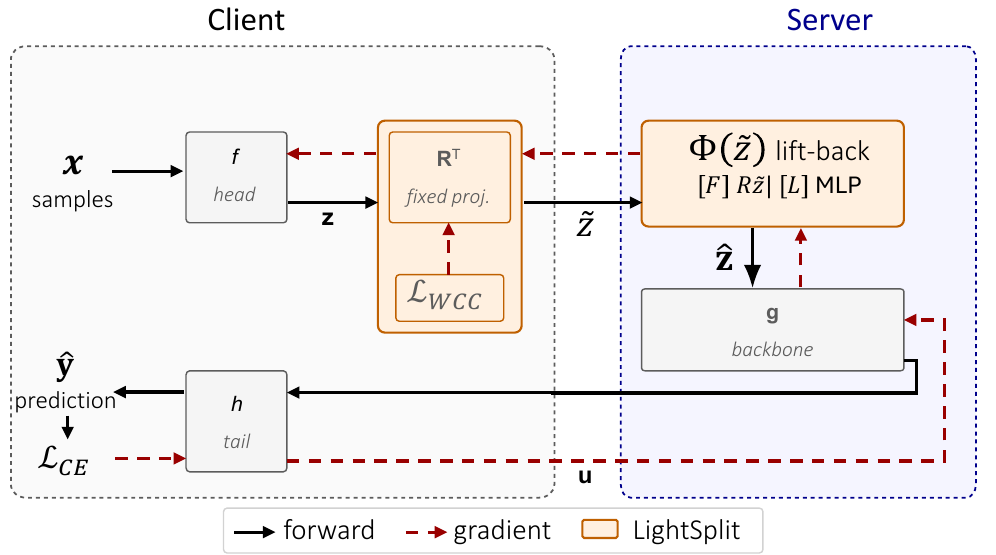}
\caption{Overview of \ourname for U-shaped SL setup with client-located head $f$ and tail $h$, and server-side backbone $g$.
\ourname components are highlighted by orange boxes. Inputs $\mathbf{x}$ are mapped to activation $\mathbf{z}$, which are projected to representations $\tilde{\mathbf{z}}\in\mathbb{R}^k$ before transmission. The server reconstructs activations as $\hat{\mathbf{z}}$ via lift-back function $\phi$ and processes by $g$, obtaining smashed data $\mathbf{u}$. Client computes the prediction $\hat{y}$ and task loss $\mathcal{L}_{\mathrm{CE}}$ and WCC loss $\mathcal{L}_{\mathrm{WCC}}$.
}

  \label{fig:lightsplit-overview}
\end{figure}

To reduce information exposure and mitigate reconstruction attacks while improving communication efficiency, \ourname projects smashed data at the head-backbone cut into a lower-dimensional subspace and lets the server operate on reconstructed, back-lifted representations (Fig.~\mbox{\ref{fig:lightsplit-overview}}). Privacy attacks typically exploit subtle, instance-specific variations in the transmitted activations. The core challenge is therefore to preserve task-relevant, discriminative features while suppressing information that enables reconstruction.

\ourname employs a projection that builds on the Johnson-Lindenstrauss (JL) lemma (see \sect\ref{sec:bg-rp}), which guarantees that pairwise distances are approximately preserved under random projection. This ensures that class structure and separability remain intact despite dimensionality reduction. Concretely, each client applies a fixed orthonormal projection matrix $\mathbf{R} \in \mathbb{R}^{d \times k}$ to map smashed activations $\mathbf{z} \in \mathbb{R}^d$ (cut-layer dimension $d$) to projected representations $\tilde{\mathbf{z}} = \mathbf{R}^\top \mathbf{z} \in \mathbb{R}^k$ (projected dimension $k$, with $k \ll d$).

On the server side, we introduce two variants for processing the projected activations: \ournameFixed and \ournameLearned. \ournameFixed uses a fixed linear lift-back mapping, while \ournameLearned employs a trainable network to reconstruct task-compatible representations. The fixed variant is parameter-free and computationally lightweight, whereas the learned variant increases flexibility and can better adapt to the projected representation.

After server-side processing, the client receives the intermediate output, completes the forward pass via the tail, and computes the task loss $\mathcal{L}_{\mathrm{CE}}$. In addition, we introduce a Within-Class Compaction (WCC) loss that reduces intra-class variation in the transmitted representations. This discourages instance-specific feature scattering that can be exploited by reconstruction attacks, while preserving class-level structure necessary for accurate prediction.

Importantly, as \ourname focuses on the head-backbone cut, making it independent of the specific SL configuration, allowing its application also to vanilla-SL configurations.

\subsection{Client-Side Projection}
\label{sec:method-projection}

A client-side projection in SL must reduce the cut-layer payload while addressing the resource constraints of devices that restrict introducing additional trainable components in some scenarios. At the same time, it should limit the amount of per-sample information exposed to the server. \ourname solves these challenges through the design of a fixed random projection at the cut layer.%

Each client projects its smashed activation $\mathbf{z} \in \mathbb{R}^d$ into a lower-dimensional representation $\tilde{\mathbf{z}} \in \mathbb{R}^k$ using a fixed orthonormal matrix $\mathbf{R} \in \mathbb{R}^{d \times k}$:
\begin{equation}
    \mathbf{R}^{\!\top}\!\colon \mathbb{R}^{d}\!\longrightarrow\!\mathbb{R}^{k},
    \qquad
    \mathbf{z} \;\longmapsto\; \tilde{\mathbf{z}} \;=\; \mathbf{R}^{\!\top}\mathbf{z},
    \qquad k \ll d.
    \label{eq:projection}
\end{equation}

Equation~\eqref{eq:projection} applies $\mathbf{R}$ to the flattened activation. When that activation is an $H{\times}W$ feature map, for larger smashed-data dimensions, the client can also cut the grid into tiles of $b{\times}b$ positions and apply the same fixed $\mathbf{R}_{b}$ to each, splitting the
$k$-scalar budget across tiles in a fixed spatial order. A single tile recovers
Eq.~\eqref{eq:projection}.\label{sec:method-block} 

The matrix $\mathbf{R}$ is broadcast in the beginning to all clients and remains fixed throughout training, and introduces no additional learnable parameters.
$\mathbf{R}$ is built by sampling a $d\!\times\!k$ matrix $\mathbf{A}$ of i.i.d.\ standard Gaussian entries and computing its thin QR decomposition,
\begin{equation}
    \mathbf{A} \;=\; \mathbf{Q}\,\mathbf{T},
    \label{eq:qr}
\end{equation}
where $\mathbf{Q}\!\in\!\mathbb{R}^{d\times k}$ has orthonormal columns ($\mathbf{Q}^{\top}\mathbf{Q}\!=\!\mathbf{I}_{k}$) and $\mathbf{T}\!\in\!\mathbb{R}^{k\times k}$ is upper triangular. We set $\mathbf{R}\!:=\!\mathbf{Q}$ and discard $\mathbf{T}$. Sampled once before training and fixed thereafter, this yields $\mathbf{R}^{\top}\mathbf{R}=\mathbf{I}_{k}$ exactly, is independent of any data, and adds no learnable parameters on the client. This is the standard random-projection setting under which the JL guarantee from \sect\ref{sec:bg-rp} applies.

The projection reduces the communication cost from $d$ to $k$ scalars per sample, with compression ratio $CR \;=\; d/k \;\geq\; 1$. Only the projected representation $\tilde{\mathbf{z}}$ is transmitted to the server, and gradients are returned with respect to $\tilde{\mathbf{z}}$, so that the original activation $\mathbf{z}$ never leaves the client.

From a privacy perspective, the projection acts as an information bottleneck. Let $I(\cdot;\cdot)$ denote the Shannon mutual information between two random variables, intuitively, the number of bits of information one carries about the other. Since $\mathbf{x}\!\to\!\mathbf{z}\!\to\!\tilde{\mathbf{z}}$ forms a Markov chain, the data-processing inequality implies  $I(\mathbf{x};\tilde{\mathbf{z}})\leq I(\mathbf{x};\mathbf{z})$. Thus, the projection cannot introduce new information about $\mathbf{x}$ that were not already present in $\mathbf{z}$. 

However, it does not eliminate information leakage, as residual instance-specific signals may still remain in $\tilde{\mathbf{z}}$. The projection reduces the overall information content, while the complementary Within-Class Compaction (see \sect\ref{sec:method-wcc}) mechanism further suppresses sample-specific variation that can be exploited by reconstruction attacks.

From a utility perspective, the projection preserves the geometric structure of the activations. As given by the JL-lemma (see \sect\ref{sec:bg-rp}), pairwise distances are approximately maintained under random projection with high probability. This is sufficient for downstream classification, where class separability primarily depends on relative distances.

\subsection{Server-Side Liftback}
\label{sec:method-liftback}

A key design objective for \ournameGen server-side components is seamless integration into existing SL architectures without requiring modifications to the underlying DNN. While the client-side design reduces information exposure and projects activations into a lower-dimensional subspace, the server-side preserves compatibility with the backbone.

Therefore, the projection $\tilde{\mathbf{z}} \in \mathbb{R}^k$ must be mapped to a $d$-dimensional input, compatible with the backbone $g$. This mapping should preserve task-relevant structure while compensating for the distribution shift induced by the projection.

We define two lift-back mechanisms (\ournameFixed and \ournameLearned) as trade-off between efficiency and adaptability. \ournameFixed uses a parameter-free linear mapping, minimizing computational overhead, while \ournameLearned employs a trainable transformation to better adapt to the projected representation and recover task-relevant structure. Both modes use the same client projection and communication budget but differ only in server-side parameterization and how the server processes incoming projected representations.

\paragraph{\ournameFixed - fixed linear lift-back}
\ournameFixed is the low-parameter server mode that directly reuses the projection matrix $\mathbf{R}$ to lift the projected representation back to the original dimensionality as:
\begin{equation}
    \hat{\mathbf{z}} \;=\; \mathbf{R}\tilde{\mathbf{z}}
    \;=\; \mathbf{R}\mathbf{R}^{\top}\mathbf{z}
    \;\in\; \mathbb{R}^{d}.
    \label{eq:fr_liftback}
\end{equation}
Let $\mathbf{P}=\mathbf{R}\mathbf{R}^{\top}\in\mathbb{R}^{d\times d}$. Since $\mathbf{R}^{\top}\mathbf{R}=\mathbf{I}_{k}$, the matrix $\mathbf{P}$ satisfies $\mathbf{P}^{2}=\mathbf{P}$ and $\mathbf{P}^{\top}=\mathbf{P}$, so $\mathbf{P}$ is the orthogonal projector onto the $k$-dimensional subspace $\mathrm{range}(\mathbf{R})\subset\mathbb{R}^{d}$, and $\hat{\mathbf{z}}=\mathbf{P}\mathbf{z}$ is the best Euclidean approximation to $\mathbf{z}$ within that subspace, with residual
\begin{equation}
    \|\mathbf{z}-\hat{\mathbf{z}}\|_2^2
    \;=\;
    \|(\mathbf{I}_{d}-\mathbf{R}\mathbf{R}^{\top})\,\mathbf{z}\|_2^2.
\end{equation}
The residual lies entirely in the $(d{-}k)$-dimensional null space of $\mathbf{P}$, so the lift-back is lossy by construction. The chain rule gives the gradient that the server returns to the client through the linear bottleneck: for any scalar loss $\mathcal{L}$ that depends on $\mathbf{z}$ only through $\tilde{\mathbf{z}}=\mathbf{R}^{\top}\mathbf{z}$,
\begin{equation}
    \nabla_{\mathbf{z}}\mathcal{L}
    \;=\;
    \mathbf{R}\,\nabla_{\tilde{\mathbf{z}}}\mathcal{L},
    \label{eq:fr_grad}
\end{equation}
so the same fixed matrix $\mathbf{R}$ that projects the forward activation also lifts the returned gradient back to the client's coordinate system. The lift-back itself adds no trainable server parameters and reduces server-side computation to a single fixed matrix multiplication defined by $\mathbf{R}$.

\paragraph{\ournameLearned - learned MLP lift-back}

\begin{figure}[t]
    \centering
    \includegraphics[width=0.55\columnwidth,trim={0.25cm 0.75cm 0.5cm 0.25cm},clip]{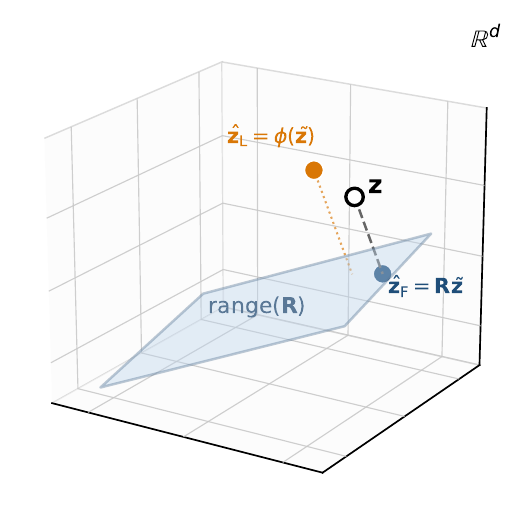}
\caption{Relation between the cut-layer activation ( $\mathbf{z}\in\mathbb{R}^{d}$), its projected representation $\tilde{\mathbf{z}}=\mathbf{R}^{\top}\mathbf{z}\in\mathbb{R}^{k}$ into the $k$-dimensional subspace $\mathrm{range}(\mathbf{R})$ (shaded plane), and the lift-back mappings $\hat{\mathbf{z}}_{\mathrm{F}}=\mathbf{R}\tilde{\mathbf{z}}$.}
\label{fig:liftback-geometry}

\end{figure}

\ournameLearned replaces the fixed map of Eq.~\eqref{eq:fr_liftback} with a learnable lift-back $\phi:\mathbb{R}^{k}\!\to\!\mathbb{R}^{d}$ trained jointly with $g$. Fig.~\ref{fig:liftback-geometry} visualizes the geometry behind the two modes. The hollow circle is the original cut-layer activation $\mathbf{z}\in\mathbb{R}^{d}$, the blue plane is $\mathrm{range}(\mathbf{R})$, the $k$-dimensional task-independent subspace fixed before training. The dashed segment is the residual $\mathbf{z}-\mathbf{P}\mathbf{z}$ that the projection discards (with $\mathbf{P}=\mathbf{R}\mathbf{R}^{\top}$). It lies in the null space of $\mathbf{P}$ and is invisible to the server in either mode. \ournameFixed returns $\hat{\mathbf{z}}_{\mathrm{F}}=\mathbf{R}\tilde{\mathbf{z}}=\mathbf{P}\mathbf{z}$ (blue dot): the orthogonal projection of $\mathbf{z}$ onto the plane, pinned inside $\mathrm{range}(\mathbf{R})$. \ournameLearned returns $\hat{\mathbf{z}}_{\mathrm{L}}=\phi(\tilde{\mathbf{z}})$ (orange dot), which sits off the plane; the dotted drop indicates that $\hat{\mathbf{z}}_{\mathrm{L}}$ is no longer constrained to $\mathrm{range}(\mathbf{R})$. Both modes operate on the same $\tilde{\mathbf{z}}$, receiving identical information about $\mathbf{z}$. Due to the data-processing inequality, $\phi$ cannot recover the discarded residual~\cite{cover2006elements}.  $\phi$ does not recover the residual, the projection discards. It only re-encodes the surviving information across $\mathbb{R}^{d}$ instead of constraining it to $\mathrm{range}(\mathbf{R})$.

\textit{Motivation.} Under \ournameFixed, the backbone $g$ receives input pinned to the same $k$-dimensional plane of $\mathbb{R}^{d}$ for every sample, fixed before training. To extract task features from that plane $g$ must spend representational capacity on a global, non-spatial linear re-mapping. This conflicts with the spatial inductive biases of the convolutional backbones. \ournameLearned places that re-mapping in an MLP at the $\mathbb{R}^{k}\!\to\!\mathbb{R}^{d}$ interface absorbs the algebra cheaply and frees $g$'s capacity for task-relevant features. The same adapter is also the only adaptable component in the server pipeline when $g$ is frozen or pre-trained. Concretely, the server trains an MLP with hidden width $m$ jointly with the backbone:
\begin{equation}
    \hat{\mathbf{z}}
    \;=\;
    \mathbf{W}_2\,\sigma\!\left(\mathrm{BN}(\mathbf{W}_1\tilde{\mathbf{z}})\right)
    \;\in\;\mathbb{R}^{d},
    \label{eq:frll_liftback}
\end{equation}
with $\mathbf{W}_1\!\in\!\mathbb{R}^{m\times k}$, $\mathbf{W}_2\!\in\!\mathbb{R}^{d\times m}$, $\mathrm{BN}(\cdot)$ batch normalization, and $\sigma(\cdot)$ element-wise ReLU.\\

The choice between \ournameFixed and \ournameLearned is a deployment decision for the server: the same payload $\tilde{\mathbf{z}}$ is sent in both modes, so client cost, bandwidth, and adversary view are identical, and the privacy properties (\sect\ref{sec:method-wcc}) are inherited unchanged. \ournameFixed is the minimal scheme, without learned lift-back, server-side parameters, or extra training dynamics. Thus, it is preferable when the backbone has spare capacity to absorb the rank-$k$ restriction. \ournameLearned is suitable when $g$ is frozen or the inductive-bias mismatch becomes the limiting factor. As shown in  \sect\ref{sec:eval}, \ournameLearned achieves better performance \ournameFixed on more complex benchmarks while providing similar results in remaining benchmarks.

The following subsection discusses the remaining privacy challenge of projection-based approaches: while projection reduces the overall information content, residual instance-level signals may persist in the transmitted representation. We therefore introduce Within-Class Compaction (WCC), which suppresses such variation and limits the information exploitable by reconstruction attacks.

\subsection{Privacy Role of Within-Class Compaction}
\label{sec:method-wcc}

Notably, projection alone provides only limited privacy protection. While the JL property preserves class structure, it may also retain cues that can be exploited by feature-inversion and model-inversion attacks. In particular, a malicious server with access to auxiliary data can train a simulator or decoder on the observed representations. Since the projection matrix $\mathbf{R}$ is known, the server can lift $\tilde{\mathbf{z}}$ back into the projected activation subspace, enabling it to exploit residual structure for reconstructing information about the clients' data. Thus, while random projection reduces dimensionality, it does not eliminate all instance-specific variation in the transmitted representation that potentially allows information leakage.

To limit this leakage, we introduce Within-Class Compaction (WCC), which targets the residual instance-level variation that remains after projection. The key idea is to preserve class-discriminative structure while suppressing variability that is not required for the learning task.

This intuition can be formalized using Shannon mutual information. Treating the label $y$ as a deterministic function of the input $\mathbf{x}$ (as is standard in supervised learning), the chain rule of mutual information yields

\begin{equation}
    I(\mathbf{x};\tilde{\mathbf{z}})
    \;=\;
    I(y;\tilde{\mathbf{z}})
    \;+\;
    I(\mathbf{x};\tilde{\mathbf{z}}\mid y).
    \label{eq:mi_decomp}
\end{equation}

$I(y;\tilde{\mathbf{z}})$, captures the class-discriminative signal required for prediction and should be preserved. The second term, $I(\mathbf{x};\tilde{\mathbf{z}}\mid y)$, represents residual instance-specific information given the label. This term upper-bounds the information available to reconstruction attacks, and reducing it increases the inherent difficulty of recovering individual inputs.

WCC targets this residual term through a tractable surrogate. By minimizing within-class scatter in the transmitted representation, it encourages samples of the same class to form compact clusters, making $\tilde{\mathbf{z}}$ behave more like a class prototype than an instance descriptor. As a result, inputs from the same class become less distinguishable, reducing the exploitable per-sample signal while preserving class-level structure.

For a mini-batch $\{(\tilde{\mathbf{z}}_i,y_i)\}_{i=1}^{b}$ of size $b$, where $y_i\in\{1,\dots,C\}$ is the class label of sample $i$ and $C$ is the number of classes, let $S_c=\{i:y_i=c\}$ denote the index set of samples in class $c$, and define the class centroid $\boldsymbol{\mu}_c$ in the projected space as:

\begin{equation}
    \boldsymbol{\mu}_c
    \;=\;
    \frac{1}{|S_c|}
    \sum_{i\in S_c}\tilde{\mathbf{z}}_i,
    \label{eq:wcc_centroid}
\end{equation}
The WCC loss  $\mathcal{L}_{\mathrm{WCC}}$ is then defined as:
\begin{equation}
    \mathcal{L}_{\mathrm{WCC}}
    \;=\;
    \sum_{c\in\mathcal{Y}_b}
    \frac{1}{|S_c|}
    \sum_{i\in S_c}
    \|\tilde{\mathbf{z}}_i-\boldsymbol{\mu}_c\|_2^2,
    \label{eq:wcc}
\end{equation}

where $\mathcal{Y}_b$ is the set of batch labels and $\|\cdot\|_2$ denotes the Euclidean norm. The loss is computed on the client, using labels that never leave the client, and adds no communication (see Appendix~\ref{app:wcc_backprop}).

The training objective $\mathcal{L}$ is defined as the combination:
\begin{equation}
    \mathcal{L}
    =
    \mathcal{L}_{\mathrm{CE}}(\hat{y},y)
    +
    \lambda_{\mathrm{WCC}}\mathcal{L}_{\mathrm{WCC}},
    \label{eq:total_loss}
\end{equation}
where $\lambda_{\mathrm{WCC}}$ controls the compaction strength. Setting $\lambda_{\mathrm{WCC}}=0$ recovers projection-only training. In our evaluation we ablate $\lambda_{\mathrm{WCC}}\in\{0,10^{-3},10^{-2},10^{-1}\}$ and find that a small weight already suffices: $\lambda_{\mathrm{WCC}}{\approx}10^{-3}$ captures the privacy gain at negligible utility cost, so we recommend it as the operating point; any $\lambda_{\mathrm{WCC}}{\ge}10^{-3}$ holds the privacy floor at a small additional utility cost.

Notably, WCC is not an adversarial loss nor trains a privacy discriminator, but a variance penalty that acts as a tractable surrogate for reducing the class-conditional per-sample signal $I(\mathbf{x};\tilde{\mathbf{z}}\mid y)$. It targets the same Shannon mutual information used in \sect\ref{sec:method-projection}, but now conditioned on the label $y$.

The full training procedure,  including where the projection, the lift-back, and $\mathcal{L}_{\mathrm{WCC}}$ are placed in one mini-batch step, is formalized in Alg.~\ref{alg:lightsplit-client} and Alg.~\ref{alg:lightsplit-server} for the client and server respectively. App.~\ref{app:wcc_backprop} expands this into a step-by-step forward and backward derivation under the two-optimizer deployment that mirrors a real client-server split.

\begin{algorithm}[t]
\caption{\ourname server-side training}
\label{alg:lightsplit-server}
\scalebox{0.72}{
\begin{minipage}{1.333\columnwidth}
\begin{algorithmic}[1]
\State \textbf{Input:} clients $\mathcal{C}$, rounds $T$, backbone $g$, projection dimension $k$, activation dimension $d$, learning rate $\eta_g$, mode $m\in\{\ournameFixed,\ournameLearned\}$
\State \textbf{Output:} updated backbone $g$ and lift-back $\phi$

\Statex \Comment{Initialize Projection and Lift-back}
\State $\mathbf{R} \gets \mathrm{QR}(\mathcal{N}(0,1)^{d\times k})$
\If{$m=\ournameFixed$} 
    \State $\phi \gets (\tilde{\mathbf{z}} \mapsto \mathbf{R}\tilde{\mathbf{z}})$
\Else
    \State $\phi \gets \mathrm{MLP}(k,d)$
\EndIf
\State $\texttt{Broadcast}(\mathbf{R}, \mathcal{C})$

\For{each training step $t\in[1,T\cdot|\mathcal{C}|]$}
    \State $C_j \gets \mathcal{C}[t \bmod |\mathcal{C}|]$

    \Statex \Comment{Receive and Lift-back Projections}
    \State $\tilde{\mathbf{z}} \gets \texttt{Receive}(C_j)$
    \State $\hat{\mathbf{z}} \gets \phi(\tilde{\mathbf{z}})$ 

    \Statex \Comment{Server Forward Pass}
    \State $\mathbf{u} \gets g(\hat{\mathbf{z}})$ 
    \State $\texttt{Send}(\mathbf{u}, C_j)$

    \Statex \Comment{Backpropagation and Model Update}
    \State $\mathbf{g}^{u} \gets \texttt{Receive}(C_j)$
    \State $\mathbf{g}^{\mathrm{cut}}, \nabla_g, \nabla_\phi \gets \texttt{Backpropagate}(\mathbf{g}^{u}, g,\phi)$
    \State $\texttt{Send}(\mathbf{g}^{\mathrm{cut}}, C_j)$
    \State $g \gets g-\eta_g\nabla_g$ 
    \If{$m=\ournameLearned$}
        \State $\phi \gets \phi-\eta_g\nabla_\phi$
    \EndIf
\EndFor

\State \Return $g$
\end{algorithmic}
\end{minipage}}
\end{algorithm}

\begin{algorithm}[t]
\caption{\ourname client-side training}
\label{alg:lightsplit-client}
\scalebox{0.72}{
\begin{minipage}{1.333\columnwidth}
\begin{algorithmic}[1]
\State \textbf{Input:} local dataset $\mathcal{D}_j$, head $f$, tail $h$, batch size $b$, WCC weight $\lambda_{\mathrm{WCC}}$, learning rate $\eta_c$
\State \textbf{Output:} updated head $f$ and tail $h$

\Statex \Comment{Receive Projection}
\State $\mathbf{R} \gets \texttt{Receive}(\mathrm{server})$

\For{each local training step}
    \Statex \Comment{Client Forward Pass and Projection}
    \State $(\mathbf{x},y) \gets \texttt{Batch}(\mathcal{D}_j,b)$
    \State $\mathbf{z} \gets f(\mathbf{x})$
    \State $\tilde{\mathbf{z}} \gets \mathbf{R}^{\top}\mathbf{z}$
    \State $\texttt{Send}(\tilde{\mathbf{z}}, \mathrm{server})$
    \State $\mathbf{u} \gets \texttt{Receive}(\mathrm{server})$

    \Statex \Comment{Client Tail and Loss Computation}
    \State $\hat{y} \gets h(\mathbf{u})$
    \State $\mathcal{L}_{\mathrm{CE}} \gets \ell_{\mathrm{CE}}(\hat{y},y)$
    \State $\mathcal{L}_{\mathrm{WCC}} \gets \texttt{WCC}(\tilde{\mathbf{z}},y)$
    \State $\mathcal{L} \gets \mathcal{L}_{\mathrm{CE}}+\lambda_{\mathrm{WCC}}\mathcal{L}_{\mathrm{WCC}}$

    \Statex \Comment{Backpropagation and Model Update}
    \State $\mathbf{g}^{u}, \nabla_h \gets \texttt{Backpropagate}(\mathcal{L}_{\mathrm{CE}}, h)$
    \State $\texttt{Send}(\mathbf{g}^{u}, \mathrm{server})$
    \State $\mathbf{g}^{\mathrm{cut}} \gets \texttt{Receive}(\mathrm{server})$
    \State $\mathbf{g}^{\tilde{\mathbf{z}}} \gets \mathbf{g}^{\mathrm{cut}}+\lambda_{\mathrm{WCC}}\nabla_{\tilde{\mathbf{z}}}\mathcal{L}_{\mathrm{WCC}}$
    \State $\nabla_f \gets \texttt{Backpropagate}(\mathbf{g}^{\tilde{\mathbf{z}}}, f,\mathbf{R}^{\top})$
    \State $h \gets h-\eta_c\nabla_h$
    \State $f \gets f-\eta_c\nabla_f$
\EndFor

\State \Return $f,h$
\end{algorithmic}
\end{minipage}}
\end{algorithm}

\section{Evaluation}
\label{sec:eval}
\subsection{Experimental Setup}
\label{sec:exp:setup}

We use the U-shaped SL protocol as the default (USL, Sec.~\ref{sec:bg:sl}), and we vary two orthogonal axes that recur across our experiments. \emph{(i) Client head size.} Each image model in App.~Table~\ref{tab:model_partitions} is split at two cut depths reported as L1H and L2H; the tabular and text models have a single cut. \emph{(ii) Client head ownership.} The client head is either per-client (PCH) (each client $i$ owns its own $f_i$) or shared SCH (a single $f$ is shared across clients). 

\subsubsection{Datasets, models, and cut layers}
For the baseline utility experiments measuring model utility and compression, we aligned with the work of Rieger~\etal~\cite{safesplit2025}. To measure utility, we use the CIFAR-10~\cite{Krizhevsky2009LearningML}, CIFAR-100~\cite{Krizhevsky2009LearningML}, and GTSRB~\cite{stallkamp2011german} datasets. We also use MNIST~\cite{726791} and Fashion-MNIST~\cite{xiao2017fashionmnist} datasets in the per-sample inversion attacks, where their public artifacts support it.We further test \ourname on ImageNet-1k~\cite{deng2009imagenet}, tabular activity recognition (UCI HAR~\cite{human_activity_recognition_using_smartphones_240}) and language modelling (GPT-2~\cite{radford2019language} fine-tuned on WikiText-103~\cite{merity2016pointer}) over the same $8$--$32\times$ range (App.~\ref{app:setup_imagenet}, \ref{app:setup_generality} and \ref{app:setup_llm}). We report the model partitions in App.~Table~\mbox{\ref{tab:model_partitions}}.

\textbf{Baselines: }We compare \ourname on three baselines that encompass the natural design choices at the cut layer:
\textbf{(1)~\textsc{Raw}} where we transmit the smashed activation unmodified and, without compression, set the upper bound on what the cut can leak.
\textbf{(2)~\textsc{Learned $1{\times}1$}} (L-1) where a lightweight BottleNet-style channel bottleneck trained end-to-end with the task loss~\cite{bottlenet2020} is used.
\textbf{(3)~top-$k$ sparse}  keeps the largest-magnitude activation coordinates at the same payload size. It is a communication-matched sparsification baseline that, unlike the projection, releases un-mixed activation values at a subset of coordinates.
We also evaluate differential privacy (DP) at the cut, on its own (\textsc{Raw}$+$DP) and composed with the projection (\ournameFixed$+$DP).
In App.~\mbox{\ref{app:setup_baselines}} we define each baseline and in App.~\mbox{\ref{app:setup_dp}} the DP mechanism.

\textbf{Compression ratios and their readings: }Compression is defined by $CR = d/k$, comparing the original activation dimension $d$ to the transmitted dimension $k$. For \textbf{\textsc{Raw}}, \textbf{\textsc{Learned $1{\times}1$}}, and \ourname, $CR$ jointly reflects bandwidth and information: fewer transmitted features imply less information available at the cut layer.\\

\subsubsection{Evaluated Attacks}
In the following, we describe the evaluated attacks and experimental setups. For each published attack, we adopt the original authors' setting and only insert \ourname after the first cutting layer between Client and the Server. We run the state-of-the-art passive reconstruction attacks; SDAR~\cite{zhu2025passiveinferenceattackssplit}, a substitute-encoder attack, UnSplit~\cite{erdogan2022unsplit}, a per-image gradient-matching inversion, and FORA~\cite{xu2024stealthy}, which trains an inversion decoder on the smashed activation, each with the configuration released by its authors; In Appendix~\mbox{\ref{app:setup_attacks}} we report the full configuration of every attack and in Appendix~\mbox{\ref{app:attacks}} its full result grids. Under these three attacks, we evaluate \ournameFixed: the attacker applies the lift-back to recover smashed activations and then runs the attack's original pipeline. The attack tables and figures therefore report \ournameFixed (LS-F). We additionally attack \ournameLearned (LS-L) with SDAR to test the learned lift-back directly.
Although our threat model assumes an honest-but-curious server (\sect\ref{sec:problem-threat}), we also stress-test \ourname with the active FSHA attack~{\cite{pasquini2021unleashing}}, where a malicious server steers the client encoder toward a feature space it can invert (See Appendix~\ref{app:fsha}).
Additionally, since the server applies the lift-back $zR^{\top}$ on every forward pass, $R$ is public; we therefore also train reconstruction decoders built around it, on $z$ alone, $z$ with $zR^{\top}$, $z$ with the returned gradient $g$, and all three fused (Appendix~\ref{app:projopt}).
On tabular data (UCI HAR) we run the analogous reconstruction~\cite{he2019model} and re-identification~\cite{song2020overlearning} probes (Appendix~\ref{app:tabular_attack}). These probes, like the decoders above, read the transmitted payload directly.
As a possible side effect, the suppression of instance-specific variation and emphasis on class-level structure might support the detection of poisoned activations. We test this with the BadNets trigger and label-flip setup~\cite{gu2017badnets} (App.~\mbox{\ref{app:backdoor_setup}}). In App.~Fig.~\ref{fig:backdoor_r} we report the per-client anomaly score under each ablation.

\textbf{Metrics: }For the utility baselines, we measure top-1 test accuracy on the task classifier. In Tables~\ref{tab:main-compact}, \ref{tab:privacy_all} and App.~Table~\ref{tab:client_scaling_ablation}), entries marked with $\Delta$ report the change from the matched \textsc{Raw} baseline ($\Delta_{Raw}$), either in percentage points or, where explicitly stated, as a relative percentage change. For the image reconstruction attacks we report MSE, PSNR, SSIM~\cite{1284395} and LPIPS~\cite{Zhang_2018_CVPR}, with arrows pointing toward stronger privacy; In App.~\ref{app:metrics} we define each metric and the foreground-masked variants used on MNIST/Fashion-MNIST.

\subsection{Experimental Results}

\subsubsection{Main Comparison with Raw, Learned, and Top-$k$ Baselines}

\begingroup
\definecolor{lsMaroon}{HTML}{7A1F2B}
\definecolor{lsBand}{gray}{0.94}
\definecolor{lsRaw}{gray}{0.90}
\newcommand{\lsheading}[2]{\textcolor{lsMaroon}{\textbf{#1\hfill #2}}}
\newcommand{\lsbest}[1]{\textcolor{lsMaroon}{\textbf{#1}}}
\newcommand{\lsband}{\rowcolor{lsBand}}

\begin{table*}[t]
\centering
\scriptsize
\setlength{\tabcolsep}{3pt}
\vspace{3pt}
\caption{Effectiveness of \ourname on maintaining privacy in terms of SSIM, LPIPS, and MSE, as well as ability to maintain utility in comparison to baseline over multiple datasets. Positive values indicate privacy increase.}
\label{tab:main-compact}
\scaleTable{%
\begin{tabular}{@{}l c c c c c c c c c c c@{}}
\toprule
 & & \multicolumn{6}{c}{\textbf{Utility Acc.: Raw / $\Delta$\% vs.\ Raw}} & \multicolumn{4}{c}{\textbf{Attack metric: Raw / $\Delta$\% vs.\ Raw}} \\
\cmidrule[0.4pt](lr{0.125em}){3-8}
\cmidrule[0.4pt](l{0.125em}){9-12}
\textbf{Method} & \textbf{CR} & CIFAR-10 & CIFAR-100 & FMNIST & GTSRB & HAR & Non-IID mean & SDAR-USL & \multicolumn{2}{c}{FORA pseudo} & UnSplit \\
& & IID & IID & IID & IID & IID & $\overline{\Delta}$ Acc & SSIM\,$\downarrow$ & LPIPS\,$\uparrow$ & SSIM\,$\downarrow$ & MSE$_\mathrm{fg}$\,$\uparrow$ \\
\midrule
\rowcolor{lsRaw}
\textsc{Raw} & $1\times$ & 87.24 & 59.47 & 91.64 & 98.55 & 94.01 & 69.30 & 0.832 & 0.206 & 0.805 & $\times 1.00$\,/\,$\times 1.00$ \\
\midrule
\lsband
\textsc{LightSplit} (F\,/\,L) & $8\times$  & $-4.3\%$\,/\,$-4.4\%$ & $-11.2\%$\,/\,$-5.0\%$ & $-0.6\%$\,/\,$-0.5\%$ & $-2.2\%$\,/\,$-2.9\%$ & $+0.2\%$\,/\,$+0.5\%$ & $-3.3\%$\,/\,$-5.0\%$ & $-37.0\%$ & \lsbest{$+217.8\%$} & \lsbest{$-96.0\%$} & \lsbest{$+27\%$\,/\,$+9\%$} \\
\textsc{LightSplit} (F\,/\,L) & $16\times$ & $-6.0\%$\,/\,\lsbest{$-3.2\%$} & $-12.8\%$\,/\,$-8.2\%$ & $-0.9\%$\,/\,$-0.7\%$ & $-3.8\%$\,/\,$-3.6\%$ & $0.0\%$\,/\,$+0.2\%$ & $-5.0\%$\,/\,$-6.8\%$ & $-44.2\%$ & $+212.8\%$ & $-91.9\%$ & \lsbest{$+27\%$\,/\,$+9\%$} \\
\lsband
\textsc{LightSplit} (F\,/\,L) & $32\times$ & $-6.0\%$\,/\,$-5.3\%$ & $-13.7\%$\,/\,$-11.7\%$ & $-1.8\%$\,/\,$-1.3\%$ & $-5.6\%$\,/\,$-5.8\%$ & $0.0\%$\,/\,$+0.3\%$ & $-8.4\%$\,/\,$-8.3\%$ & \lsbest{$-58.3\%$} & $+212.0\%$ & $-93.3\%$ & $+27\%$\,/\,$+8\%$ \\
\addlinespace[2pt]
\textsc{Learned $1{\times}1$} & $8\times$  & \lsbest{$-0.7\%$} & $-4.1\%$ & \lsbest{$-0.5\%$} & \lsbest{$-0.5\%$} & $+0.8\%$ & \lsbest{$-2.8\%$} & $-18.5\%$ & $+97.7\%$ & $-30.1\%$ & $+3\%$\,/\,$+6\%$ \\
\lsband
\textsc{Learned $1{\times}1$} & $16\times$ & $-1.7\%$ & $-7.5\%$ & $-0.9\%$ & $-1.0\%$ & $-0.1\%$ & $-4.7\%$ & $-16.8\%$ & $+114.4\%$ & $-36.5\%$ & $-3\%$\,/\,$+8\%$ \\
\textsc{Learned $1{\times}1$} & $32\times$ & $-3.8\%$ & $-16.9\%$ & $-1.0\%$ & $-1.0\%$ & $+0.2\%$ & $-7.5\%$ & $-19.9\%$ & $+165.0\%$ & $-58.1\%$ & $+8\%$\,/\,$-7\%$ \\
\addlinespace[2pt]
\lsband
\topksparse & $8\times$ & $-1.8\%$ & \lsbest{$-3.8\%$} & $-0.5\%$ & $-1.1\%$ & $+0.1\%$ & $-4.9\%$ & $-13.6\%$ & $+78.9\%$ & $-28.7\%$ & $+12\%$\,/\,$+5\%$ \\
\topksparse & $16\times$ & $-3.0\%$ & $-5.8\%$ & $-1.1\%$ & $-2.2\%$ & $-0.8\%$ & $-4.6\%$ & $-19.2\%$ & $+64.8\%$ & $-17.9\%$ & $+21\%$\,/\,$+9\%$ \\
\lsband
\topksparse & $32\times$ & $-4.8\%$ & $-10.9\%$ & $-2.0\%$ & $-5.6\%$ & $-0.8\%$ & $-10.2\%$ & $-31.8\%$ & $+101.5\%$ & $-34.0\%$ & $+24\%$\,/\,$+8\%$ \\
\bottomrule
\end{tabular}%
}\\[2pt]
\begin{minipage}{\textwidth}
\footnotesize
\emph{Notes.} Utility: L2H-SCH cells of Table~\ref{tab:full_sweep_baseline_appendix}. SDAR: USL L4, CIFAR-10. FORA: pseudo decoder (App.~\ref{app:fora}). HAR: App.~\ref{app:har}, outside the Non-IID mean. UnSplit: $\mathrm{MSE}_{\mathrm{fg}}/\mathrm{Raw}\!-\!1$, \texttt{MnistNet} split-2, MNIST\,/\,F-MNIST; Attack cells: \textsc{LightSplit-F}, $\lambda_{\mathrm{WCC}}{=}0$. \topksparse: largest-$|h|$ coordinates kept, rest zeroed.
\end{minipage}
\end{table*}
\endgroup

Tab.~\ref{tab:main-compact} summarizes our default privacy-utility trade-off at the L2H-SCH setting. Raw activations are highly invertible across all three attacks. \textsc{SDAR} reaches USL level-4 SSIM $0.832$, \textsc{FORA}'s pseudo decoder an LPIPS $0.206$, and \textsc{UnSplit} recovers inputs nearly-pixel-perfectly at the leakiest split-2 cut, with MSE of $0.13$ (MNIST) and $0.30$ (Fashion-MNIST).

\textsc{Learned $1\!\times\!1$} mostly preserves utility, with IID accuracy dropping by only $0.5 - 3.8\%$ on CIFAR-10, FashionMNIST, and GTSRB, with CIFAR-100 the worst case at $-16.9\%$ at a $32\times$ compression. Nevertheless, the activations transmitted remain easily invertible compared to \ourname. \textsc{SDAR} SSIM falls only $16.8-19.9\%$, \textsc{FORA} LPIPS rises $+98\%/{+}114\%/{+}165\%$, (which still leaves a recognizable image (Fig.~\ref{fig:unified_recon_fig})) well under \ourname's gain at every ratio, and \textsc{UnSplit} MSE jumps between $-7\%$ and $+8\%$, relative to Raw, occasionally leaking \emph{more} than the case of \textsc{No Defense}. \topksparse preserves utility comparably, with IID accuracy dropping by $0.5-5.6\%$ on CIFAR-10, FashionMNIST, and GTSRB, and CIFAR-100 the worst case at $-10.9\%$ at a $32\times$ compression. Its activations also remain invertible: \textsc{SDAR} SSIM falls only $13.6-31.8\%$ across all CR ratios, \textsc{FORA} LPIPS rises only $+79\%$, $+65\%$ and $+102\%$, the weakest gain of the three baseline methods, and \textsc{UnSplit} MSE rises by $+12-24\%$ on MNIST and $+5-9\%$ on Fashion-MNIST.

In contrast, \ourname, even using $\lambda_{\text{WCC}}=0$, delivers substantially larger privacy gains against every image reconstruction attack and at every CR: \textsc{SDAR} SSIM falls $37.0\%/44.2\%/58.3\%$ across $\{8\times,16\times,32\times\}$, \textsc{FORA} LPIPS rises by $+218\%/{+}213\%/{+}212\%$, and \textsc{UnSplit} MSE rises by $+27\%$ on MNIST and $+8-9\%$ on Fashion-MNIST, even at the leakiest split-2 head. At $32\times$, its best setting, top-$k$ still removes less SSIM than \ourname removes at $8\times$, and costs $10.2\%$ of the non-IID mean to get there against \ourname's $3.3-8.4\%$ at any CR. Only \textsc{UnSplit}, the weakest attack, brings the two close. Utility costs are modest on simple datasets (FashionMNIST $-0.5\%$ to $-1.8\%$, GTSRB $-2.2\%$ to $-5.8\%$) and grow on feature-rich ones (CIFAR-10 $-3.2\%$ to $-6.0\%$, CIFAR-100 $-5.0\%$ to $-13.7\%$). On CIFAR-100, \ournameLearned recovers about half of the accuracy lost under \ournameFixed, matching what the intuition from \sect\ref{sec:method-liftback}. For complex tasks, the CNN backbone alone is not sufficient to deal with the rank-$k$ restriction, and putting an MLP at the $\mathbb{R}^{k}\!\to\!\mathbb{R}^{d}$ interface stabilizes the accuracy. \\

\begin{figure}[t]
    \centering
    \includegraphics[width=0.85\columnwidth]{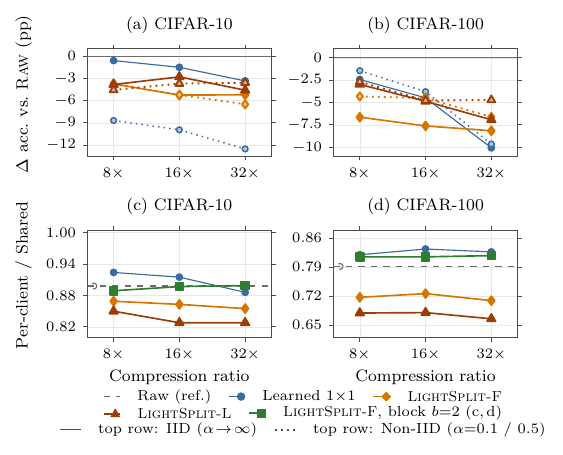}
        \caption{Utility of \ourname at the L2H cut. (a,\,b)~SCH: $\Delta$
    accuracy against matched \textsc{Raw}, marked by the zero line.
    (c,\,d)~IID: PCH\,/\,SCH accuracy ratio, where $1.0$ is no loss from
    per-client heads. The $b{=}2$ series in (c,\,d) is the
    block-local projection (\sect\ref{sec:method-block}).}
    \label{fig:utility_distilled}
\end{figure}

\subsubsection{Utility Behavior}
\label{subsec:utility_behavior}
Fig.~\ref{fig:utility_distilled} shows the changes in utility for CIFAR-10
and CIFAR-100 across CR, data distribution, and head-ownership configurations. The \emph{top row} (a, b) plots accuracy drop vs.\ \textsc{Raw} under IID and
Non-IID. \ournameGen cost stays within $\le\!8.2$\,pp on CIFAR-100 across all CRs and barely varies between IID and Non-IID scenarios. At low CRs it tracks
\textsc{Learned}~$1{\times}1$, but at higher CRs under Non-IID
\textsc{Learned}~$1{\times}1$ degrades faster: for CR=$32\times$ it loses
${\sim}12.5$\,pp on CIFAR-10 and ${\sim}9.6$\,pp on CIFAR-100, while
\textsc{\ournameLearned} stays within $\le\!7$\,pp on both. So
\ourname maintains the models' utility \emph{and} achieves significantly
stronger privacy, as reported in Table~\ref{tab:main-compact}. Further, we observe that \ournameLearned performs better for harder semantic tasks like CIFAR-100, while \ournameFixed is more robust on simpler datasets under Non-IID. FashionMNIST and GTSRB show milder versions of both trends
(App.~Table~\ref{tab:full_sweep_baseline_appendix}).

Subfigs. (c, d) reports the PCH\,/\,SCH accuracy ratio at IID. On CIFAR-10 the PCH drop is modest, both variants of \ourname maintain $83-87\%$ of SCH accuracy and \textsc{Learned}~$1{\times}1$ retains
${\sim}90\%$. On CIFAR-100 the drop increases: \textsc{\ournameLearned}
is reduced to $0.67$ and \textsc{\ournameFixed} to $0.71$ of SCH accuracy, while \textsc{Learned}~$1{\times}1$ is decreased to $0.83$. SCH becomes particularly relevant for complex tasks and \ourname is more sensitive to it than
\textsc{Learned}~$1{\times}1$. Per-client heads drift apart, and the global projection spreads each client's deviation over the whole lifted input the backbone reads. Projecting within $b{\times}b$ tiles (\sect\ref{sec:method-block}) confines the mixing to each tile and narrows the PCH gap to \textsc{Raw} from $10.8$ to $4.7$ points at $\lambda_{\mathrm{WCC}}{=}0$ (Tab.~\ref{tab:block_util_F}). App.~\ref{app:block} gives the full sweep. The $b{=}2$ series in (c,\,d) plots this remedy. For the L2H\,+\,SCH configuration
\ourname achieves a comparable utility as \textsc{Raw} while reducing privacy leakages (see Tab.~\ref{tab:main-compact}).

\paragraph{Scaling to ImageNet-1k}
\label{subsec:imagenet}
\ourname scales to ImageNet-1k (Fig.~\ref{fig:imagenet_generality}, Appendix~\ref{app:imagenet}). With $2{\times}2$ tiles it retains $92\%$ of the \textsc{Raw} top-1 at $8\times$, $85\%$ at $16\times$ and $71\%$ at $32\times$, ahead of both matched-budget baselines at $8\times$ and $16\times$ and level with top-$K$ at $32\times$.

\subsubsection{Beyond Image Classifiers}
\label{subsec:generality}
 We test \ourname on two non-image
settings of \sect\ref{sec:exp:setup}.

\paragraph{Tabular data}
On HAR, every \ourname variant stays within noise of the
$94.0\%$ no-defense baseline (Table~\ref{tab:main-compact}; full
sweep in App.~\ref{app:har}); the matched-budget learned $1{\times}1$ and top-$K$ sparsification land in the same band ($94.7\%$/$94.1\%$).

\paragraph{Language modelling}
On GPT-2 we test \ourname on fine-tuning. It works: with the cut held at a tenth of the model depth, perplexity inflation at CR\,$8$/$32\times$ falls from $1.15\times$/$1.36\times$ on the $124$M model to $1.06\times$/$1.11\times$ at $1.5$B (Fig.~\ref{fig:llm_generality} in App.~\ref{app:llm}).

\subsubsection{Privacy Attacks}
\label{subsec:privacy_attacks}
We report the attacks in three groups. Passive image reconstruction comes first: SDAR and FORA train a server-side substitute encoder that inverts the smashed activations, and UnSplit inverts each image on its own by gradient matching. We then repeat the reconstruction test on tabular data. The last group strengthens the adversary: the active FSHA attack, where the server hijacks client training, and attackers that optimize directly against the projection. We close with the privacy of the learned lift-back. 

\paragraph{SDAR}

\begin{table}[t]
\centering
\caption{Privacy attacks at CR\,$8\times$. \textsc{Raw} is absolute; other rows are the relative change vs.\ \textsc{Raw}; arrows point toward more privacy.}
\label{tab:privacy_all}
\label{tab:sdar-main-compact}
\label{tab:fora-privacy}
\label{tab:unsplit_attack_headline}
\setlength{\tabcolsep}{3pt}
\renewcommand{\arraystretch}{1.2}
\resizebox{0.95\columnwidth}{!}{%
\begin{tabular}{@{}l cc cc cc cc@{}}
\toprule
 & \multicolumn{2}{c}{\textbf{SDAR}} & \multicolumn{2}{c}{\textbf{FORA}} & \multicolumn{2}{c}{\textbf{UnSplit}} & \multicolumn{2}{c}{\textbf{Tabular probes}} \\
\cmidrule(lr){2-3}\cmidrule(lr){4-5}\cmidrule(lr){6-7}\cmidrule(lr){8-9}
\textbf{Method} & LPIPS\,$\uparrow$ & SSIM\,$\downarrow$ & LPIPS\,$\uparrow$ & SSIM\,$\downarrow$ & \multicolumn{2}{c}{$\mathrm{MSE}_{\mathrm{fg}}$\,$\uparrow$} & $R^2$\,$\downarrow$ & re-ID\,$\downarrow$ \\
 & & & & & M & F & & \\
\midrule
\textsc{Raw} (absolute) & $0.290$ & $0.832$ & $0.206$ & $0.805$ & $0.548$ & $0.537$ & $0.904$ & $0.507$ \\
\textsc{Raw} $+$ DP & \cellcolor{lsheat!18}$+57.7\%$ & \cellcolor{lsheat!9}$-41.6\%$ & \cellcolor{lsheat!30}$+212.3\%$ & \cellcolor{lsheat!30}$-95.6\%$ & \cellcolor{lsheat!30}$+22.8\%$ & \cellcolor{lsheat!9}$+6.5\%$ & \cellcolor{lsheat!18}$-19.6\%$ & \cellcolor{lsheat!30}$-51.9\%$ \\
\midrule
\textsc{Learned $1{\times}1$} & $+21.5\%$ & $-18.5\%$ & \cellcolor{lsheat!18}$+97.7\%$ & \cellcolor{lsheat!9}$-30.1\%$ & $+3.3\%$ & \cellcolor{lsheat!9}$+6.5\%$ & $-5.1\%$ & $-23.1\%$ \\
\topksparse & $+12.7\%$ & $-13.6\%$ & \cellcolor{lsheat!9}$+78.9\%$ & $-28.7\%$ & \cellcolor{lsheat!9}$+12.0\%$ & $+5.0\%$ & \cellcolor{lsheat!9}$-11.8\%$ & $-22.9\%$ \\
\midrule
\textsc{LS-F}, $\lambda{=}0$ & \cellcolor{lsheat!18}$+61.3\%$ & \cellcolor{lsheat!9}$-37.0\%$ & \cellcolor{lsheat!30}$+217.8\%$ & \cellcolor{lsheat!30}$-96.0\%$ & \cellcolor{lsheat!30}$+27.0\%$ & \cellcolor{lsheat!30}$+8.8\%$ & $-4.8\%$ & $-13.2\%$ \\
\textsc{LS-F}, $\lambda{=}10^{-3}$ & \cellcolor{lsheat!30}$+74.7\%$ & \cellcolor{lsheat!18}$-49.0\%$ & \cellcolor{lsheat!18}$+195.6\%$ & \cellcolor{lsheat!18}$-84.5\%$ & \cellcolor{lsheat!30}$+27.0\%$ & \cellcolor{lsheat!30}$+10.5\%$ & $-7.4\%$ & $-20.4\%$ \\
\textsc{LS-F}, $\lambda{=}10^{-2}$ & \cellcolor{lsheat!30}$+71.4\%$ & \cellcolor{lsheat!18}$-55.1\%$ & \cellcolor{lsheat!30}$+211.3\%$ & \cellcolor{lsheat!18}$-90.6\%$ & \cellcolor{lsheat!30}$\mathbf{+27.4\%}$ & \cellcolor{lsheat!30}$+8.8\%$ & \cellcolor{lsheat!9}$-10.9\%$ & $-30.1\%$ \\
\textsc{LS-F}, $\lambda{=}10^{-1}$ & \cellcolor{lsheat!30}$+83.9\%$ & \cellcolor{lsheat!9}$-36.7\%$ & \cellcolor{lsheat!30}$\mathbf{+221.0\%}$ & \cellcolor{lsheat!30}$\mathbf{-101.9\%}$ & \cellcolor{lsheat!30}$\mathbf{+27.4\%}$ & \cellcolor{lsheat!30}$+8.8\%$ & \cellcolor{lsheat!30}$\mathbf{-26.4\%}$ & \cellcolor{lsheat!30}$\mathbf{-55.0\%}$ \\
\textsc{LS-F} $+$ DP & \cellcolor{lsheat!30}$\mathbf{+86.5\%}$ & \cellcolor{lsheat!30}$\mathbf{-64.0\%}$ & \cellcolor{lsheat!30}$+212.1\%$ & \cellcolor{lsheat!30}$-96.9\%$ & \cellcolor{lsheat!30}$\mathbf{+27.4\%}$ & \cellcolor{lsheat!30}$\mathbf{+11.0\%}$ & \cellcolor{lsheat!18}$-17.4\%$ & \cellcolor{lsheat!18}$-43.4\%$ \\
\bottomrule
\end{tabular}%
}
\vspace{2pt}

\parbox{0.95\columnwidth}{\footnotesize\raggedright \textsc{LS-F} is \textsc{LightSplit-F}; UnSplit's M/F columns are MNIST and Fashion-MNIST; the tabular probes run on UCI HAR, re-ID chance is $0.048$. Attack protocols: App.~\ref{app:sdar_setup}, \ref{app:fora_setup}, \ref{app:unsplit_setup}; DP: App.~\ref{app:setup_dp}. Absolute values and full CR sweeps are in the per-attack appendix grids (App.~Tables~\mbox{\ref{tab:final-sdar-usl-all-split-points}}, \mbox{\ref{tab:fora-full}}, \mbox{\ref{tab:unsplit_attack_full}}; App.~Fig.~\mbox{\ref{fig:har_privacy}} for HAR). A structureless reconstruction can score slightly negative SSIM, so a drop can exceed $100\%$.}
\end{table}

\begin{figure}[t]
\centering
\includegraphics[width=0.8\columnwidth]{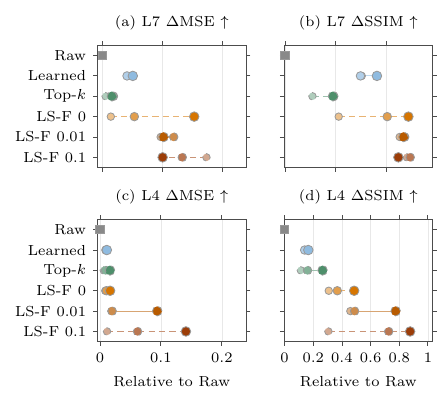}
\caption{USL SDAR reconstruction degradation relative to \textsc{Raw} at SL cutting $L{=}4$ and $L{=}7$. Large x-values mean the defense increased MSE or reduced SSIM compared with raw SDAR. Dot shading denotes CR $8/16/32\times$ from light to dark.}
\label{fig:sdar-usl-l7-bars}
\end{figure}

\begin{figure*}[t]
    \centering
    \includegraphics[width=0.75\textwidth]{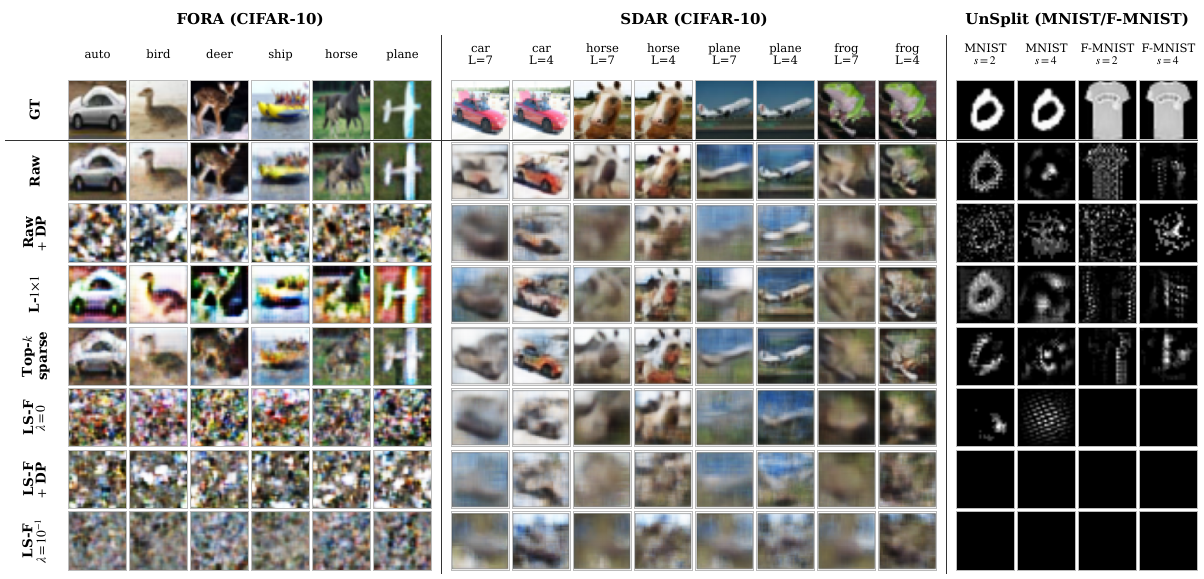}
        \caption{Reconstruction-attack comparison across the passive attacks; each column shows reconstructed samples.}%

    \label{fig:unified_recon_fig}
\end{figure*}

In Table~\ref{tab:sdar-main-compact} we report SDAR at the default USL $L{=}4$, CR$\,{=}\,8\times$. \textsc{Learned $1{\times}1$} barely shifts the leakage profile (LPIPS ${+}21.5\%$, SSIM ${-}18.5\%$): a trainable channel preserves the structure SDAR's surrogate exploits. \topksparse barely moves either metric (LPIPS ${+}12.7\%$, SSIM ${-}13.6\%$): the un-mixed coordinates it releases keep the structure SDAR's surrogate exploits.  \textsc{\ournameFixed} cuts SSIM by more than a third from the random projection alone (SSIM ${-}37.0\%$, LPIPS ${+}61.3\%$ at $\lambda_{\mathrm{WCC}}{=}0$), and WCC compounds the effect on the perceptual metric: ${+}74.7\%$ at $\lambda_{\mathrm{WCC}}{=}10^{-3}$, the recommended operating point, and up to ${+}83.9\%$ at $\lambda_{\mathrm{WCC}}{=}10^{-1}$. Fig.~\ref{fig:unified_recon_fig} shows the effect of WCC on SDAR reconstructions: at $\lambda{=}0$ the random projection $R$ leaves enough residual structure for the attack to recover the object shape, while at $\lambda{=}10^{-1}$ the reconstructions collapse further and become visually unidentifiable at the same CR level. The pixel metrics do not track $\lambda_{\mathrm{WCC}}$ monotonically at this cell: the SSIM drop returns to ${-}36.7\%$ at $\lambda_{\mathrm{WCC}}{=}10^{-1}$, level with its $\lambda_{\mathrm{WCC}}{=}0$ value, while LPIPS never falls back (${+}71\%$ or more under WCC, ${+}61.3\%$ without), because WCC makes the decoder return a class-typical image that the pixel metrics reward and LPIPS penalizes. Both stay far below the undefended attack; App.~\ref{app:sdar_metrics} reports how often the two metric families agree across the full sweep. Raising the compression deepens the effect: at $\lambda_{\mathrm{WCC}}{=}10^{-1}$ the SSIM drop reaches ${-}87.3\%$ at $16\times$ and ${-}105.3\%$ at $32\times$, where the reconstruction scores below zero (Table~\ref{tab:sdar-main-compact}).

Fig.~\ref{fig:sdar-usl-l7-bars} extends the analysis to the full USL sweep across split levels $\{4,7\}$ and every (CR, $\lambda_{\mathrm{WCC}}$) combination, and reveals two complementary effects of WCC on the $\Delta$MSE and $\Delta$SSIM relative to Raw. At $L{=}7$, $\lambda_{\mathrm{WCC}}{=}0$ produces a wide CR-dependent spread. WCC reduces information leakage regardless of the compression ratio. At $L{=}4$ the outcome changes: increasing the CR alone barely adds any additional privacy benefit, but combining WCC with high CR pushes both $\Delta$MSE and $\Delta$SSIM far above what either term reaches in isolation. \textsc{Learned $1{\times}1$} shows no such interaction at either level, with $\Delta$MSE and $\Delta$SSIM almost flat across CR. \topksparse stays in the same low band as \textsc{Learned $1{\times}1$} at $L{=}4$ and trails every method at $L{=}7$, where both its $\Delta$MSE and $\Delta$SSIM sit lowest at every CR. Further, we report the VSL counterpart and the full table in App.~\ref{app:sdar}.

\paragraph{FORA}

In Table~\ref{tab:fora-privacy} we report FORA at the default CR$\,{=}\,8\times$ slice using the \emph{pseudo} decoder. Every method in that slice sends the same payload, so the cells compare defences directly: \textsc{\ournameFixed} triples the attacker's perceptual error from the random projection alone (LPIPS ${+}217.8\%$), while \textsc{Learned $1{\times}1$} reaches ${+}97.7\%$. \topksparse reaches ${+}78.9\%$: releasing raw activation values at their own coordinates leaves the most for a decoder to learn. These percentages run high partly because \textsc{Raw}'s absolute LPIPS base is small (Table~\ref{tab:privacy_all}); the SSIM column confirms the gap at ordinary scale: ${-}28.7\%$ for \topksparse and ${-}30.1\%$ for \textsc{Learned $1{\times}1$} against ${-}96.0\%$ for \textsc{\ournameFixed}. 
Against the pseudo decoder, WCC adds no further gain as the projection alone is already sufficient (\textsc{\ournameFixed} moves only from ${+}217.8\%$ to ${+}221.0\%$ at $\lambda_{\mathrm{WCC}}{=}10^{-1}$); against the white-box reference decoder it lifts \textsc{\ournameFixed} from $\times3.8$ to $\times4.9$. Fig.~\ref{fig:unified_recon_fig} provides a visual confirmation on the automobile class: \textsc{\ournameFixed} reconstructions degrade into noise while \textsc{Raw}, \textsc{Learned $1{\times}1$}, and \topksparse remain identifiable. The white-box reference ceiling and the full CR$\,{\in}\,\{8,16,32\}\times$ results are also reported in Appendix~\ref{app:fora}. Both preserve the ordering with substantially larger ratios (up to $\Delta_{\text{Raw}}\,\times5.6$).

\paragraph{UnSplit}
Table~\ref{tab:unsplit_attack_headline} reports UnSplit at the default split-2 cut, the shallow point where UnSplit's gradient-matching loop is most informative. We use foreground-masked MSE so that trivially-matched zero pixels in MNIST/F-MNIST do not dominate the metric. \textsc{Learned $1{\times}1$} stays close to the \textsc{Raw} baseline, with $\Delta_{\text{Raw}}$ between ${-}7\%$ and ${+}8\%$ across CR$\,{\in}\,\{8,16,32\}\times$ on both datasets and several rows dipping below $0\%$ (App.~Table~\ref{tab:unsplit_attack_full}). \textsc{\ournameFixed} yields $\Delta_{\text{Raw}}\,{+}27\%$ on MNIST at every evaluated CR and $\lambda_{\mathrm{WCC}}$, and ${+}8\%$ to ${+}11\%$ on F-MNIST. \topksparse lands between them, $\Delta_{\text{Raw}}\,{+}12.0\%$ on MNIST and ${+}5.0\%$ on F-MNIST.

As shown in Fig.~\ref{fig:unified_recon_fig}, \textsc{Raw} recovers identifiable digit and garment silhouettes on both datasets. \textsc{Learned $1{\times}1$} still leaks the digit shape on MNIST split-2 and remains identifiable but visibly degraded at split-4, while on F-MNIST it produces scattered white pixels with no recognizable garment class. \topksparse behaves like \textsc{Learned $1{\times}1$}: the digit ring stays legible at split-2 and F-MNIST degrades to scattered pixels. The \textsc{\ournameFixed} columns are uniformly black across datasets and splits. The split-4 results and whole-image MSE variants are reported in Appendix~\ref{app:unsplit} and follow the same pattern.

\paragraph{Tabular reconstruction}
\label{subsec:tabular-privacy}
On HAR both probes weaken as compression tightens, and WCC weakens them further: with $\lambda_{\mathrm{WCC}}{=}10^{-1}$, \ournameFixed cuts median per-feature $R^2$ by $26.4\%$ and subject re-identification by $55.0\%$ at $8\times$, and the re-identification drop deepens to $61.4\%$ at $32\times$ (Table~\ref{tab:privacy_all}; sweeps in App.~Fig.~\ref{fig:har_privacy}). Without WCC, \ournameFixed, \textsc{Learned $1{\times}1$} and \topksparse stay close to each other on both probes; WCC is what separates them. The lower $R^2$ of \topksparse comes from its zeroed coordinates, which the regressor cannot recover, while its re-identification rate matches the other baselines (App.~\ref{app:tabular_privacy}).

\paragraph{Fully-malicious server (FSHA)}
\ournameFixed is the only compression mechanism that degrades the
attacker at unchanged accuracy, raising reconstruction LPIPS to $8.4\times$ (MNIST) and
$2.5\times$ (Fashion-MNIST) its no-defense value, while the learned $1{\times}1$ conv and
\topksparse leave it unimpeded (${\approx}\,1.2$--$1.5\times$); Table~\ref{tab:fsha_active} and App.~\ref{app:fsha_active} give the full results and the reconstructions.

\paragraph{Attackers optimized around the projection}
$R$ is public by construction, so we also build attackers directly
around it (App.~Table~\ref{tab:proj_opt}). Handing the adversary the analytic lift-back $zR^{\top}$ gains at most $+0.032$
SSIM over a decoder reading $z$ alone, and fusing every signal the server observes at the
cut ($z$, the returned gradient $g$, and the lift-back) into one decoder does not beat it
either: $R$ discards $89.8/95.6/98.5\%$ of the activation energy at CR\,$8/16/32\times$ (App.~\ref{app:energy_discarded}),
so it is the information removal that protects the client empirically; the secrecy of $R$ plays no part.

\paragraph{Privacy of the learned lift-back}
\label{sec:lsl-privacy}
\ournameFixed and \ournameLearned send the identical projected
payload, and the learned lift-back is server-side post-processing on an already-released
$k$-dimensional vector, so it cannot widen the adversary's view
(\sect\ref{sec:method-liftback}). Running SDAR against \ournameLearned at the $L{=}7$ cut confirms this (App.~Table~\ref{tab:sdar_lsl}): reconstruction SSIM is $0.18/0.12/0.14$ at CR\,$8/16/32\times$ under \ournameLearned and $0.36/0.17/0.08$ under \ournameFixed, both far below the no-defense $0.57$. LPIPS$_{20}$ shows the same parity: $0.64$--$0.66$ under \ournameLearned against $0.57$--$0.68$ under \ournameFixed.

\subsubsection{Differential Privacy}
\label{subsec:dpbody}
As a formal guarantee we also apply DP at the cut: $\sigma{=}0.02$ gives the best accuracy trade-off, keeping ${\approx}\,97\%$ of the un-noised accuracy (noise sweep in App.~\ref{app:dp}). We run the attacks at that noise level. \ournameFixed$+$DP carries the largest SDAR drop of the comparison at the default $L{=}4$ cut (SSIM ${-}64.0\%$), ahead of DP alone (${-}41.6\%$) and \ournameFixed alone (${-}37.0\%$); Table~\ref{tab:privacy_all} reports \textsc{Raw}$+$DP and \ournameFixed$+$DP against SDAR, FORA, UnSplit and the tabular HAR probes, and App.~Table~\ref{tab:fsha_active} extends the composition to the active FSHA. The deeper $L{=}7$ cut shows the composition sharper still: SSIM falls from $0.572$ to $0.360$ under \ournameFixed alone, to $0.181$ under DP alone, and to $0.139$ with both; the two defenses add up, and DP and the projection together are more private than either alone (See App.~Table~\ref{tab:dp_privacy}).

\paragraph{Compression-ratio sweep}
We sweep the compression ratio from $1\times$ to $4096\times$ ($k{=}1$). Accuracy is nearly flat across the $8$--$32\times$ band we use: CIFAR-10 keeps above $93\%$ of its uncompressed accuracy and CIFAR-100 above $85\%$, and gentler ratios buy the rest back at a proportionally larger payload. The curves break only well past the band: CIFAR-10 still holds $89\%$ at $512\times$ ($k{=}8$) and $55\%$ at $4096\times$ ($k{=}1$), while CIFAR-100 falls earlier, to $63\%$ at $512\times$ and $12\%$ at $4096\times$ (Fig.~\ref{fig:dp_lowcr}(a), App.~\ref{app:lowcr_results}).

\subsubsection{Backdoor Detection Signal}
\label{subsec:backdoor}
While not the focus of this work, \ournameGen ability to emphasize class-level structure also highlights inconsistencies between smashed data, being characteristic for backdoored inputs (see \sect\ref{sec:exp:setup}). To evaluate this intuition, we run a basic backdoor experiment to determine whether \ournameGen compression interferes with the cosine-similarity signal the server uses to flag malicious clients. App.~Fig.~\ref{fig:backdoor_r} shows the full results over Dirichlet concentration $\alpha$, client number $n$, malicious-client fraction $\mu$, and per-client poison ratio $p$. Notably, the figure shows across the four methods (\textsc{Raw}, \textsc{Learned $1{\times}1$}, \textsc{\ournameFixed}, \textsc{\ournameLearned}) that \ourname has the potential to support the server by recognizing poisoned training contributions as malicious clients remain identifiable from the server-side backbone activations, with $|\text{MAD-}z|$ above the $3\sigma$ threshold in most cells of the sweep. Detection weakens for strongly non-IID data ($\alpha\,{\leq}\,1$ in subfigure~\ref{fig:backdoor_r}a) and we observe minor drops in a few $n$ and $p$ cells. Thus, especially in more IID settings, \ourname does not degrade the capability to recognize poisoned training contributions.

\subsubsection{Ablations}
For the ablations on WCC strength, client-count scaling, and the \textsc{\ournameLearned} lift-back hidden width, see Appendix~\ref{app:ablations}.

\subsubsection{Deployment Guidance}
\label{sec:deployment}
\ournameFixed is the default lift-back. \ournameLearned is beneficial where the backbone cannot absorb the structure the projection removes, as on CIFAR-100, where it recovers accuracy over \ournameFixed at every ratio (Table~\ref{tab:main-compact}). The choice of projection follows head ownership. With a shared head, the global projection with WCC is preferable: its privacy gain is largest at gentle compression ratios and shrinks as the ratio grows, where the projection alone removes most of the signal (App.~Table~\ref{tab:final-sdar-usl-all-split-points}). With per-client heads, the block-local projection with DP should be used: tiling absorbs the drift between client heads and narrows the gap to \textsc{Raw} from $10.8$ to $4.7$ points (App.~Table~\ref{tab:block_util_F}), smaller tiles return part of the privacy (App.~Table~\ref{tab:block_privacy_unsplit}), and DP restores it at every tile size at little utility cost (App.~Tables~\ref{tab:block_privacy_unsplit} and~\ref{tab:dp_utility}).

\section{Discussion}
\label{sec:discussion}
In the previous sections, we introduced \ourname that employs orthogonal projections building on the JL-lemma, together with the WCC constraint to limit information exposure during the training and reduce communication overhead. In the following, we discuss its security (\sect\ref{sec:discussion-security}) as and limitations (\sect\ref{sec:discussion-limitations}).

\subsection{Security Considerations}
\label{sec:discussion-security}
We defined the adversarial threat model in \sect\ref{sec:problem-threat} and derived the requirements for a secure and practical SL defense in \sect\ref{sec:problem-requirements}. In \sect\ref{sec:eval}, we extensively evaluated \ourname under multiple reconstruction attacks and system settings, demonstrating that it effectively limits privacy leakage while substantially reducing communication overhead. Although the server knows projection-matrix R, rank-reduction/information-bottleneck causes inherent information loss (pigeonhole-principle/rank-deficiency), as described in \sect\ref{sec:method-liftback}. Together with WCC-loss, this is mainly responsible for \ournameGen privacy-gain. The information retained in the reconstructable signal seems to focus on task-relevant information rather than information allowing reconstructing individual samples. Our extensive evaluation shows significantly degraded performance. Thus, information reduction seems to affect particularly subtle differences between samples that these attacks exploit. While \ourname does not inherently provide formal guarantees (see \sect\ref{sec:discussion-limitations}), we showed its effectiveness in our extensive evaluation and demonstrated in \sect\ref{subsec:dpbody} that \ourname can be combined with Differential Privacy to achieve formal privacy gains. In the following, we discuss how the design of \ourname addresses the identified requirements and clarify its remaining limitations.

\ourname addresses R1 (attack prevention) by limiting the amount of instance-specific information exposed at the cut layer. Clients transmit only the projected representation, irreversibly discarding information before it leaves the client. WCC further suppresses residual intra-class variation exploitable by reconstruction attacks. As shown in \sect\ref{sec:eval}, this significantly reduces reconstruction quality across multiple attacks. To preserve R2 (model utility), the projection maintains the geometric structure of the representation space, while the server-side lift-back mechanisms preserve compatibility with existing SL architectures. As demonstrated in \sect\ref{sec:eval}, \ourname retains most of the baseline accuracy even under aggressive projection ratios. Finally, \ourname addresses R3 (communication efficiency) by reducing the dimensionality of transmitted cut-layer activations and returned gradients from $d$ to $k$, directly lowering communication overhead with only negligible client-side computation.

\noindent\textbf{Out-of-Scope Threats at the Cut-Layer Interface:}
\ourname targets passive input reconstruction and does not defend against other attacks on the smashed-data channel.
Label leakage attacks~\cite{li2022labelleak,liu2023clustering} exploit cut-layer gradients or similarity-based clustering, orthogonal to per-sample activation compression.
Active servers (e.g., FSHA~\cite{pasquini2021unleashing}) deviate from the protocol and are caught by client-side detection mechanisms~\cite{erdogan2022splitguard,fu2023gradscrutinizer}. We nevertheless evaluate FSHA as an active-attack stress test (App.~\ref{app:fsha_active}): \ournameFixed alone raises the reconstruction error to $8.4\times$ (MNIST) and $2.5\times$ (Fashion-MNIST) the undefended value at unchanged accuracy, while the learned $1{\times}1$ conv and top-$k$ sparse leave the attack unimpeded.

Model poisoning and backdoor injection through the gradient channel are likewise outside the scope of this work. While first observations suggest that class separation causes only a negligible drop in the server's ability to detect backdoors, a robust interpretation of this correlation requires more comprehensive evaluation of backdoor attacks by future work.\\
\noindent\textbf{Residual Leakage at the Class Level:}
WCC pulls same-class activations toward shared centroids, deliberately collapsing the per-sample signal that reconstruction attacks exploit. The server can still recover class-level prototypes from the smashed-data stream. We treat this as the privacy floor of \ourname: an attacker can at best learn what an average sample of a given class looks like, not the specific input.

\subsection{Limitations}
\label{sec:discussion-limitations}
We discuss \ournameGen limitations and how they affect efficiency and practical applicability.\\
\noindent\textbf{Restriction to Attacks Analyzing Smashed Data:} \ourname limits sample-specific information in smashed activations and addresses privacy risks from a curious server. White-box inference attacks that exploit shared model updates are a separate threat. \ourname operates solely on smashed data and provides no protection against them. Sharing model updates is a design choice that can be adapted to the threat model, whereas transmitting smashed activations is fundamental to SL. We therefore focus on the cut-layer interface and leave shared-parameter settings to future work.\\
\noindent\textbf{Formal Guarantees:}
\ournameGen privacy gain is motivated by formal concepts (pigeonhole principle/rank-deficiency) that enforce information reduction. A fixed linear view with $k{\ll}d$ cannot retain all of the activation's detail, while the Johnson-Lindenstrauss lemma explains why task-relevant structures survive~\cite{johnson1984extensions}. In our extensive evaluation, we showed \ournameGen effectiveness, while plain \ourname does not provide formal guarantees. To achieve formal guarantees, \ourname can be combined with Differential-Privacy~\cite{abadi2016deep}, as shown in  \sect\ref{subsec:dpbody}.

\noindent\textbf{Overhead through Learnable Backlifting:} \ournameLearned uses a trainable MLP for lift-back, adding parameters and compute. This cost falls on the server, assumed to have sufficient resources (see \sect\ref{sec:problem-system}). In resource-constrained scenarios, \ournameFixed provides a parameter-free alternative. As shown in \sect\ref{sec:eval}, the performance gap is tolerable.

\section{Related Work}
\label{sec:sota}
\subsection{Privacy Threats at the Split Interface}
\label{sec:rw:attacks}

In SL, raw data stays on the client, but the smashed-data interface still leaks substantial private information. In our work, we focus on \emph{training-time} attacks: the server exploits the smashed-data stream observed during collaborative training. \emph{Inference} attacks~\cite{he2019model,ginver2023} target a frozen model at deployment and are outside the scope of our work.

\paragraph{Passive reconstruction attacks}
Gradient-based methods (DLG~\cite{zhu2019deepleakagegradients}, iDLG~\cite{zhao2020idlg}, Inverting Gradients~\cite{geiping2020inverting}) recover training data from gradient signals. In SL the threat is amplified: the server observes \emph{per-sample} activations rather than aggregated updates. More recent attacks train a surrogate of the client encoder.
UnSplit~\cite{erdogan2022unsplit} alternates between optimising a clone encoder and a candidate input from the smashed-data stream, requiring only the client's architecture and no auxiliary data. PCAT~\cite{285505} drops the architecture assumption and builds a pseudo-client from the server's own evolving checkpoints. FORA~\cite{xu2024stealthy} trains a substitute encoder by feature-level transfer on public data. SDAR~\cite{zhu2025passiveinferenceattackssplit} is the current state of the art among passive attacks. It trains a simulator $\tilde{f}$ on auxiliary data while keeping the server's backbone $g$ frozen. The attack exploits unintentional memorisation: $g$, co-trained with the real client encoder $f$, retains distributional information about the private data~\cite{carlini2019secret}. Two adversarial discriminators align the simulator's activations with the client's and force the decoded outputs to look realistic.
SDAR matches the reconstruction quality of the active attack FSHA~\cite{pasquini2021unleashing} while remaining fully passive. These surrogate attacks interpret the smashed-data distribution rather than inverting it analytically. A compression scheme that preserves task structure may therefore preserve enough signal to train them. We evaluate this directly.

\paragraph{Active reconstruction attacks and detection}
FSHA~\cite{pasquini2021unleashing} reconstructs inputs by hijacking the client's encoder through manipulated gradients.
Such protocol deviations are detectable~\cite{erdogan2022splitguard,fu2023gradscrutinizer}, leaving passive reconstruction as the primary practical threat and our main focus. We still report FSHA as an active-attack stress test (App.~\ref{app:fsha_active}).

\paragraph{Label leakage}
Cut-layer gradients can reveal training labels~\cite{li2022labelleak}, and similarity-based clustering achieves label inference at the same interface~\cite{liu2023clustering}.
Label leakage is outside our focus and left to future work.

\subsection{Existing Defenses}
\label{sec:rw:defences}

\paragraph{Representation perturbation}
NoPeek~\cite{vepakomma2020nopeek} penalizes input-activation correlation via distance correlation. Shredder~\cite{mireshghallah2020shredderlearningnoisedistributions} injects learned additive noise. DISCO~\cite{singh2021disco} learns a pruning filter that obfuscates sensitive channels at inference time. Patch shuffling~\cite{patchshuffle2020} permutes spatial patches in vision transformers. None of these reduces the communication payload, and adaptive attackers can defeat them by learning latent privacy features from the defended representations~\cite{zhu2025passiveinferenceattackssplit}.
Because \ourname targets joint compression and privacy, we compare against compression baselines at matched ratios and against DP noise (\sect\ref{subsec:dpbody}).

Censor targets privacy attacks in federated learning. By letting clients sample multiple candidates of parameter updates and determine the most suitable one, it seeks to avoid privacy leakage~\cite{zhang2025censor}. However, creating multiple projections of activations and selecting the best is infeasible in SL, as assessing quality requires involving the server and sharing these options, including privacy-leaking projections. Instead, SL requires deterministic projections that preserve privacy and can be calculated on the client, as demonstrated by \ourname.

\paragraph{Cryptographic mechanisms}
FedVS~\cite{li2023fedvs} employs secret sharing to provide information-theoretic confidentiality of uploaded embeddings.
Homomorphic encryption (HE) has also been proposed for privacy-preserving neural network computation~\cite{gilad2016cryptonets}, enabling the server to process ciphertexts without observing plaintext activations, at orders-of-magnitude computational overhead unsuitable for edge deployments.

\subsection{Communication Compression at the Cut Layer}
\label{sec:rw:compression}

Top-$k$ sparsification transmits the $k$ largest activations and their indices~\cite{randtopk2023}; mask-encoded variants replace the indices with a bit-mask~\cite{ZhouQLCY24}. Both expose which positions are most active per input. Learned codecs train an encoder/decoder pair across the cut layer (BottleNet++~\cite{bottlenet2020}, FrankenSplit~\cite{frankensplit2024}, the latter additionally requiring ImageNet pretraining), adding trainable parameters and forward-pass cost to the client. SplitFC~\cite{splitfc2025} drops low-variance features and quantizes the rest with closed-form adaptive levels: parameter-free on the client but still reveals which positions are kept. C3-SL~\cite{hsieh2022c3slcircularconvolutionbasedbatchwise} superposes a batch of features via circular convolution with fixed random keys, reducing transmissions per round but not the per-sample payload.

\subsection{Random Projections in Distributed Learning}
\label{sec:rw:rp}

Random projections appear in distributed learning, but applied to different objects: RanPAC~\cite{mcdonnell2024ranpacrandomprojectionspretrained} projects pretrained features for continual learning, Han et al.~\cite{JMLR:v25:23-0215} project local model parameters for personalized FL, and ProjPert~\cite{projpert2024} projects backward gradients for label protection in vertical FL. To our knowledge, no prior work applies a random projection at the SL cut layer for per-sample activation compression.

\subsection{Positioning Our Approach}
\label{sec:rw:position}

Existing methods leave a gap that motivates our work. Privacy mechanisms do not compress.
Compression mechanisms either leak structural information (top-$k$ and variants), add client-side overhead (learned codecs), or do not reduce the per-sample payload (C3-SL).
No existing method simultaneously compresses, avoids client overhead, and restructures the server's observation under a semi-honest threat model.

We address this gap with a fixed, non-trainable orthogonal projection at the cut layer.
Clients map their $d$-dimensional activation to $k$ dimensions via $\mathbf{R}$, reducing the per-sample payload by $d/k$ without trainable client parameters. The server, which knows $\mathbf{R}$, either lifts back via $\mathbf{R}^\top$ (\ournameFixed) or applies a small learned module on the compressed representation (\ournameLearned). Both variants differ only in server-side capacity. An optional WCC regularizer pulls same-class activations toward shared centroids, removing the per-sample signal that reconstruction attacks exploit while preserving class separability.
The result is a representation limiting attackers to recovering class-level prototypes, not a private input.

We do not claim formal privacy guarantees. Formal results for JL projections require added noise~\cite{blocki2012jlprivacy,kenthapadi2013privacyjl}. Plain \ourname adds none; \sect\ref{subsec:dpbody} evaluates the combination with DP noise. We measure the privacy benefit empirically under state-of-the-art attacks with full knowledge of $\mathbf{R}$.

\section*{Acknowledgment}
This work was supported by the state of Styria, Austria  (project Cybersecurity-Next). This research also received funding from the European Union's Horizon program under grant agreements No. 101335589 (MIRAGE) and No. 101055025 (HYDRANOS), as well as the Federal Ministry of Education and Research of Germany (BMBF) within the IoTGuard project.

\appendix
\section{Appendix}
\label{sec:appendix}

\section{Experimental Setup}
\label{app:setup}

\begin{table*}[!tbp]
\centering
\small
\caption{Model partitions for U-shaped SL. Parameter counts of the client head at both cut depths, the server backbone, and the client tail; $d$ is the cut activation the projection compresses. The \ournameLearned lift-back (Linear$\to$BN$\to$ReLU$\to$Linear) is optional and server-side. Tabular and text models have a single client cut.}
\label{tab:model_partitions}
\scaleTable{%
\begin{tabular}{llrrrrcrr}
\toprule
& & \multicolumn{2}{c}{\textbf{Client head}} & \textbf{Server} & \textbf{Client} & \textbf{Cut} & \multicolumn{2}{c}{\textbf{\ournameLearned lift-back}$^{*}$} \\
\cmidrule(lr){3-4}\cmidrule(lr){8-9}
\textbf{Model} & \textbf{Dataset} & \textbf{L1H} & \textbf{L2H} & \textbf{backbone} & \textbf{tail} & $\boldsymbol{d}$ & $\boldsymbol{h_{\text{MLP}}}$ & \textbf{Params} \\
\midrule
\multicolumn{9}{l}{\emph{Image}}\\
ResNet-18   & CIFAR-10       & 9{,}536  & 83{,}520 & 11{,}166{,}976 & 5{,}130   & $64 \times 8 \times 8$    & 512   & 2{,}364{,}928  \\
ResNet-18   & CIFAR-100      & 9{,}536  & 83{,}520 & 11{,}166{,}976 & 51{,}300  & $64 \times 8 \times 8$    & 2{,}048 & 9{,}447{,}424  \\
ResNet-50 & ImageNet-1k & 9{,}536 & 83{,}520 & 23{,}498{,}496 & 2{,}049{,}000 & $64 \times 8 \times 8$ & 512 & 2{,}364{,}928 \\
MicronNet   & GTSRB          & 824      & 8{,}451  & 411{,}192      & 12{,}943  & $29 \times 14 \times 14^{\ddagger}$  & 128   & 824{,}500     \\
SimpleCNN   & FMNIST         & 520      & 4{,}160  & 435{,}162      & 5{,}130   & $20 \times 12 \times 12$  & 128   & 417{,}984     \\
\midrule
\multicolumn{9}{l}{\emph{Tabular}}\\
MLP & UCI HAR & 210{,}688 & -- & 99{,}456 & 774 & $256$ & 256 & 74{,}752\\
\midrule
\multicolumn{9}{l}{\emph{Text}}\\
GPT-2 (124M) & WikiText-103 & 46.5\,M  & -- & 78.0\,M  & tied$^{\S}$ & $768$  & 512 & 444{,}672\\
GPT-2 (355M) & WikiText-103 & 77.7\,M  & -- & 277.1\,M & tied$^{\S}$ & $1024$ & 512 & 592{,}384\\
GPT-2 (774M) & WikiText-103 & 144.3\,M & -- & 629.7\,M & tied$^{\S}$ & $1280$ & 512 & 740{,}096\\
GPT-2 (1.5B) & WikiText-103 & 235.8\,M & -- & 1.32\,B  & tied$^{\S}$ & $1600$ & 512 & 924{,}736\\
\bottomrule
\end{tabular}%
}\\[2pt]
\parbox{0.95\textwidth}{\footnotesize\raggedright
$^{*}$At CR$\,{=}\,8\times$; the parameter count shrinks linearly with $k$. $h_{\text{MLP}}$ is the per-dataset hidden width chosen in App.~\ref{app:mlp_hidden_ablation}.\\
$^{\ddagger}$MicronNet's first max-pool is moved from the backbone into the head, so $d{=}29 \times 14 \times 14{=}5684$ instead of $29 \times 28 \times 28$; parameter counts are unchanged.\\
$^{\S}$The LM head shares its weights with the client embeddings, so there is no separate client tail.}
\end{table*}

\subsection{Model partitions}
\label{app:models}
Table~\ref{tab:model_partitions} lists the model partitions used throughout the paper: for each architecture and dataset, the parameter counts of the client head at both cut depths (L1H and L2H), the server backbone, and the client tail, together with the cut activation shape and the lift-back MLP width used by \ournameLearned.

\subsection{Utility experiments}
\label{app:setup_utility}
\sect\ref{sec:exp:setup} states the configuration of the baseline utility experiments on the four image datasets. In this appendix we define the baselines and give the setups that extend them: ImageNet-1k, the block-local projection, the tabular and language experiments, the lower compression ratios, and differential privacy.

\subsubsection{Baselines and compression ratios}
\label{app:setup_baselines}
\textsc{Raw} transmits the smashed activation unmodified. \textsc{Learned $1{\times}1$} is a $1{\times}1$ convolution that compresses the $C$-channel feature map to $k$ channels at the client, paired with a $1{\times}1$ convolution on the server that restores $C$ channels~\cite{bottlenet2020}; it isolates the benefit of a trainable client-side compressor, a learnable codec, without reproducing BottleNet++'s full channel-aware codec. Because it compresses along channels, its CR is approximated to the nearest realizable channel count. Top-$k$ sparse keeps the $k_s{=}\lfloor 4k/6\rfloor$ largest-magnitude activation coordinates and transmits their \emph{true} values, zeroing the rest, sized to the same $512/256/128$ transmitted floats as CR\,$8/16/32\times$. DP clips each per-sample payload to an $L_2$ bound and adds calibrated Gaussian noise (App.~\ref{app:setup_dp}).

\subsubsection{ImageNet-1k}
\label{app:setup_imagenet}
We split a ResNet-50 over $10$ IID clients. Every client runs the same head: one extra residual block past the L1H cut, pooled to an $8{\times}8{\times}64$ map, so the cut stays $d{=}4096$ and matches the CIFAR cut. We train with SGD (momentum $0.9$, learning rate $0.1$ at batch $256$, weight decay $10^{-4}$, label smoothing $0.1$), mixed precision. We sweep CR\,${\in}\,\{8,16,32\}\times$ against \textsc{Raw}, a learned $1{\times}1$ convolution, and top-$K$ sparsification, all at the same payload size, and for \ourname also the tile size $b\,{\in}\,\{1,2,4,8\}$ (\sect\ref{sec:method-block}): $b{=}1$ projects each grid position on its own, $b{=}8$ covers the whole grid and recovers the global projection of Eq.~\eqref{eq:projection}.

\subsubsection{Block-local projection}
\label{app:setup_block}
The block-local variant (\sect\ref{sec:method-block}) operates on spatial cut maps: block size $b$ tiles the $H{\times}W{\times}C$ activation into $b{\times}b$ patches and projects each patch with a shared fixed $\mathbf{R}_b$, so $b{=}8$ recovers the global projection of Eq.~\eqref{eq:projection} and $b{=}1$ mixes only across channels at each position. With $T$ tiles, $\mathbf{R}_{b}\in\mathbb{R}^{b^{2}C\times k/T}$, so the client stores a $T^{2}$-times smaller matrix at the same compression ratio.

\subsubsection{Tabular data (UCI HAR)}
\label{app:setup_generality}
UCI HAR~\cite{human_activity_recognition_using_smartphones_240} has $561$ inertial features per window and six activities; we keep its official subject-disjoint train and test split. The client head is a two-layer MLP ($561\,{\to}\,256\,{\to}\,256$, batch norm and ReLU), the server backbone maps $256\,{\to}\,128$, and the client tail closes the U shape; the $d{=}256$ cut is already narrower than the input, so no pseudo-inverse recovers it. We transmit $k\,{\in}\,\{8,16,32\}$ scalars (CR\,${=}\,256/k$) and train with Adam (learning rate $10^{-3}$, batch $128$) over three seeds with $n{=}5$ clients; the heterogeneous partitions draw each client's label shares from a Dirichlet$(\alpha)$ distribution; the WCC arms use $\lambda_{\mathrm{WCC}}{=}10^{-1}$. The tabular attack configuration is in
App.~\ref{app:tabular_attack}.

\subsubsection{Language modelling (GPT-2)}
\label{app:setup_llm}
We fine-tune four GPT-2~\cite{radford2019language} sizes, $124$M to $1.5$B ($12$ to $48$ blocks, $d_{\mathrm{model}}$ $768$ to $1600$). The client keeps $\mathrm{round}(0.1L)$ blocks, that is $1$ to $5$, so the cut stays a tenth of the depth as the backbone grows. We project the residual state at that cut (CR\,${=}\,d_{\mathrm{model}}/k$) and
sweep CR\,${\in}\,\{8,16,32\}\times$, so at CR\,$8\times$ the payload keeps $k{=}96$ dimensions on the $124$M model and $k{=}200$ on the $1.5$B model. We train on WikiText-103~\cite{merity2016pointer} over $80{,}000$ blocks of $256$ tokens (about $20.5$M tokens) at learning rate $5\times10^{-5}$ and batch $8$ ($4$ for the $1.5$B model, which changes memory, not perplexity). We evaluate on $1{,}500$ held-out blocks and report the best validation perplexity as an inflation ratio over \textsc{Raw} at the same budget. \ournameLearned uses a lift-back MLP of width $512$.

\subsubsection{Compression range and cut width}
\label{app:setup_lowcr}
Two knobs set the size of the cut: the compression ratio, which decides how many of the $d$ cut activations survive as the payload $k$, and the cut width $d$ itself. We sweep them separately. Fig.~\ref{fig:dp_lowcr}(a) sweeps \ournameFixed across the full compression range at the default $d{=}4096$, from the uncompressed anchor (CR\,$1\times$, $k{=}d{=}4096$) all the way to the extreme $4096\times$ ($k{=}1$). Fig.~\ref{fig:cutwidth} sweeps the width instead: on CIFAR-10 we keep the cut depth, its $8{\times}8$ grid, the client head body and the server backbone fixed, and vary only the number of channels crossing the cut, so $d\in\{512,1024,2048,4096,8192\}$ with the default $d{=}4096$ inside the sweep. (Moving between the $L7$ and $L4$ cuts would change the width too, but it also changes the layer, its spatial grid and its abstraction level, so it does not isolate the width.) At every width we measure SDAR reconstruction and held-out accuracy at CR\,$8/16/32\times$ and against that width's own \textsc{Raw} run.

\subsubsection{Differential privacy}
\label{app:setup_dp}
To obtain a formal guarantee we add DP at the cut: each client norm-clips its transmitted payload to a per-sample $\ell_2$ bound and adds calibrated Gaussian noise (standard deviation $\sigma\,{\cdot}\,\text{clip}$), in training and at evaluation, so the noise is part of the released representation rather than a training-only regularizer. A noise sweep on CIFAR-10 (L2H) locates the operating point $\sigma{=}0.02$, where DP keeps ${\approx}\,97\%$ of the un-noised accuracy.

\subsection{Attack experiments}
\label{app:setup_attacks}

\subsubsection{SDAR}
\label{app:sdar_setup}

We adopt the SDAR~\cite{zhu2025passiveinferenceattackssplit} pipeline at the upstream defaults; the only edit is the insertion of the method object at the cut, applied identically to the target path and to the simulator path so that \textsc{Raw} reduces to the released attack.

USL is our operating point (\sect\ref{sec:problem-threat}). SDAR also releases the attack for vanilla SL, so we run every method under both protocols to show that the defense does not depend on the protocol. In the U-shaped setting (\textsc{USL}) the client keeps the head and the tail and the server never sees the labels; in the vanilla setting (\textsc{VSL}) the server holds the classifier and the labels. \ourname sits at the same head-backbone cut in either case, so the \textsc{USL} and \textsc{VSL} column groups of Table~\ref{tab:final-sdar-usl-all-split-points} measure the method under each protocol, at every split point and compression ratio.

\textbf{Per-method $k$ values.} For \textsc{\ournameFixed} at L7,
$\CR{=}d/k\in\{8,16,32\}$ fixes $k\in\{512,256,128\}$; at L4 the
same ratios fix $k\in\{1024,512,256\}$. The flat activation is
projected onto an orthonormal fixed random basis, then lifted back
into the original smashed-feature shape before $g$.
\textsc{Learned $1{\times}1$} keeps the same nominal CR with
$k_\mathrm{ch}\in\{8,4,2\}$ at L7 and $\{4,2,1\}$ at L4.

\textbf{LPIPS$_{20}$ computation.} MSE and SSIM are the full client-set averages
printed by SDAR's \texttt{evaluate()}; PSNR is computed from that MSE
as $-10\log_{10}(\mathrm{MSE})$ with image range $[0,1]$.
LPIPS$_{20}$ is computed afterward with LPIPS-Alex on the saved
20-image reconstruction grid, because the SDAR artifacts save loss
histories and PNGs but not full-dataset reconstruction tensors.

\textbf{Block-local projection.} We run the block variant of App.~\ref{app:setup_block} under SDAR, FORA, UnSplit and FSHA at every tile size $b$ with $\lambda_{\mathrm{WCC}}{=}0$, and repeat every tile size at $8\times$ with the DP noise of App.~\ref{app:setup_dp} aadded under SDAR, UnSplit and FSHA. Only SDAR also sweeps $\lambda_{\mathrm{WCC}}$ on the block variant: it is the strongest attack we run, the only one whose decoder co-trains with the victim, and it reconstructs the undefended $L{=}4$ cut at SSIM $0.83$ (Table~\ref{tab:final-sdar-usl-all-split-points}), so a WCC setting that holds under SDAR holds under the others.

\subsubsection{FORA}
\label{app:fora_setup}

We adopt the FORA~\cite{xu2024stealthy} pipeline at the upstream defaults (substitute encoder, residual discriminator with $\lambda_{\mathrm{GP}}{=}1000$, \texttt{cifar\_decoder} inversion stack, auxiliary public set, and the MK-MMD objective); the only edit is the insertion of the method object at the released cut.

\textbf{Per-method $k$ values.} The compression sweep $\CR{=}d/k{\in}\{8,16,32\}$ fixes $k{\in}\{2048,1024,512\}$ on the
flat dimension for \textsc{\ournameFixed}, and $k_\mathrm{ch}{\in}\{8,4,2\}$ for \textsc{Learned $1{\times}1$}, which compresses along channels.

\textbf{Reference decoder.} Table~\ref{tab:fora-full} additionally reports FORA's white-box reference decoder, trained directly on the victim's $(\mathbf{z}, \mathbf{x})$ pairs. It is non-deployable in U-shaped SL (the server never sees $\mathbf{x}$) and is included as the leakage ceiling: $\Delta_{\mathrm{Raw}}$ in Table~\ref{tab:fora-full} is computed on this column because it measures information removed from the cut rather than what the deployable surrogate happens to recover.

\textbf{Attacker seeds.} Adversarial training of the pseudo decoder fails in some runs. We train it with several seeds per cell, leave out the collapsed runs, and average the rest; the reference decoder converged in every run.

\textbf{Pseudo-head pixelwise instability.} Pseudo SSIM and PSNR are noisier than LPIPS: the surrogate-trained decoder drifts in mean RGB ($\approx 130$ on \textsc{Raw} vs.\ $\approx 115$ at ground truth), which perturbs SSIM's luminance term. LPIPS is robust to this drift and remains the primary FORA metric; Tables~\ref{tab:main-compact} and~\ref{tab:fora-privacy} report SSIM next to it as multi-seed means over the converged runs.

\subsubsection{UnSplit}
\label{app:unsplit_setup}

We adopt the UnSplit~\cite{erdogan2022unsplit} pipeline at the upstream defaults; the only edit is the insertion of the method object between the client head and the server backbone. The victim is the \texttt{MnistNet} architecture from the released attack repository, used unchanged for both datasets. We evaluate two cuts: $\ell{=}2$ (post-pool$_1$, $d{=}1152$) and $\ell{=}4$ (post-ReLU$_2$, $d{=}1024$).

\textbf{Joint training.} For each (dataset, method, $\ell$) cell we train for $20$ epochs with the upstream optimizer
(Adam~\cite{kingma2015adam}, lr $10^{-3}$, AMSGrad, batch $64$), using $\mathcal{L}_\mathrm{CE}+\lambda_\mathrm{wcc}\mathcal{L}_\mathrm{WCC}$ on the \ourname rows (Sec.~\ref{sec:method-wcc}). The compression ratios CR$\in\{8,16,32\}\times$ fix the bottleneck rank at $k\in\{144,72,36\}$ at $\ell{=}2$ and $k\in\{128,64,32\}$ at $\ell{=}4$.

\textbf{Foreground mask.} MSE and PSNR are computed only over pixels where the ground truth exceeds $0.1$; SSIM and LPIPS use a tight bounding-box crop around the same mask, so background pixels (zero in both target and reconstruction) cannot dominate the metric. Table~\ref{tab:unsplit_defenses_whole} reports the foreground ratio and the whole-image metrics side by side.

\subsubsection{Tabular probes}
\label{app:tabular_attack}
Both tabular attacks invert the $256$-d HAR cut (App.~\ref{app:setup_generality}). The reconstruction probe (\sect\ref{subsec:tabular-privacy}) trains a decoder (payload\,${\to}\,256\,{\to}\,512\,{\to}\,561$) on non-member data and evaluates it on members; features are standardized, so predicting the mean scores $R^2{=}0$. The re-identification probe, the identity-level analogue of property-inference attacks on shared representations~\cite{song2020overlearning}, trains a subject classifier on the payload (group-aware split), read against the $1/|\text{subjects}|$ chance rate. Members and non-members are stratified by activity, and a held-out member slice selects the checkpoint, so the test set never drives selection.

\subsubsection{Active attack: FSHA}
\label{app:fsha}
We run the authors' released FSHA implementation and keep its training loop, WGAN objective, and learning rates unchanged, inserting \ourname at the client-encoder output so the attack observes exactly the payload that crosses the cut; the attacker's pilot network passes through the same bottleneck, so it knows $R$. The victim is the repository's ResNet setup cut at level 4, whose activation is $4{\times}4{\times}256\,{=}\,4096$-d and therefore matches our CIFAR cut dimension; at CR\,${=}\,8\times$ the client transmits $k{=}512$ scalars. \textsc{Raw}, the learned $1{\times}1$ conv, and \topksparse keep the released convolutional decoder and discriminator, which take the $4{\times}4{\times}256$ map; the projected payload is a
$k$-vector, so for \ournameFixed the attacker's decoder and discriminator are fully-connected networks of matching input. We evaluate these defenses at this single ratio, so Table~\ref{tab:fsha_active} gives their ranking at that point rather than a trend across compression ratios; the block variant's FSHA columns (Table~\mbox{\ref{tab:block_privacy_unsplit}}) also cover $16\times$ and $32\times$. Each run trains for $10{,}000$ iterations at batch $64$, with Adam at learning rate $10^{-4}$ for the client encoder, the attacker's decoder, and the discriminator alike. Because FSHA is GAN training and its per-run reconstruction quality is noisy, we repeat every reconstruction cell over several seeds and score $20$ images per seed, reporting the mean. The DP rows follow App.~\ref{app:setup_dp_attacks}.

\subsubsection{Projection-optimized attackers}
\label{app:projopt}
We run the four attackers on the CIFAR-10 L7 cut at CR\,${\in}\,\{8,16,32\}\times$. Every attacker is a
supervised reconstruction decoder trained to convergence against a converged split model, and all
four observe the same payload $z$, differing only in the features they consume. The four decoders consume (i)~$z$ alone; (ii)~$z$ with the analytic lift-back $zR^{\top}$; (iii)~$z$ with the returned cut-layer gradient $g$; and (iv)~all three fused. The decoder that sees $z$ alone receives a wider
fully-connected stage, which matches its parameter count to the lift-back decoder, so the two decoders differ only in
the features they see, not in capacity.

\subsubsection{Differential privacy under each attack}
\label{app:setup_dp_attacks}
Every DP row in the paper uses the mechanism of App.~\ref{app:setup_dp} at its operating point: the clip is the per-configuration median transmitted-payload norm, so half of the payloads are clipped in every configuration, and the noise level is $\sigma{=}0.02$. Under SDAR the clip and the noise sit inside the method object at the cut (App.~\ref{app:sdar_setup}), so the target path and the simulator path see the same privatized payload; these runs give the SDAR DP cells of Table~\ref{tab:privacy_all} and Table~\ref{tab:dp_privacy}. For FORA, UnSplit and the HAR probes we run the sweeps of App.~\ref{app:fora_setup}--\ref{app:tabular_attack} with the mechanism added at the cut, alongside a no-DP run under the same schedule. The FSHA DP rows (App.~\ref{app:fsha}) and the block-local DP rows of Table~\ref{tab:block_privacy_unsplit} follow the same recipe.

\subsubsection{Backdoor detection}
\label{app:backdoor_setup}

We use the canonical BadNets trigger and label-flip recipe~\cite{gu2017badnets}, a $3{\times}3$ corner trigger with a fixed target class, in a multi-client SL setting on CIFAR-10. The defaults are $n{=}10$ clients, $\alpha{=}10^{7}$
(Dirichlet), malicious-client fraction $\mu{=}0.1$, per-client poison ratio $p{=}0.3$, target class $0$ and CR\,${=}\,16$, and we vary one factor at a time. The server flags clients by a cosine-to-consensus signature against a $3\sigma$-MAD threshold, the standard detector in distributed learning settings (FoolsGold~\cite{fung2020foolsgold}, FLTrust~\cite{cao2021fltrust}, CONTRA~\cite{awan2021contra}, SafeSplit~\cite{safesplit2025}). As \ourname's main focus is privacy and communication efficiency, we keep to this standard detector and do not consider more advanced attack or
defense strategies.

\subsubsection{Center loss vs.\ WCC}
\label{app:center_loss_setup}
WCC is the stateless per-batch instance of the class-compaction idea that center loss~\cite{wen2016discriminative} realizes with persistent, EMA-updated class centers (\sect\ref{sec:method-wcc}). We test whether the extra state buys privacy by running the classic center loss in place of WCC on the same SDAR setup (L7 cut, CR\,$8\times$, CIFAR-10) and comparing reconstruction.

\subsection{Reconstruction metrics}
\label{app:metrics}
We score the image reconstruction attacks with four standard image-quality metrics. \emph{Mean squared error} (MSE) is the pixel-wise mean squared error between the private image and the reconstruction, and \emph{peak signal-to-noise ratio} (PSNR) follows from it as $-10\log_{10}(\mathrm{MSE})$ for images scaled to $[0,1]$, so higher MSE or lower PSNR means a weaker reconstruction. \emph{Structural similarity} (SSIM)~\cite{1284395} compares luminance, contrast and structure between two images, so lower SSIM means less structural similarity to the private input. \emph{Learned perceptual image patch similarity} (LPIPS)~\cite{Zhang_2018_CVPR} measures perceptual distance in deep feature space, which ties it less to exact pixel alignment than MSE, so higher LPIPS means the reconstruction is less perceptually similar to the target. Throughout, arrows point toward stronger privacy: MSE/LPIPS $\uparrow$ and SSIM/PSNR $\downarrow$ indicate weaker visual leakage. Where a black background dominates the image, as in the UnSplit MNIST/Fashion-MNIST runs, we additionally report foreground-masked versions of these metrics so that matched background pixels do not hide reconstruction failures (App.~\ref{app:unsplit_setup} gives the mask). Two experiment families use their own scores. We score the language-modelling runs by validation perplexity, reported as an inflation ratio over \textsc{Raw} at the same budget (App.~\ref{app:setup_llm}), and the tabular probes by decoder $R^2$ and re-identification accuracy against the chance rate (App.~\ref{app:tabular_attack}).

\begin{table}[!htbp]
\centering
\caption{SDAR reconstruction SSIM under DP (CIFAR-10, USL) at the $4096$-d and $8192$-d cuts, lower $=$ more private; parentheses give the change relative to \textsc{Raw} at the same cut.}
\label{tab:dp_privacy}
\scaleTable{%
\begin{tabular}{l cc}
\toprule
\textbf{Configuration} & $4096$-d cut & $8192$-d cut \\
\midrule
\rowcolor{rawgray} \textsc{Raw}      & $0.572$              & $0.832$ \\
\rowcolor{rawgray} \textsc{Raw} $+$ DP            & $0.181$~{\scriptsize$(-68\%)$} & $0.486$~{\scriptsize$(-42\%)$} \\
\ournameFixed                  & $0.360$~{\scriptsize$(-37\%)$} & $0.524$~{\scriptsize$(-37\%)$} \\
\ournameFixed $+$ DP  & $0.139$~{\scriptsize$(-76\%)$} & $0.300$~{\scriptsize$(-64\%)$} \\
\bottomrule
\end{tabular}}
\end{table}

\begin{table}[!tbp]
\centering
\small
\setlength{\tabcolsep}{6pt}
\caption{Block-local projection under four reconstruction attacks over tile size $b$ and compression tier, lower $=$ more private: SDAR as SSIM in \% of the matched \textsc{Raw} at the $4096$-d and $8192$-d cuts; FORA (pseudo and reference decoder), UnSplit and FSHA as SSIM. The $+$DP rows add DP ($\sigma{=}0.02$) at $8\times$.}
\label{tab:block_privacy_unsplit}\label{tab:block_privacy_fsha}
\resizebox{\columnwidth}{!}{%
\begin{tabular}{l c cc cc cc cc}
\toprule
 & & \multicolumn{2}{c}{\textbf{SDAR} (\%\,of\,\textsc{Raw}\,$\downarrow$)} & \multicolumn{2}{c}{\textbf{FORA} (SSIM\,$\downarrow$)} & \multicolumn{2}{c}{\textbf{UnSplit} (SSIM\,$\downarrow$)} & \multicolumn{2}{c}{\textbf{FSHA} (SSIM\,$\downarrow$)} \\
\cmidrule(lr){3-4}\cmidrule(lr){5-6}\cmidrule(lr){7-8}\cmidrule(lr){9-10}
\textbf{Block} & \textbf{CR} & $4096$-d & $8192$-d & Pseudo & Ref. & MNIST & F-MNIST & MNIST & F-MNIST \\
\midrule
$b{=}1$ & $8\times$   & 56\% & 86\% & -- & 0.86 & 0.224 & 0.073 & 0.544 & 0.430 \\
        & $16\times$ & 30\% & 79\% & 0.58 & 0.78 & 0.099 & 0.036 & 0.646 & 0.562 \\
        & $32\times$ & 28\% & 74\% & 0.16 & 0.57 & 0.130 & 0.070 & 0.636 & 0.538 \\
\addlinespace[1pt]
$b{=}2$ & $8\times$   & 36\% & 83\% & 0.01 & 0.87 & 0.116 & 0.053 & 0.652 & 0.547 \\
        & $16\times$ & 48\% & 78\% & 0.27 & 0.83 & 0.073 & 0.048 & 0.662 & 0.533 \\
        & $32\times$ & 26\% & 71\% & 0.16 & 0.75 & 0.229 & 0.042 & 0.424 & 0.454 \\
\addlinespace[1pt]
$b{=}4$ & $8\times$ & 38\% & 72\% & 0.10 & 0.82 & 0.106 & 0.045 & 0.655 & 0.526 \\
        & $16\times$ & 18\% & 46\% & 0.17 & 0.76 & 0.112 & 0.045 & 0.550 & 0.492 \\
        & $32\times$ & 29\% & 53\% & 0.20 & 0.72 & 0.110 & 0.038 & 0.480 & 0.539 \\
\addlinespace[1pt]
\addlinespace[2pt]
$b{=}1$$+$DP & $8\times$ & 23\% & 69\% & -- & -- & 0.114 & 0.046 & 0.814 & 0.620 \\
$b{=}2$$+$DP & $8\times$ & 25\% & 63\% & -- & -- & 0.127 & 0.045 & 0.513 & 0.461 \\
$b{=}4$$+$DP & $8\times$ & 26\% & 46\% & -- & -- & 0.127 & 0.043 & 0.439 & 0.303 \\
\bottomrule
\end{tabular}}
\\[2pt]
\parbox{0.95\columnwidth}{\footnotesize\raggedright The global projection $b{=}8$ is the $\lambda_{\mathrm{WCC}}{=}0$ column of Table~\ref{tab:block_privacy_grid} for SDAR, the \ournameFixed rows of Table~\ref{tab:fora-full} for FORA, and the \ournameFixed rows of Table~\ref{tab:unsplit_attack_full} (split~2, $\lambda_{\mathrm{WCC}}{=}0$) for UnSplit; FSHA's $4{\times}4$ cut makes $b{=}4$ its global projection. FORA cells come from one attacker run per cell at the matched payloads; the empty pseudo cell is a run whose pseudo decoder did not converge (App.~\mbox{\ref{app:fora_setup}}), and the FORA $+$DP cells were not run. Victim accuracy stays in a narrow band in every cell (UnSplit $\ge 96.7\%$ MNIST, $\ge 84.0\%$ F-MNIST; FSHA $\ge 98.7\%$ MNIST, $\ge 90.6\%$ F-MNIST).}
\end{table}

\section{Full Results}
\label{app:results}

\subsection{Utility}
\label{app:utility}

\begin{table*}[!tbp]
\centering
\caption{U-shaped SL: best test accuracy (\%) at $\lambda_{\mathrm{WCC}}{=}0$ across the L1H and L2H cuts (SCH and PCH merged), with $\Delta$ vs.\ matched \textsc{Raw}. \textsc{\ournameLearned} uses the per-dataset width of App.~\ref{app:mlp_hidden_ablation}.}
\label{tab:full_sweep_baseline_appendix}
\scaleTable{%
\begin{tabular}{ll c c c cc cc cc cc cc cc cc cc}
\toprule
 & & & & & \multicolumn{4}{c}{\textbf{CIFAR-10}} & \multicolumn{4}{c}{\textbf{CIFAR-100}} & \multicolumn{4}{c}{\textbf{FashionMNIST}} & \multicolumn{4}{c}{\textbf{GTSRB}} \\
\cmidrule(lr){6-9} \cmidrule(lr){10-13} \cmidrule(lr){14-17} \cmidrule(lr){18-21}
 & & & & & \multicolumn{2}{c}{$\alpha{=}0.1$} & \multicolumn{2}{c}{$\alpha{\to}\infty$} & \multicolumn{2}{c}{$\alpha{=}0.5$} & \multicolumn{2}{c}{$\alpha{\to}\infty$} & \multicolumn{2}{c}{$\alpha{=}0.1$} & \multicolumn{2}{c}{$\alpha{\to}\infty$} & \multicolumn{2}{c}{$\alpha{=}0.1$} & \multicolumn{2}{c}{$\alpha{\to}\infty$} \\
\cmidrule(lr){6-7} \cmidrule(lr){8-9} \cmidrule(lr){10-11} \cmidrule(lr){12-13} \cmidrule(lr){14-15} \cmidrule(lr){16-17} \cmidrule(lr){18-19} \cmidrule(lr){20-21}
\textbf{Method} & $k$ & \textbf{CR} & \textbf{Depth} & \textbf{Head} & Acc & $\Delta$ & Acc & $\Delta$ & Acc & $\Delta$ & Acc & $\Delta$ & Acc & $\Delta$ & Acc & $\Delta$ & Acc & $\Delta$ & Acc & $\Delta$ \\
\midrule
\rowcolor{rawgray}
 \textsc{Raw} & 4096 & $1\times$ & L1H & SCH & 46.69 & -- & 86.86 & -- & 58.11 & -- & 59.56 & -- & 86.11 & -- & 91.86 & -- & 96.44 & -- & 97.61 & -- \\
\rowcolor{rawgray}
  &  &  & L1H & PCH & 22.82 & -- & 81.28 & -- & 39.78 & -- & 49.95 & -- & 57.21 & -- & 88.82 & -- & 80.80 & -- & 96.10 & -- \\
\rowcolor{rawgray}
  &  &  & L2H & SCH & 49.08 & -- & 87.24 & -- & 57.46 & -- & 59.47 & -- & 72.21 & -- & 91.64 & -- & 98.44 & -- & 98.55 & -- \\
\rowcolor{rawgray}
  &  &  & L2H & PCH & 22.67 & -- & 78.31 & -- & 31.65 & -- & 47.09 & -- & 56.03 & -- & 88.67 & -- & 79.58 & -- & 97.66 & -- \\
\midrule
 \textsc{LightSplit-F} & 512 & 8$\times$ & L1H & SCH & 41.56 & -5.13 & 76.34 & -10.52 & 44.87 & -13.24 & 43.71 & -15.85 & 82.41 & -3.70 & 90.27 & -1.59 & 87.39 & -9.06 & 90.19 & -7.42 \\
  &  &  & L1H & PCH & 21.31 & -1.51 & 71.79 & -9.49 & 23.45 & -16.33 & 38.62 & -11.33 & 52.31 & -4.90 & 86.72 & -2.10 & 56.69 & -24.11 & 88.17 & -7.93 \\
  &  &  & L2H & SCH & 45.29 & -3.79 & 83.50 & -3.74 & 53.13 & -4.33 & 52.83 & -6.64 & 74.00 & +1.79 & 91.08 & -0.56 & 95.75 & -2.69 & 96.42 & -2.13 \\
  &  &  & L2H & PCH & 20.01 & -2.66 & 72.57 & -5.74 & 24.04 & -7.61 & 37.89 & -9.20 & 51.49 & -4.54 & 87.17 & -1.50 & 64.62 & -14.96 & 94.06 & -3.59 \\
\addlinespace[1pt]
  & 256 & 16$\times$ & L1H & SCH & 36.52 & -10.17 & 72.70 & -14.16 & 42.16 & -15.95 & 42.20 & -17.36 & 79.72 & -6.39 & 89.75 & -2.11 & 84.96 & -11.49 & 87.47 & -10.14 \\
  &  &  & L1H & PCH & 17.25 & -5.57 & 67.57 & -13.71 & 22.27 & -17.51 & 35.69 & -14.26 & 51.72 & -5.49 & 86.13 & -2.69 & 42.89 & -37.91 & 85.76 & -10.35 \\
  &  &  & L2H & SCH & 43.86 & -5.22 & 81.99 & -5.25 & 52.95 & -4.51 & 51.85 & -7.62 & 72.59 & +0.38 & 90.83 & -0.81 & 94.05 & -4.39 & 94.77 & -3.78 \\
  &  &  & L2H & PCH & 19.78 & -2.89 & 70.76 & -7.55 & 23.18 & -8.47 & 37.62 & -9.47 & 51.79 & -4.24 & 86.38 & -2.29 & 54.10 & -25.48 & 91.69 & -5.97 \\
\addlinespace[1pt]
  & 128 & 32$\times$ & L1H & SCH & 34.02 & -12.67 & 69.97 & -16.89 & 38.59 & -19.52 & 39.04 & -20.52 & 76.75 & -9.36 & 89.26 & -2.60 & 77.59 & -18.85 & 83.19 & -14.42 \\
  &  &  & L1H & PCH & 17.82 & -5.00 & 63.92 & -17.36 & 18.11 & -21.67 & 32.97 & -16.98 & 51.21 & -6.00 & 85.68 & -3.14 & 39.77 & -41.03 & 77.42 & -18.69 \\
  &  &  & L2H & SCH & 42.59 & -6.49 & 82.03 & -5.21 & 50.85 & -6.61 & 51.31 & -8.16 & 70.78 & -1.43 & 90.00 & -1.64 & 89.81 & -8.63 & 93.07 & -5.48 \\
  &  &  & L2H & PCH & 19.88 & -2.79 & 70.13 & -8.18 & 18.25 & -13.40 & 36.37 & -10.72 & 49.96 & -6.07 & 86.20 & -2.47 & 46.65 & -32.93 & 88.60 & -9.06 \\
\midrule
 \textsc{LightSplit-L} & 512 & 8$\times$ & L1H & SCH & 43.56 & -3.13 & 76.72 & -10.14 & 48.05 & -10.06 & 45.89 & -13.67 & 69.18 & -16.93 & 89.97 & -1.89 & 82.21 & -14.24 & 89.47 & -8.14 \\
  &  &  & L1H & PCH & 24.26 & +1.44 & 72.58 & -8.70 & 26.68 & -13.10 & 41.90 & -8.05 & 45.73 & -11.48 & 84.69 & -4.13 & 46.83 & -33.97 & 87.08 & -9.03 \\
  &  &  & L2H & SCH & 44.52 & -4.56 & 83.42 & -3.82 & 54.77 & -2.69 & 56.49 & -2.98 & 70.87 & -1.34 & 91.18 & -0.46 & 93.20 & -5.24 & 95.71 & -2.84 \\
  &  &  & L2H & PCH & 22.00 & -0.67 & 70.93 & -7.38 & 26.96 & -4.69 & 38.33 & -8.76 & 47.59 & -8.44 & 85.38 & -3.29 & 61.76 & -17.82 & 94.87 & -2.79 \\
\addlinespace[1pt]
  & 256 & 16$\times$ & L1H & SCH & 47.50 & +0.81 & 75.00 & -11.86 & 45.68 & -12.43 & 43.18 & -16.38 & 65.43 & -20.68 & 89.29 & -2.57 & 79.14 & -17.30 & 87.14 & -10.47 \\
  &  &  & L1H & PCH & 23.13 & +0.31 & 70.52 & -10.76 & 26.00 & -13.78 & 38.59 & -11.36 & 37.29 & -19.92 & 83.41 & -5.41 & 35.19 & -45.61 & 86.54 & -9.56 \\
  &  &  & L2H & SCH & 45.38 & -3.70 & 84.45 & -2.79 & 52.68 & -4.78 & 54.60 & -4.87 & 70.64 & -1.57 & 91.04 & -0.60 & 89.52 & -8.92 & 94.98 & -3.57 \\
  &  &  & L2H & PCH & 17.85 & -4.82 & 69.95 & -8.36 & 26.21 & -5.44 & 37.12 & -9.97 & 29.29 & -26.74 & 84.71 & -3.96 & 57.09 & -22.49 & 91.70 & -5.95 \\
\addlinespace[1pt]
  & 128 & 32$\times$ & L1H & SCH & 42.79 & -3.90 & 71.84 & -15.02 & 42.50 & -15.61 & 39.50 & -20.06 & 61.17 & -24.94 & 88.77 & -3.09 & 73.18 & -23.26 & 83.58 & -14.03 \\
  &  &  & L1H & PCH & 20.31 & -2.51 & 66.35 & -14.93 & 23.29 & -16.49 & 34.92 & -15.03 & 23.97 & -33.24 & 81.12 & -7.70 & 30.80 & -50.00 & 78.04 & -18.07 \\
  &  &  & L2H & SCH & 45.50 & -3.58 & 82.65 & -4.59 & 52.74 & -4.72 & 52.54 & -6.93 & 69.00 & -3.21 & 90.45 & -1.19 & 87.05 & -11.39 & 92.80 & -5.75 \\
  &  &  & L2H & PCH & 17.62 & -5.05 & 68.45 & -9.86 & 24.79 & -6.86 & 34.94 & -12.15 & 36.03 & -20.00 & 83.83 & -4.84 & 49.64 & -29.94 & 90.21 & -7.44 \\
\midrule
 \textsc{Learned $1{\times}1$} & 512 & 8$\times$ & L1H & SCH & 47.72 & +1.03 & 85.54 & -1.32 & 56.67 & -1.44 & 56.81 & -2.75 & 82.25$^{\dagger}$ & -3.86 & 90.65$^{\dagger}$ & -1.21 & 96.61$^{\dagger}$ & +0.17 & 97.39$^{\dagger}$ & -0.22 \\
  &  &  & L1H & PCH & 15.87 & -6.95 & 83.04 & +1.76 & 43.10 & +3.32 & 49.97 & +0.02 & 67.80$^{\dagger}$ & +10.59 & 88.85$^{\dagger}$ & +0.03 & 82.79$^{\dagger}$ & +2.00 & 95.09$^{\dagger}$ & -1.01 \\
  &  &  & L2H & SCH & 40.40 & -8.68 & 86.65 & -0.59 & 56.01 & -1.45 & 57.04 & -2.43 & 75.47$^{\dagger}$ & +3.26 & 91.20$^{\dagger}$ & -0.44 & 97.60$^{\dagger}$ & -0.84 & 98.01$^{\dagger}$ & -0.54 \\
  &  &  & L2H & PCH & 16.52 & -6.15 & 80.09 & +1.78 & 33.84 & +2.19 & 46.78 & -0.31 & 60.42$^{\dagger}$ & +4.39 & 89.05$^{\dagger}$ & +0.38 & 84.45$^{\dagger}$ & +4.87 & 97.35$^{\dagger}$ & -0.31 \\
\addlinespace[1pt]
  & 256 & 16$\times$ & L1H & SCH & 39.13 & -7.56 & 83.32 & -3.54 & 50.68 & -7.43 & 50.45 & -9.11 & 79.42$^{\dagger}$ & -6.69 & 89.25$^{\dagger}$ & -2.61 & 95.00$^{\dagger}$ & -1.45 & 95.87$^{\dagger}$ & -1.74 \\
  &  &  & L1H & PCH & 14.87 & -7.95 & 80.17 & -1.11 & 42.16 & +2.38 & 47.88 & -2.07 & 68.34$^{\dagger}$ & +11.13 & 88.44$^{\dagger}$ & -0.38 & 82.79$^{\dagger}$ & +1.99 & 94.51$^{\dagger}$ & -1.59 \\
  &  &  & L2H & SCH & 39.14 & -9.94 & 85.74 & -1.50 & 53.67 & -3.79 & 54.99 & -4.48 & 74.93$^{\dagger}$ & +2.72 & 90.82$^{\dagger}$ & -0.82 & 96.30$^{\dagger}$ & -2.14 & 97.55$^{\dagger}$ & -1.01 \\
  &  &  & L2H & PCH & 23.69 & +1.02 & 78.45 & +0.14 & 31.50 & -0.15 & 45.86 & -1.23 & 59.97$^{\dagger}$ & +3.94 & 88.57$^{\dagger}$ & -0.10 & 84.13$^{\dagger}$ & +4.55 & 96.61$^{\dagger}$ & -1.05 \\
\addlinespace[1pt]
  & 128 & 32$\times$ & L1H & SCH & 38.42 & -8.27 & 75.99 & -10.87 & 41.75 & -16.36 & 41.66 & -17.90 & 82.88$^{\dagger}$ & -3.23 & 89.26$^{\dagger}$ & -2.60 & 94.65$^{\dagger}$ & -1.80 & 95.87$^{\dagger}$ & -1.74 \\
  &  &  & L1H & PCH & 13.24 & -9.58 & 73.90 & -7.38 & 33.99 & -5.79 & 41.02 & -8.93 & 67.09$^{\dagger}$ & +9.88 & 88.38$^{\dagger}$ & -0.44 & 85.26$^{\dagger}$ & +4.46 & 94.33$^{\dagger}$ & -1.77 \\
  &  &  & L2H & SCH & 36.56 & -12.52 & 83.90 & -3.34 & 47.82 & -9.64 & 49.40 & -10.07 & 75.45$^{\dagger}$ & +3.24 & 90.75$^{\dagger}$ & -0.89 & 96.56$^{\dagger}$ & -1.88 & 97.59$^{\dagger}$ & -0.96 \\
  &  &  & L2H & PCH & 16.59 & -6.08 & 74.31 & -4.00 & 27.86 & -3.79 & 40.86 & -6.23 & 60.16$^{\dagger}$ & +4.13 & 88.76$^{\dagger}$ & +0.09 & 84.83$^{\dagger}$ & +5.25 & 96.33$^{\dagger}$ & -1.33 \\
\midrule
\rowcolor{rawgray}
 \topksparse & 512 & 8$\times$ & L1H & SCH & 39.60 & -7.09 & 81.41 & -5.45 & 50.35 & -7.76 & 51.84 & -7.72 & 88.57 & +2.46 & 90.32 & -1.54 & 92.95 & -3.49 & 84.59 & -13.02 \\
\rowcolor{rawgray}
  &  &  & L1H & PCH & 20.27 & -2.55 & 70.27 & -11.01 & 24.73 & -15.05 & 37.08 & -12.87 & 49.33 & -7.88 & 86.49 & -2.33 & 37.49 & -43.31 & 77.98 & -18.12 \\
\rowcolor{rawgray}
  &  &  & L2H & SCH & 42.73 & -6.35 & 85.66 & -1.58 & 54.48 & -2.98 & 57.19 & -2.28 & 69.34 & -2.87 & 91.19 & -0.45 & 97.16 & -1.28 & 97.48 & -1.07 \\
\rowcolor{rawgray}
  &  &  & L2H & PCH & 25.35 & +2.68 & 71.55 & -6.76 & 25.10 & -6.55 & 37.59 & -9.50 & 48.18 & -7.85 & 86.83 & -1.84 & 52.07 & -27.51 & 88.42 & -9.24 \\
\addlinespace[1pt]
\rowcolor{rawgray}
  & 256 & 16$\times$ & L1H & SCH & 34.66 & -12.03 & 76.85 & -10.01 & 44.93 & -13.18 & 46.46 & -13.10 & 87.69 & +1.58 & 89.52 & -2.34 & 84.62 & -11.82 & 82.20 & -15.41 \\
\rowcolor{rawgray}
  &  &  & L1H & PCH & 17.45 & -5.37 & 65.01 & -16.27 & 17.30 & -22.48 & 30.13 & -19.82 & 51.73 & -5.48 & 84.85 & -3.97 & 26.66 & -54.14 & 63.65 & -32.45 \\
\rowcolor{rawgray}
  &  &  & L2H & SCH & 40.84 & -8.24 & 84.59 & -2.65 & 53.54 & -3.92 & 56.02 & -3.45 & 74.10 & +1.89 & 90.60 & -1.04 & 95.87 & -2.57 & 96.36 & -2.19 \\
\rowcolor{rawgray}
  &  &  & L2H & PCH & 21.05 & -1.62 & 67.20 & -11.11 & 21.96 & -9.69 & 34.98 & -12.11 & 52.14 & -3.89 & 86.82 & -1.85 & 35.06 & -44.52 & 80.59 & -17.07 \\
\addlinespace[1pt]
\rowcolor{rawgray}
  & 128 & 32$\times$ & L1H & SCH & 30.75 & -15.94 & 70.84 & -16.02 & 37.77 & -20.34 & 40.87 & -18.69 & 86.94 & +0.83 & 87.26 & -4.60 & 66.83 & -29.61 & 78.51 & -19.10 \\
\rowcolor{rawgray}
  &  &  & L1H & PCH & 17.10 & -5.72 & 52.94 & -28.34 & 14.45 & -25.33 & 24.74 & -25.21 & 49.14 & -8.07 & 82.88 & -5.94 & 26.22 & -54.58 & 61.72 & -34.38 \\
\rowcolor{rawgray}
  &  &  & L2H & SCH & 39.13 & -9.95 & 83.03 & -4.21 & 51.13 & -6.33 & 52.99 & -6.48 & 67.67 & -4.54 & 89.79 & -1.85 & 91.00 & -7.44 & 93.06 & -5.49 \\
\rowcolor{rawgray}
  &  &  & L2H & PCH & 17.77 & -4.90 & 62.98 & -15.33 & 18.07 & -13.58 & 26.76 & -20.33 & 49.09 & -6.94 & 85.21 & -3.46 & 31.58 & -48.00 & 76.10 & -21.56 \\
\bottomrule
\end{tabular}}\\[2pt]
\parbox{0.95\textwidth}{\footnotesize\raggedright
Shaded: \textsc{Raw} and the \topksparse baseline.\\
$^{\dagger}$The learned $1{\times}1$ keeps whole channels. At $8\times$ it keeps $2$ of $20$ channels on FashionMNIST and $3$ of $29$ on GTSRB (realised CR ${\approx}10\times$); at $16\times$ and $32\times$ it keeps one channel on both (realised CR $20\times$ and $29\times$), so those two rows train the same one-channel model twice.}
\end{table*}

\begin{table*}[!tbp]
\centering
\scriptsize
\setlength{\tabcolsep}{3pt}
\caption{Block-local utility over tile size $b$: $\Delta$ accuracy (points) vs.\ matched \textsc{Raw}; each cell pairs the transpose and the learned MLP lift-back (\ournameFixed\,/\,\ournameLearned).}
\label{tab:block_util_F}\label{tab:block_util_L}
\scaleTable{%
\begin{tabular}{l c c c cc cc cc cc}
\toprule
 & & & & \multicolumn{2}{c}{\textbf{CIFAR-10}} & \multicolumn{2}{c}{\textbf{CIFAR-100}} & \multicolumn{2}{c}{\textbf{FashionMNIST}} & \multicolumn{2}{c}{\textbf{GTSRB}} \\
\cmidrule(lr){5-6} \cmidrule(lr){7-8} \cmidrule(lr){9-10} \cmidrule(lr){11-12}
\textbf{Block} & \textbf{CR} & \textbf{Depth} & \textbf{Head} & $\alpha{=}0.1$ & $\alpha{\to}\infty$ & $\alpha{=}0.5$ & $\alpha{\to}\infty$ & $\alpha{=}0.1$ & $\alpha{\to}\infty$ & $\alpha{=}0.1$ & $\alpha{\to}\infty$ \\
\midrule
\rowcolor{rawgray} \textsc{Raw} (abs.) & $1\times$ & L1H & SCH & 46.69 & 86.86 & 58.11 & 59.56 & 86.11 & 91.86 & 96.44 & 97.61 \\
\rowcolor{rawgray}  &  & L1H & PCH & 22.82 & 81.28 & 39.78 & 49.95 & 57.21 & 88.82 & 80.80 & 96.10 \\
\rowcolor{rawgray}  &  & L2H & SCH & 49.08 & 87.24 & 57.46 & 59.47 & 72.21 & 91.64 & 98.44 & 98.55 \\
\rowcolor{rawgray}  &  & L2H & PCH & 22.67 & 78.31 & 31.65 & 47.09 & 56.03 & 88.67 & 79.58 & 97.66 \\
\midrule
$b{=}1$ & 8$\times$ & L1H & SCH & $-2.23$\,/\,$-4.33$ & $-2.04$\,/\,$-8.46$ & $-11.31$\,/\,$-11.23$ & $-6.86$\,/\,$-14.09$ & $-7.08$\,/\,$-24.91^{\dagger}$ & $-1.16$\,/\,$-1.88^{\dagger}$ & $-0.51$\,/\,$-11.02^{\dagger}$ & $-1.19$\,/\,$-6.53^{\dagger}$ \\
 &  & L1H & PCH & $-3.51$\,/\,$-2.68$ & $+1.24$\,/\,$-9.79$ & $-1.01$\,/\,$-12.34$ & $-1.52$\,/\,$-10.87$ & $+3.64$\,/\,$-25.70^{\dagger}$ & $+0.29$\,/\,$-2.00^{\dagger}$ & $-9.47$\,/\,$-36.54^{\dagger}$ & $-1.91$\,/\,$-7.30^{\dagger}$ \\
 &  & L2H & SCH & $-6.34$\,/\,$-2.17$ & $-1.76$\,/\,$-3.46$ & $-10.65$\,/\,$-11.93$ & $-5.13$\,/\,$-6.23$ & $+0.70$\,/\,$-4.45^{\dagger}$ & $-0.39$\,/\,$-0.55^{\dagger}$ & $-0.36$\,/\,$-7.26^{\dagger}$ & $-0.44$\,/\,$-2.49^{\dagger}$ \\
 &  & L2H & PCH & $-2.02$\,/\,$-0.25$ & $-0.58$\,/\,$-7.10$ & $-0.75$\,/\,$-6.28$ & $-1.96$\,/\,$-9.52$ & $+7.07$\,/\,$-20.62^{\dagger}$ & $+0.25$\,/\,$-0.93^{\dagger}$ & $-12.92$\,/\,$-15.28^{\dagger}$ & $-0.80$\,/\,$-5.52^{\dagger}$ \\
\addlinespace[1pt]
 & 16$\times$ & L1H & SCH & $-5.62$\,/\,$-9.27$ & $-3.89$\,/\,$-10.95$ & $-16.98$\,/\,$-22.19$ & $-11.95$\,/\,$-20.03$ & $-7.72$\,/\,$-21.91^{\dagger}$ & $-1.82$\,/\,$-2.57^{\dagger}$ & $-3.73$\,/\,$-13.87^{\dagger}$ & $-2.49$\,/\,$-6.82^{\dagger}$ \\
 &  & L1H & PCH & $-8.18$\,/\,$-1.33$ & $-1.07$\,/\,$-9.69$ & $-4.25$\,/\,$-13.20$ & $-5.54$\,/\,$-15.61$ & $+6.27$\,/\,$-32.34^{\dagger}$ & $-0.81$\,/\,$-11.33^{\dagger}$ & $-30.60$\,/\,$-27.58^{\dagger}$ & $-2.95$\,/\,$-7.64^{\dagger}$ \\
 &  & L2H & SCH & $-3.86$\,/\,$-4.39$ & $-2.22$\,/\,$-4.14$ & $-14.36$\,/\,$-12.17$ & $-8.92$\,/\,$-7.00$ & $-0.73$\,/\,$-7.53^{\dagger}$ & $-0.51$\,/\,$-1.17^{\dagger}$ & $-2.92$\,/\,$-11.61^{\dagger}$ & $-1.64$\,/\,$-5.46^{\dagger}$ \\
 &  & L2H & PCH & $-3.35$\,/\,$-2.66$ & $-1.44$\,/\,$-8.76$ & $-1.48$\,/\,$-8.23$ & $-3.18$\,/\,$-11.45$ & $+5.37$\,/\,$-18.75^{\dagger}$ & $+0.59$\,/\,$-1.40^{\dagger}$ & $-11.80$\,/\,$-15.01^{\dagger}$ & $-1.07$\,/\,$-6.14^{\dagger}$ \\
\addlinespace[1pt]
 & 32$\times$ & L1H & SCH & $-11.80$\,/\,$-14.08$ & $-11.16$\,/\,$-16.81$ & $-24.82$\,/\,$-30.59$ & $-18.59$\,/\,$-24.58$ & $-7.28$\,/\,$-22.74^{\dagger}$ & $-2.34$\,/\,$-2.30^{\dagger}$ & $-4.25$\,/\,$-15.26^{\dagger}$ & $-2.40$\,/\,$-6.86^{\dagger}$ \\
 &  & L1H & PCH & $-10.57$\,/\,$-9.39$ & $-8.18$\,/\,$-20.49$ & $-8.59$\,/\,$-18.85$ & $-10.39$\,/\,$-19.93$ & $+7.67$\,/\,$-22.21^{\dagger}$ & $+0.20$\,/\,$-1.97^{\dagger}$ & $-13.78$\,/\,$-27.80^{\dagger}$ & $-1.74$\,/\,$-7.90^{\dagger}$ \\
 &  & L2H & SCH & $-11.12$\,/\,$-5.83$ & $-4.36$\,/\,$-4.79$ & $-17.60$\,/\,$-15.81$ & $-12.00$\,/\,$-9.95$ & $-4.06$\,/\,$-5.08^{\dagger}$ & $-0.45$\,/\,$-0.86^{\dagger}$ & $-2.79$\,/\,$-16.67^{\dagger}$ & $-1.34$\,/\,$-6.08^{\dagger}$ \\
 &  & L2H & PCH & $-3.37$\,/\,$-9.51$ & $-5.79$\,/\,$-12.23$ & $-6.44$\,/\,$-9.48$ & $-7.84$\,/\,$-12.91$ & $+7.77$\,/\,$-15.14^{\dagger}$ & $+0.57$\,/\,$-2.20^{\dagger}$ & $-1.63$\,/\,$-28.33^{\dagger}$ & $-1.47$\,/\,$-7.40^{\dagger}$ \\
\midrule
$b{=}2$ & 8$\times$ & L1H & SCH & $-4.85$\,/\,$-3.59$ & $-4.74$\,/\,$-8.51$ & $-14.54$\,/\,$-18.91$ & $-8.59$\,/\,$-11.77$ & $-4.46$\,/\,$-19.97$ & $-1.09$\,/\,$-1.64$ & $-1.35$\,/\,$-15.28$ & $-1.65$\,/\,$-6.56$ \\
 &  & L1H & PCH & $-8.91$\,/\,$-3.90$ & $-3.08$\,/\,$-9.58$ & $-6.37$\,/\,$-12.82$ & $-5.06$\,/\,$-10.26$ & $-1.07$\,/\,$-22.31$ & $-0.89$\,/\,$-3.49$ & $-37.59$\,/\,$-37.13$ & $-3.38$\,/\,$-9.50$ \\
 &  & L2H & SCH & $-5.16$\,/\,$-4.40$ & $-2.11$\,/\,$-3.38$ & $-11.85$\,/\,$-10.33$ & $-6.12$\,/\,$-4.11$ & $+0.39$\,/\,$-0.21$ & $-0.21$\,/\,$-0.33$ & $-1.31$\,/\,$-5.40$ & $-0.91$\,/\,$-2.73$ \\
 &  & L2H & PCH & $+2.35$\,/\,$+6.58$ & $-2.67$\,/\,$-7.05$ & $+0.62$\,/\,$-3.05$ & $-3.62$\,/\,$-7.60$ & $+1.75$\,/\,$-12.26$ & $-0.19$\,/\,$-2.07$ & $-13.63$\,/\,$-31.00$ & $-1.38$\,/\,$-6.18$ \\
\addlinespace[1pt]
 & 16$\times$ & L1H & SCH & $-4.90$\,/\,$-3.68$ & $-5.45$\,/\,$-9.46$ & $-15.50$\,/\,$-19.52$ & $-10.23$\,/\,$-13.15$ & $-4.38$\,/\,$-17.77$ & $-1.41$\,/\,$-1.83$ & $-3.61$\,/\,$-16.08^{\dagger}$ & $-2.07$\,/\,$-8.28^{\dagger}$ \\
 &  & L1H & PCH & $+1.59$\,/\,$+2.27$ & $-3.82$\,/\,$-11.53$ & $-5.02$\,/\,$-12.65$ & $-6.22$\,/\,$-10.50$ & $+2.54$\,/\,$-27.09$ & $-0.75$\,/\,$-4.24$ & $-8.87$\,/\,$-30.05^{\dagger}$ & $-3.65$\,/\,$-10.66^{\dagger}$ \\
 &  & L2H & SCH & $-6.74$\,/\,$-3.63$ & $-2.80$\,/\,$-3.83$ & $-12.32$\,/\,$-11.20$ & $-7.06$\,/\,$-5.64$ & $-2.08$\,/\,$-3.32$ & $-0.29$\,/\,$-0.68$ & $-1.60$\,/\,$-10.09^{\dagger}$ & $-0.99$\,/\,$-3.65^{\dagger}$ \\
 &  & L2H & PCH & $-8.93$\,/\,$-6.97$ & $-2.53$\,/\,$-6.37$ & $-0.72$\,/\,$-4.29$ & $-4.37$\,/\,$-7.34$ & $+1.17$\,/\,$-11.49$ & $+0.36$\,/\,$-2.19$ & $-19.91$\,/\,$-24.12^{\dagger}$ & $-2.26$\,/\,$-6.56^{\dagger}$ \\
\addlinespace[1pt]
 & 32$\times$ & L1H & SCH & $-8.49$\,/\,$-7.84$ & $-7.93$\,/\,$-11.18$ & $-18.97$\,/\,$-21.43$ & $-13.72$\,/\,$-17.16$ & $-8.63$\,/\,$-23.60^{\dagger}$ & $-2.01$\,/\,$-2.73^{\dagger}$ & $-6.04$\,/\,$-15.25^{\dagger}$ & $-3.32$\,/\,$-7.42^{\dagger}$ \\
 &  & L1H & PCH & $-5.31$\,/\,$-4.97$ & $-4.06$\,/\,$-10.03$ & $-9.72$\,/\,$-14.14$ & $-8.78$\,/\,$-12.36$ & $+1.67$\,/\,$-23.57^{\dagger}$ & $-0.54$\,/\,$-4.34^{\dagger}$ & $-8.24$\,/\,$-36.73^{\dagger}$ & $-3.97$\,/\,$-9.63^{\dagger}$ \\
 &  & L2H & SCH & $-5.70$\,/\,$-5.53$ & $-3.41$\,/\,$-3.82$ & $-14.23$\,/\,$-14.21$ & $-8.73$\,/\,$-7.44$ & $+0.16$\,/\,$-9.86^{\dagger}$ & $-0.77$\,/\,$-0.76^{\dagger}$ & $-2.64$\,/\,$-7.84^{\dagger}$ & $-1.98$\,/\,$-4.84^{\dagger}$ \\
 &  & L2H & PCH & $-3.96$\,/\,$-6.94$ & $-2.91$\,/\,$-7.06$ & $+0.29$\,/\,$-4.32$ & $-5.59$\,/\,$-7.67$ & $-0.36$\,/\,$-21.77^{\dagger}$ & $-0.62$\,/\,$-1.92^{\dagger}$ & $-24.07$\,/\,$-28.96^{\dagger}$ & $-2.79$\,/\,$-6.39^{\dagger}$ \\
\midrule
$b{=}4$ & 8$\times$ & L1H & SCH & $-7.58$\,/\,$-7.60$ & $-6.88$\,/\,$-9.61$ & $-19.85$\,/\,$-19.08$ & $-14.38$\,/\,$-13.40$ & $-5.77$\,/\,$-22.69$ & $-1.10$\,/\,$-1.90$ & \textsc{n/a} & \textsc{n/a} \\
 &  & L1H & PCH & $-1.53$\,/\,$+0.59$ & $-3.21$\,/\,$-10.81$ & $-14.90$\,/\,$-12.53$ & $-9.04$\,/\,$-10.63$ & $-1.92$\,/\,$-20.75$ & $-1.05$\,/\,$-4.44$ & \textsc{n/a} & \textsc{n/a} \\
 &  & L2H & SCH & $-5.57$\,/\,$-2.57$ & $-3.68$\,/\,$-3.72$ & $-14.22$\,/\,$-12.24$ & $-8.95$\,/\,$-5.36$ & $-2.77$\,/\,$-2.19$ & $-0.34$\,/\,$-0.38$ & \textsc{n/a} & \textsc{n/a} \\
 &  & L2H & PCH & $+4.32$\,/\,$-5.73$ & $-5.84$\,/\,$-7.72$ & $-5.74$\,/\,$-4.55$ & $-6.86$\,/\,$-8.12$ & $-2.32$\,/\,$-18.99$ & $-1.23$\,/\,$-3.15$ & \textsc{n/a} & \textsc{n/a} \\
\addlinespace[1pt]
 & 16$\times$ & L1H & SCH & $-11.18$\,/\,$-7.52$ & $-12.18$\,/\,$-11.76$ & $-22.36$\,/\,$-20.64$ & $-16.01$\,/\,$-15.70$ & $-6.27$\,/\,$-16.56$ & $-1.45$\,/\,$-2.49$ & \textsc{n/a} & \textsc{n/a} \\
 &  & L1H & PCH & $-0.58$\,/\,$-3.52$ & $-7.74$\,/\,$-11.59$ & $-15.25$\,/\,$-13.75$ & $-10.86$\,/\,$-12.19$ & $-5.85$\,/\,$-26.23$ & $-1.87$\,/\,$-5.55$ & \textsc{n/a} & \textsc{n/a} \\
 &  & L2H & SCH & $-3.21$\,/\,$-4.01$ & $-3.87$\,/\,$-4.20$ & $-14.28$\,/\,$-12.44$ & $-9.19$\,/\,$-5.87$ & $-0.14$\,/\,$-0.47$ & $-0.50$\,/\,$-0.89$ & \textsc{n/a} & \textsc{n/a} \\
 &  & L2H & PCH & $-6.01$\,/\,$+0.90$ & $-6.36$\,/\,$-7.85$ & $-9.68$\,/\,$-4.56$ & $-7.90$\,/\,$-8.98$ & $-4.84$\,/\,$-15.70$ & $-0.82$\,/\,$-3.35$ & \textsc{n/a} & \textsc{n/a} \\
\addlinespace[1pt]
 & 32$\times$ & L1H & SCH & $-12.28$\,/\,$-9.73$ & $-14.49$\,/\,$-14.99$ & $-25.20$\,/\,$-23.78$ & $-18.53$\,/\,$-18.56$ & $-9.65$\,/\,$-22.69$ & $-2.11$\,/\,$-2.63$ & \textsc{n/a} & \textsc{n/a} \\
 &  & L1H & PCH & $-1.81$\,/\,$-4.47$ & $-11.56$\,/\,$-13.56$ & $-17.75$\,/\,$-17.52$ & $-14.46$\,/\,$-13.63$ & $-7.03$\,/\,$-28.36$ & $-1.97$\,/\,$-4.66$ & \textsc{n/a} & \textsc{n/a} \\
 &  & L2H & SCH & $-8.89$\,/\,$-5.99$ & $-4.35$\,/\,$-4.30$ & $-16.09$\,/\,$-14.45$ & $-11.57$\,/\,$-7.23$ & $-1.39$\,/\,$-5.16$ & $-0.69$\,/\,$-0.98$ & \textsc{n/a} & \textsc{n/a} \\
 &  & L2H & PCH & $+0.92$\,/\,$-4.37$ & $-7.12$\,/\,$-8.03$ & $-9.62$\,/\,$-5.38$ & $-9.34$\,/\,$-9.35$ & $-5.29$\,/\,$-11.61$ & $-1.09$\,/\,$-2.89$ & \textsc{n/a} & \textsc{n/a} \\
\bottomrule
\end{tabular}%
}\\[2pt]
\parbox{0.95\textwidth}{\footnotesize\raggedright
$^{\dagger}$Both numbers of the cell: the realised CR differs from the requested one because the tile payload cannot shrink further, which binds at lower CR for smaller tiles. \textsc{n/a}: the tile does not divide the activation.}
\end{table*}

\subsubsection{Full utility baseline experiments}
\label{app:full_sweep}

Table~\ref{tab:full_sweep_baseline_appendix} reports the complete utility grid behind the main-text utility analysis: best test accuracy for every (dataset, distribution, head ownership, head depth, CR) cell, with $\Delta$ measured against the matched \textsc{Raw} baseline at the same head depth, head ownership, and dataset. The four head configurations are the cross of L1H/L2H with SCH/PCH, so each row group corresponds to one system configuration.

\subsubsection{Block-local projection}
\label{app:block}

In Table~\ref{tab:block_util_F}, averaged over all four datasets, both cut depths, both head-ownership settings and the three ratios, the gap of \ournameFixed to \textsc{Raw} is $4.5$ points at $b{=}1$, $5.0$ at $b{=}2$, $7.4$ at $b{=}4$ and $9.2$ for the global projection $b{=}8$ (the \ournameFixed rows of Table~\ref{tab:full_sweep_baseline_appendix}), at matched bytes. A small tile keeps the coarse spatial grid the backbone reads, so the projection removes detail within each patch but not the layout across patches. The gain concentrates in the per-client-head cells, where the gap shrinks from $10.8$ points at $b{=}8$ to $3.1$ at $b{=}1$ and $4.7$ at $b{=}2$, because each client's head drift stays inside its own tiles instead of spreading over the whole lifted input; in the shared-head cells, where there is no drift to contain, it moves only from $7.6$ to $5.9$ and $5.3$. The learned lift-back (\ournameLearned) costs more at every tile size, $11.1$, $10.0$ and $9.4$ points at $b{=}1$, $2$ and $4$, and does not gain from small tiles: its dense MLP maps the whole payload back to the whole activation, so it re-mixes across tiles and gives up the locality the projection preserved.

\textbf{WCC on block-local cuts.} The $b{=}2$ rows of Table~\ref{tab:wcc_ablation} show that block locality helps most where $\lambda_{\mathrm{WCC}}$ is small: at $\lambda_{\mathrm{WCC}}{=}0$ the $b{=}2$ tile improves every \ournameFixed configuration, turning the L2H-PCH loss of $-9.1\%$ into $+0.2\%$ and the L1H-SCH loss of $-9.5\%$ into $-3.8\%$. The advantage reverses as $\lambda_{\mathrm{WCC}}$ grows, and reverses hardest on a per-client head: at $\lambda_{\mathrm{WCC}}{=}10^{-2}$ the L2H-PCH tile falls to $-58.2\%$ where the global projection holds $-33.0\%$. A smaller tile leaves WCC a smaller subspace to whiten, so the same $\lambda_{\mathrm{WCC}}$ costs more accuracy. With a shared head the two knobs remain a genuine choice; with a per-client head, block locality calls for a small or zero $\lambda_{\mathrm{WCC}}$. \ournameLearned follows the same pattern with a steeper penalty, and on L1H-PCH the tile already costs more at $\lambda_{\mathrm{WCC}}{=}0$.

\subsubsection{ImageNet-1k}
\label{app:imagenet}
\begin{figure}[!htbp]
    \centering
    \includegraphics[width=\columnwidth]{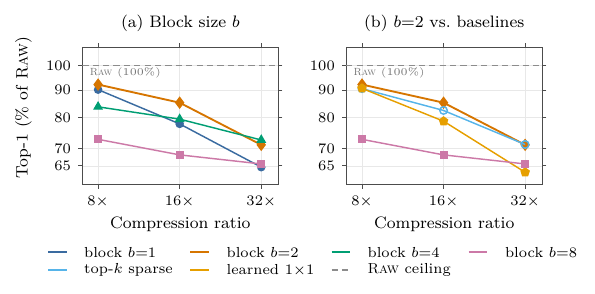}
    \caption{\ourname on ImageNet-1k ($1000$ classes, ResNet, $8{\times}8{\times}64$ cut, shared deep head, IID.}
    \label{fig:imagenet_generality}
\end{figure}

Fig.~\ref{fig:imagenet_generality} reports top-1 accuracy on ImageNet-1k at $8\times$, $16\times$ and $32\times$: panel~(a) sweeps the tile size $b\,{\in}\,\{1,2,4,8\}$, panel~(b) compares the $b{=}2$ operating point of \sect\ref{subsec:imagenet} with the matched-budget baselines (setup in App.~\ref{app:setup_imagenet}).

\subsubsection{Tabular (HAR): full compression sweeps}
\label{app:har}
\begin{figure}[!htbp]
    \centering
    \includegraphics[width=0.95\columnwidth]{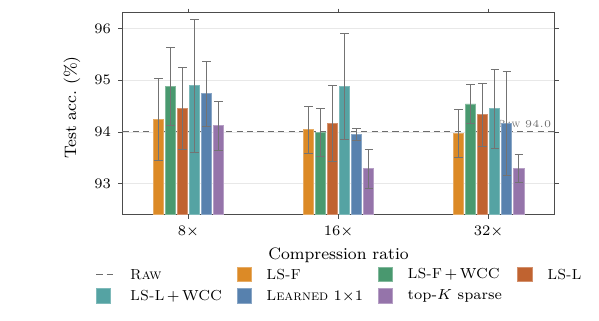}
    \caption{Utility on tabular data (UCI HAR) over the compression ratio; the dashed line is \textsc{Raw}, error bars span the seeds.}
    \label{fig:har_generality}
\end{figure}

In Fig.~\ref{fig:har_generality} every \ourname variant, with and without WCC, stays within one point of the $94.0\%$ \textsc{Raw} accuracy at all three ratios, with error bars that overlap the \textsc{Raw} line; \topksparse is the only method clearly below \textsc{Raw} at $16\times$ and $32\times$. The $256$-d cut carries more than the six-class task needs, so a rank-$k$ view keeps the task signal at every ratio; the learned lift-back and WCC add nothing the MLP backbone cannot already recover, while \topksparse drops coordinates outright instead of mixing them and loses the signal they carried. Over heterogeneous clients (Table~\ref{tab:har_noniid}), \ournameFixed stays within $1.5$ points of \textsc{Raw} at every compression ratio and every $\alpha$. The trained bottlenecks separate as the partition sharpens: \textsc{Learned $1{\times}1$} loses $0.6$ to $0.9$ points at $\alpha{=}0.3$ and $5.7$ to $10.9$ points at $\alpha{=}0.1$, and \ournameLearned follows the same trend more mildly. A fixed $\mathbf{R}$ has no trainable parameters, so a client's skewed data cannot bias it; a trained bottleneck fits the few activities its client holds and transfers worse to the rest. \topksparse stays within $1.7$ points of \textsc{Raw} at $8$--$16\times$ for every $\alpha$, but loses about $3$ points at $32\times$ once $\alpha\le0.3$: with only a handful of entries kept per activation, the selection made on a client that holds few activities discards the coordinates the remaining classes need. At $\alpha{=}0.5$ all arms still match the IID result (\textsc{Raw} $94.17\%$ against $94.01\%$ IID); the arms separate only from $\alpha\le0.3$.

\begin{table}[!tbp]
\centering
\small
\setlength{\tabcolsep}{4pt}
\caption{Tabular generality under label heterogeneity: $\Delta$ best test accuracy (points) vs.\ the matched \textsc{Raw} of the same column; $\alpha$ is the Dirichlet concentration, smaller is more heterogeneous; the WCC arms use $\lambda_{\mathrm{WCC}}{=}10^{-1}$.}
\label{tab:har_noniid}
\scaleTable{%
\begin{tabular}{l c c ccc}
\toprule
 & & \textbf{IID} & \multicolumn{3}{c}{\textbf{Non-IID} $\boldsymbol{\alpha}$} \\
\cmidrule(lr){3-3} \cmidrule(lr){4-6}
\textbf{Method} & \textbf{CR} & $\alpha{\to}\infty$ & $0.5$ & $0.3$ & $0.1$ \\
\midrule
\rowcolor{rawgray}
\textsc{Raw} (acc.\ \%) & $1\times$ & $94.01$ & $94.17$ & $92.47$ & $88.31$ \\
\midrule
\ournameFixed & $8\times$   & $+0.2$ & $+0.6$ & $-0.1$ & $-0.2$ \\
              & $16\times$  & $+0.0$ & $+0.5$ & $+0.5$ & $+1.5$ \\
              & $32\times$  & $-0.0$ & $+0.4$ & $+0.1$ & $+0.2$ \\
\addlinespace[2pt]
\ournameFixed$+$WCC    & $8\times$   & $+0.9$ & $+0.3$ & $+1.8$ & $+0.6$ \\
                       & $16\times$  & $-0.0$ & $+0.0$ & $+0.1$ & $-0.2$ \\
                       & $32\times$  & $+0.5$ & $+0.8$ & $+0.8$ & $-3.2$ \\
\addlinespace[2pt]
\ournameLearned        & $8\times$   & $+0.5$ & $+0.4$ & $-1.3$ & $-2.1$ \\
                       & $16\times$  & $+0.2$ & $+0.4$ & $-1.7$ & $-2.5$ \\
                       & $32\times$  & $+0.3$ & $+1.3$ & $-1.1$ & $-6.1$ \\
\addlinespace[2pt]
\ournameLearned$+$WCC  & $8\times$   & $+0.9$ & $+1.1$ & $-0.5$ & $-6.9$ \\
                       & $16\times$  & $+0.9$ & $+0.3$ & $-0.7$ & $-4.9$ \\
                       & $32\times$  & $+0.4$ & $+0.9$ & $-4.3$ & $-5.2$ \\
\midrule
\textsc{Learned $1{\times}1$} & $8\times$  & $+0.7$ & $+0.2$ & $-0.9$ & $-10.9$ \\
                       & $16\times$  & $-0.1$ & $+0.1$ & $-0.8$ & $-5.7$ \\
                       & $32\times$  & $+0.2$ & $+0.7$ & $-0.6$ & $-6.1$ \\
\addlinespace[2pt]
\topksparse & $8\times$  & $+0.1$ & $-0.4$ & $-1.7$ & $+0.8$ \\
                       & $16\times$  & $-0.7$ & $-1.1$ & $-0.3$ & $-0.2$ \\
                       & $32\times$  & $-0.7$ & $-0.9$ & $-3.1$ & $-3.2$ \\
\bottomrule
\end{tabular}}\\[2pt]
\parbox{0.95\columnwidth}{\footnotesize\raggedright Seed spread is ${\pm}0.3$--$1.2$ points at $\alpha{\to}\infty$ and widens to ${\pm}1$--$3.6$ at $\alpha{=}0.1$, so differences below ${\sim}2$ points are noise; no per-cell best is marked. At $\alpha{=}0.1$ one Dirichlet draw hands a client a near-single-activity shard and collapses every arm, \textsc{Raw} included; that column excludes it.}
\end{table}

\subsubsection{Language modelling}
\label{app:llm}
\begin{figure}[!htbp]
    \centering
    \includegraphics[width=0.95\columnwidth]{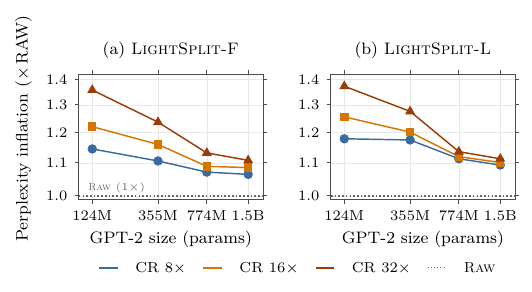}
    \caption{Validation-perplexity inflation over \textsc{Raw} on GPT-2, by model size and compression ratio: (a)~\ournameFixed, (b)~\ournameLearned; the dashed line is \textsc{Raw} ($1\times$).}
    \label{fig:llm_generality}
\end{figure}

Fig.~\ref{fig:llm_generality} shows how much the projection inflates validation perplexity over \textsc{Raw} at each GPT-2 size and compression ratio. Compressing harder costs more, as expected: on the $124$M model the inflation grows from $1.15\times$ at CR\,$8\times$ to $1.36\times$ at $32\times$. The more interesting pattern is that larger models pay less. At CR\,$8\times$ the inflation drops from $1.15\times$ on the $124$M model to $1.06\times$ on the $1.5$B model, and the gap between CR\,$8\times$ and $32\times$ shrinks from $0.21$ to $0.05$ over the same range. Since the cut sits a tenth of the way into every model, what changes between sizes is the width of the residual state at the cut, from $768$ to $1{,}600$ dimensions, so at CR\,$8\times$ the payload keeps $96$ dimensions on the smallest model and $200$ on the largest (App.~\ref{app:setup_llm}). A wider residual state carries more redundancy, and dropping the same fraction of it removes less of what the server-side blocks actually need. \ournameLearned costs a little more than \ournameFixed everywhere ($1.18\times$ against $1.15\times$ on the $124$M model at CR\,$8\times$), but this gap closes as well by $1.5$B ($1.09\times$ against $1.06\times$).

\subsection{Privacy Attacks}
\label{app:attacks}
Each subsection reports the full grid of one attack, in the order of the main text.

\subsubsection{SDAR}
\label{app:sdar}

\begin{table*}[!tbp]
\centering
\small
\setlength{\tabcolsep}{4pt}
\caption{SDAR reconstruction on CIFAR-10 under both split-learning protocols: one row per split level, method, CR and $\lambda_{\mathrm{WCC}}$, \textsc{USL} and \textsc{VSL} side by side. PSNR replaces MSE ($\mathrm{PSNR}{=}{-}10\log_{10}\mathrm{MSE}$, pixel range $[0,1]$); arrows point toward more privacy; a dash in the $\lambda_{\mathrm{WCC}}$ column marks arms without WCC.}
\label{tab:final-sdar-usl-all-split-points}\label{tab:final-sdar-vsl-all-split-points}
\scaleTable{%
\begin{tabular}{l l c c c c c c c c}
\toprule
 & & & & \multicolumn{3}{c}{\textsc{USL}} & \multicolumn{3}{c}{\textsc{VSL}} \\
\cmidrule(lr){5-7}
\cmidrule(l){8-10}
\textbf{Level} & \textbf{Method} & \textbf{CR} & $\boldsymbol{\lambda_{\mathrm{WCC}}}$ & SSIM\,$\downarrow$ & PSNR\,$\downarrow$ & LPIPS$_{20}$\,$\uparrow$ & SSIM\,$\downarrow$ & PSNR\,$\downarrow$ & LPIPS$_{20}$\,$\uparrow$ \\
\midrule
\rowcolor{rawgray} 7 & \textsc{Raw} & $1\times$ & -- & 0.572 & 16.34 & 0.4469 & 0.565 & 16.59 & 0.4415 \\
\rowcolor{rawgray} 7 & \textsc{Raw}\,$+$\,DP & $1\times$ & -- & 0.181 & 12.63 & 0.6290 & 0.254 & 13.59 & 0.6052 \\
\addlinespace[2pt]
7 & \textsc{Learned $1{\times}1$} & $8\times$ & -- & 0.268 & 13.10 & 0.5781 & 0.410 & 15.00 & 0.5247 \\
7 & \textsc{Learned $1{\times}1$} & $16\times$ & -- & 0.274 & 13.58 & 0.5623 & 0.340 & 14.12 & 0.5453 \\
7 & \textsc{Learned $1{\times}1$} & $32\times$ & -- & 0.208 & 13.13 & 0.5979 & 0.252 & 13.23 & 0.5969 \\
\addlinespace[2pt]
7 & \ournameFixed & $8\times$ & $0$ & 0.360 & 15.18 & 0.5738 & 0.342 & 14.93 & 0.5485 \\
7 & \ournameFixed & $16\times$ & $0$ & 0.168 & 13.02 & 0.6453 & 0.330 & 15.31 & 0.5868 \\
7 & \ournameFixed & $32\times$ & $0$ & 0.084 & 10.01 & 0.6845 & 0.250 & 14.24 & 0.6000 \\
\addlinespace[1pt]
7 & \ournameFixed & $8\times$ & $10^{-2}$ & 0.118 & 11.43 & 0.6338 & 0.253 & 14.05 & 0.5904 \\
7 & \ournameFixed & $16\times$ & $10^{-2}$ & 0.098 & 10.82 & 0.6938 & 0.199 & 13.36 & 0.6172 \\
7 & \ournameFixed & $32\times$ & $10^{-2}$ & 0.103 & 11.30 & 0.6861 & 0.181 & 13.01 & 0.6254 \\
\addlinespace[1pt]
7 & \ournameFixed & $8\times$ & $10^{-1}$ & 0.090 & 9.58 & 0.6401 & 0.100 & 11.26 & 0.6545 \\
7 & \ournameFixed & $16\times$ & $10^{-1}$ & 0.077 & 10.46 & 0.6791 & 0.124 & 12.06 & 0.6214 \\
7 & \ournameFixed & $32\times$ & $10^{-1}$ & 0.123 & 11.34 & 0.6977 & 0.109 & 11.35 & 0.6373 \\
\addlinespace[2pt]
7 & \topksparse & $8\times$ & -- & 0.464 & 15.85 & 0.5003 & 0.513 & 16.33 & 0.4848 \\
7 & \topksparse & $16\times$ & -- & 0.379 & 14.88 & 0.5338 & 0.447 & 15.58 & 0.5069 \\
7 & \topksparse & $32\times$ & -- & 0.382 & 15.05 & 0.5480 & 0.346 & 14.52 & 0.5419 \\
\addlinespace[1pt]
7 & \ournameFixed\,$+$\,DP & $8\times$ & $0$ & 0.139 & 11.92 & 0.6451 & 0.264 & 14.07 & 0.5992 \\
\midrule
\rowcolor{rawgray} 4 & \textsc{Raw} & $1\times$ & -- & 0.832 & 21.21 & 0.2899 & 0.794 & 20.47 & 0.2811 \\
\rowcolor{rawgray} 4 & \textsc{Raw}\,$+$\,DP & $1\times$ & -- & 0.486 & 15.14 & 0.4572 & 0.504 & 15.29 & 0.4566 \\
\addlinespace[2pt]
4 & \textsc{Learned $1{\times}1$} & $8\times$ & -- & 0.678 & 17.33 & 0.3524 & 0.685 & 17.72 & 0.3651 \\
4 & \textsc{Learned $1{\times}1$} & $16\times$ & -- & 0.692 & 17.61 & 0.3621 & 0.650 & 16.33 & 0.3732 \\
4 & \textsc{Learned $1{\times}1$} & $32\times$ & -- & 0.667 & 17.24 & 0.4191 & 0.655 & 17.03 & 0.4010 \\
\addlinespace[2pt]
4 & \ournameFixed & $8\times$ & $0$ & 0.524 & 17.88 & 0.4677 & 0.495 & 17.70 & 0.4770 \\
4 & \ournameFixed & $16\times$ & $0$ & 0.465 & 17.40 & 0.4997 & 0.425 & 16.94 & 0.5006 \\
4 & \ournameFixed & $32\times$ & $0$ & 0.347 & 16.15 & 0.5319 & 0.382 & 16.44 & 0.5293 \\
\addlinespace[1pt]
4 & \ournameFixed & $8\times$ & $10^{-2}$ & 0.374 & 15.80 & 0.4968 & 0.519 & 17.10 & 0.4744 \\
4 & \ournameFixed & $16\times$ & $10^{-2}$ & 0.344 & 15.59 & 0.5315 & 0.121 & 10.68 & 0.5660 \\
4 & \ournameFixed & $32\times$ & $10^{-2}$ & 0.057 & 9.93 & 0.6411 & 0.240 & 14.56 & 0.5405 \\
\addlinespace[1pt]
4 & \ournameFixed & $8\times$ & $10^{-1}$ & 0.527 & 17.18 & 0.5331 & 0.081 & 9.95 & 0.5889 \\
4 & \ournameFixed & $16\times$ & $10^{-1}$ & 0.106 & 11.60 & 0.5718 & 0.075 & 10.23 & 0.6018 \\
4 & \ournameFixed & $32\times$ & $10^{-1}$ & -0.044 & 8.28 & 0.6272 & 0.234 & 13.78 & 0.5534 \\
\addlinespace[2pt]
4 & \topksparse & $8\times$ & -- & 0.719 & 18.67 & 0.3266 & 0.697 & 18.53 & 0.3387 \\
4 & \topksparse & $16\times$ & -- & 0.672 & 17.79 & 0.3480 & 0.700 & 18.17 & 0.3410 \\
4 & \topksparse & $32\times$ & -- & 0.567 & 16.19 & 0.3894 & 0.588 & 16.55 & 0.3887 \\
\addlinespace[1pt]
4 & \ournameFixed\,$+$\,DP & $8\times$ & $0$ & 0.300 & 13.90 & 0.5406 & 0.309 & 14.27 & 0.5311 \\
\bottomrule
\end{tabular}%
}\\[2pt]
\parbox{0.95\textwidth}{\footnotesize\raggedright $\mathrm{LPIPS}_{20}$ is computed on the saved $20$-image reconstruction grid. The DP rows ($\sigma{=}0.02$) come from a separate set of seeds whose own \textsc{Raw} attack reaches $0.542$ SSIM at level~7 and $0.821$ at level~4, close to the canonical $0.572$/$0.832$ that Table~\mbox{\ref{tab:dp_privacy}} uses as its reference.}
\end{table*}

\begin{table}[!tbp]
\centering
\caption{SDAR reconstruction against \ournameLearned (CIFAR-10, USL, $L{=}7$ cut, $\lambda_{\mathrm{WCC}}{=}0$), with \textsc{Raw} and \ournameFixed from Table~\ref{tab:final-sdar-usl-all-split-points} for reference; arrows point toward stronger privacy.}
\label{tab:sdar_lsl}
\setlength{\tabcolsep}{5pt}
\begin{tabular}{l c c c c}
\toprule
\textbf{Method} & \textbf{CR} & SSIM\,$\downarrow$ & PSNR\,$\downarrow$ & LPIPS$_{20}$\,$\uparrow$ \\
\midrule
\rowcolor{rawgray} \textsc{Raw} & $1\times$ & 0.572 & 16.34 & 0.4469 \\
\addlinespace[2pt]
\ournameFixed & $8\times$  & 0.360 & 15.18 & 0.5738 \\
\ournameFixed & $16\times$ & 0.168 & 13.02 & 0.6453 \\
\ournameFixed & $32\times$ & 0.084 & 10.01 & 0.6845 \\
\addlinespace[2pt]
\ournameLearned & $8\times$  & 0.177 & 12.45 & 0.6375 \\
\ournameLearned & $16\times$ & 0.117 & 11.25 & 0.6564 \\
\ournameLearned & $32\times$ & 0.139 & 11.86 & 0.6603 \\
\bottomrule
\end{tabular}\\[2pt]
\parbox{0.95\columnwidth}{\footnotesize\raggedright The \ournameLearned rows are means over seeds; $\mathrm{LPIPS}_{20}$ is computed on the saved $20$-image reconstruction grid.}
\end{table}

\begin{table*}[!tbp]
\centering
\small
\setlength{\tabcolsep}{4pt}
\caption{Block-local projection under SDAR (CIFAR-10): reconstruction SSIM as \% of \textsc{Raw}, lower $=$ more private, over tile size $b$, CR and $\lambda_{\mathrm{WCC}}$ at the $4096$-d (left) and $8192$-d (right) cuts; $b{=}8$ is the global projection.}
\label{tab:block_privacy_grid}
\scaleTable{%
\begin{tabular}{l l *{6}{c}}
\toprule
 & & \multicolumn{6}{c}{$\boldsymbol{\lambda_{\mathrm{WCC}}}$} \\
\cmidrule(lr){3-8}
\textbf{CR} & \textbf{Block} & $0$ & $10^{-3}$ & $3{\cdot}10^{-3}$ & $10^{-2}$ & $3{\cdot}10^{-2}$ & $10^{-1}$ \\
\midrule
\multicolumn{8}{c}{\cellcolor{rawgray}\textbf{$4096$-d cut} (\textsc{Raw} SSIM $0.572$)} \\
\midrule
 8$\times$  & $b{=}1$                & 56 & 54 & 44 & 40 & 39 & 17 \\
            & $b{=}2$                & 36 & 30 & 15 & 16 & 16 & 17 \\
            & $b{=}4$                & 38 & 26 & 22 & 16 & 43 & 22 \\
            & $b{=}8$ (\ournameFixed) & 29 & 28 & 20 & 20 & 17 & 20 \\
\addlinespace[2pt]
 16$\times$ & $b{=}1$                & 30 & 29 & 19 & 56 & 22 & 50 \\
            & $b{=}2$                & 48 & 17 & 32 & 21 & 21 & 16 \\
            & $b{=}4$                & 18 & 21 & 19 & 17 & 19 & 17 \\
            & $b{=}8$ (\ournameFixed) & 26 & 23 & 22 & 17 & 35 & 16 \\
\addlinespace[2pt]
 32$\times$ & $b{=}1$                & 28 & 33 & 33 & 25 & 47 & 45 \\
            & $b{=}2$                & 26 & 18 & 28 & 18 & 28 & 21 \\
            & $b{=}4$                & 29 & 19 & 22 & 20 & 20 & 16 \\
            & $b{=}8$ (\ournameFixed) & 16 & 24 & 25 & 20 & 17 & 18 \\
\bottomrule
\end{tabular}\hspace{1.5em}%
\begin{tabular}{l l *{6}{c}}
\toprule
 & & \multicolumn{6}{c}{$\boldsymbol{\lambda_{\mathrm{WCC}}}$} \\
\cmidrule(lr){3-8}
\textbf{CR} & \textbf{Block} & $0$ & $10^{-3}$ & $3{\cdot}10^{-3}$ & $10^{-2}$ & $3{\cdot}10^{-2}$ & $10^{-1}$ \\
\midrule
\multicolumn{8}{c}{\cellcolor{rawgray}\textbf{$8192$-d cut} (\textsc{Raw} SSIM $0.832$)} \\
\midrule
 8$\times$  & $b{=}1$                & 86 & 92 & 94 & 91 & 87 & 82 \\
            & $b{=}2$                & 83 & 78 & 72 & 74 & 73 & 82 \\
            & $b{=}4$                & 72 & 63 & 58 & 61 &  8 & 26 \\
            & $b{=}8$ (\ournameFixed) & 57 & 51 & 33 & 40 &  7 & 47 \\
\addlinespace[2pt]
 16$\times$ & $b{=}1$                & 79 & 84 & 81 & 80 & 85 & 80 \\
            & $b{=}2$                & 78 & 71 & 74 & 75 & 75 & 75 \\
            & $b{=}4$                & 46 & 59 & 55 & 56 & 54 & 46 \\
            & $b{=}8$ (\ournameFixed) & 56 & 41 & 43 & 31 & 36 &  7 \\
\addlinespace[2pt]
 32$\times$ & $b{=}1$                & 74 & 74 & 77 & 79 & 64 & 67 \\
            & $b{=}2$                & 71 & 61 & 63 & 71 & 75 & 49 \\
            & $b{=}4$                & 53 & 45 & 12 & 35 & 41 & 60 \\
            & $b{=}8$ (\ournameFixed) & 42 & 19 & 20 & 20 & 19 &  1 \\
\bottomrule
\end{tabular}}\\[2pt]
\parbox{0.95\textwidth}{\footnotesize\raggedright Each cell is the mean over seeds. Above $\lambda_{\mathrm{WCC}}{=}10^{-2}$ the cells vary strongly between seeds; read them as trends.}
\end{table*}

\begin{figure*}[!tbp]
\centering
\setlength{\tabcolsep}{1.5pt}%
\begin{tabular}{@{}m{9pt}@{\,}cccc@{}}
 & \multicolumn{1}{c}{\scriptsize L4 USL} & \multicolumn{1}{c}{\scriptsize L4 VSL} & \multicolumn{1}{c}{\scriptsize L7 USL} & \multicolumn{1}{c}{\scriptsize L7 VSL} \\
\rotatebox{90}{\scriptsize $\CR{=}8\times$} &
\includegraphics[width=0.235\textwidth]{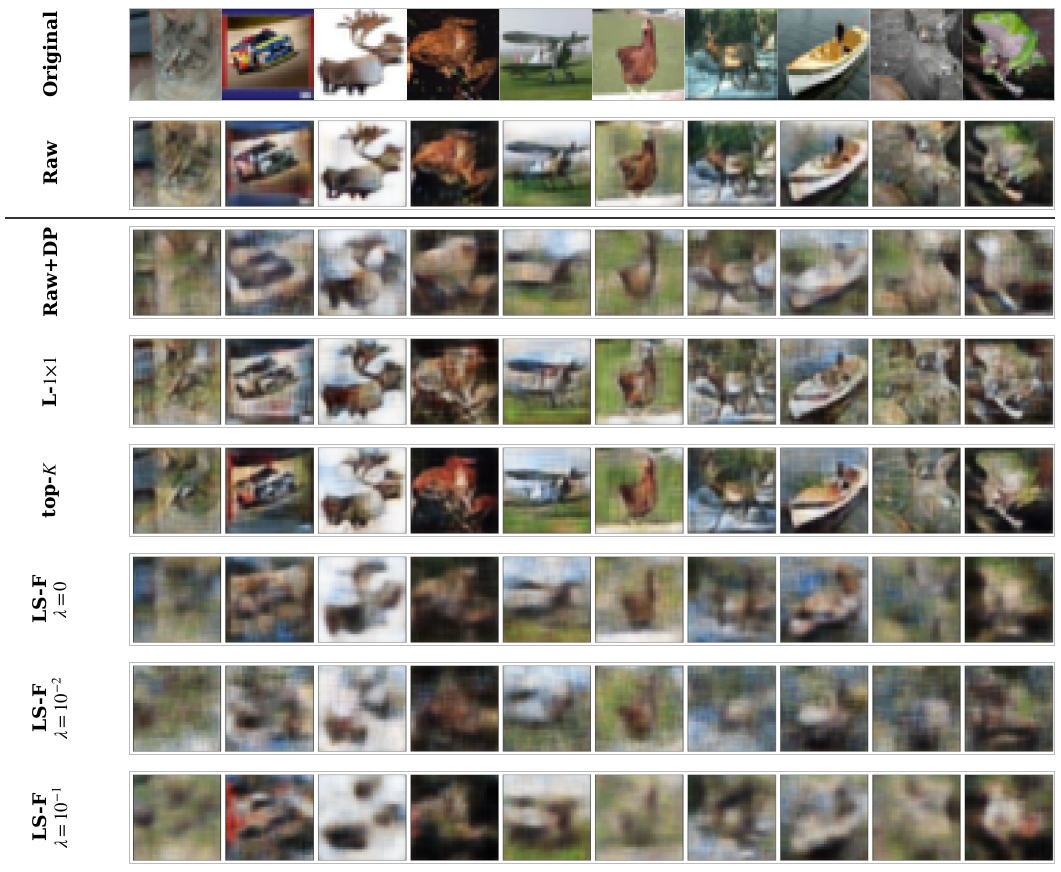} &
\includegraphics[width=0.235\textwidth]{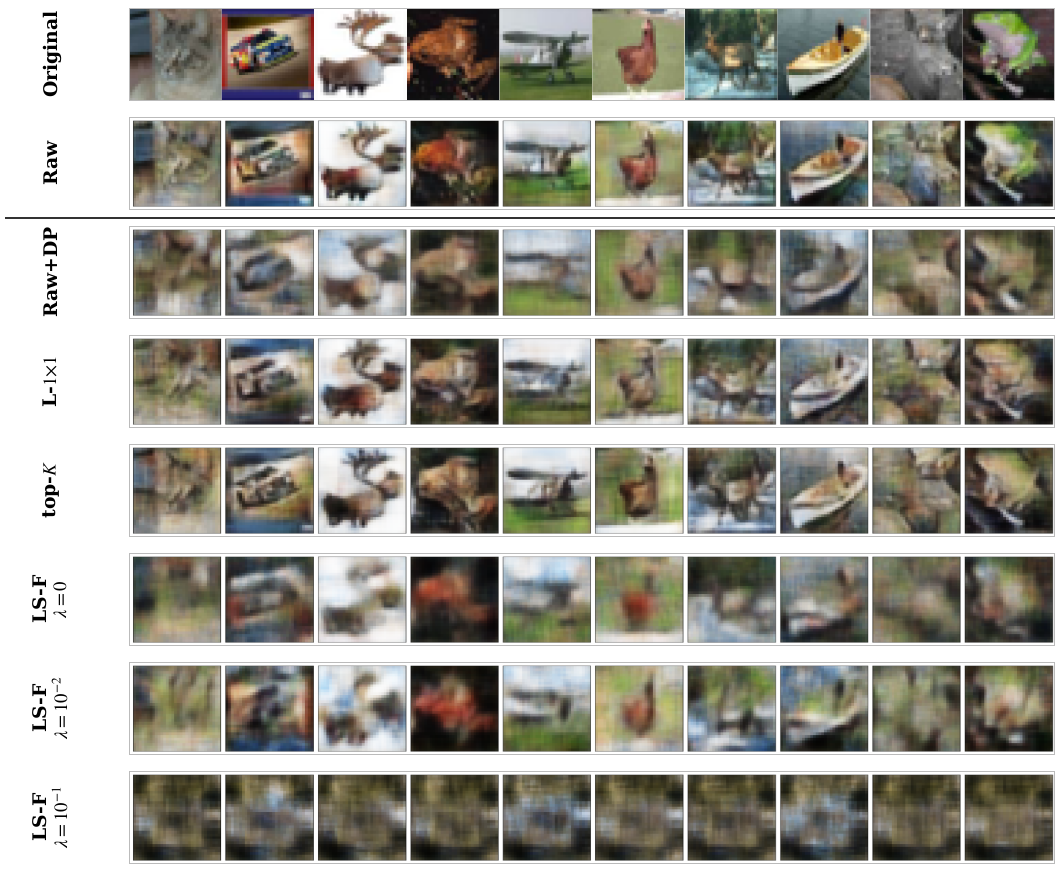} &
\includegraphics[width=0.235\textwidth]{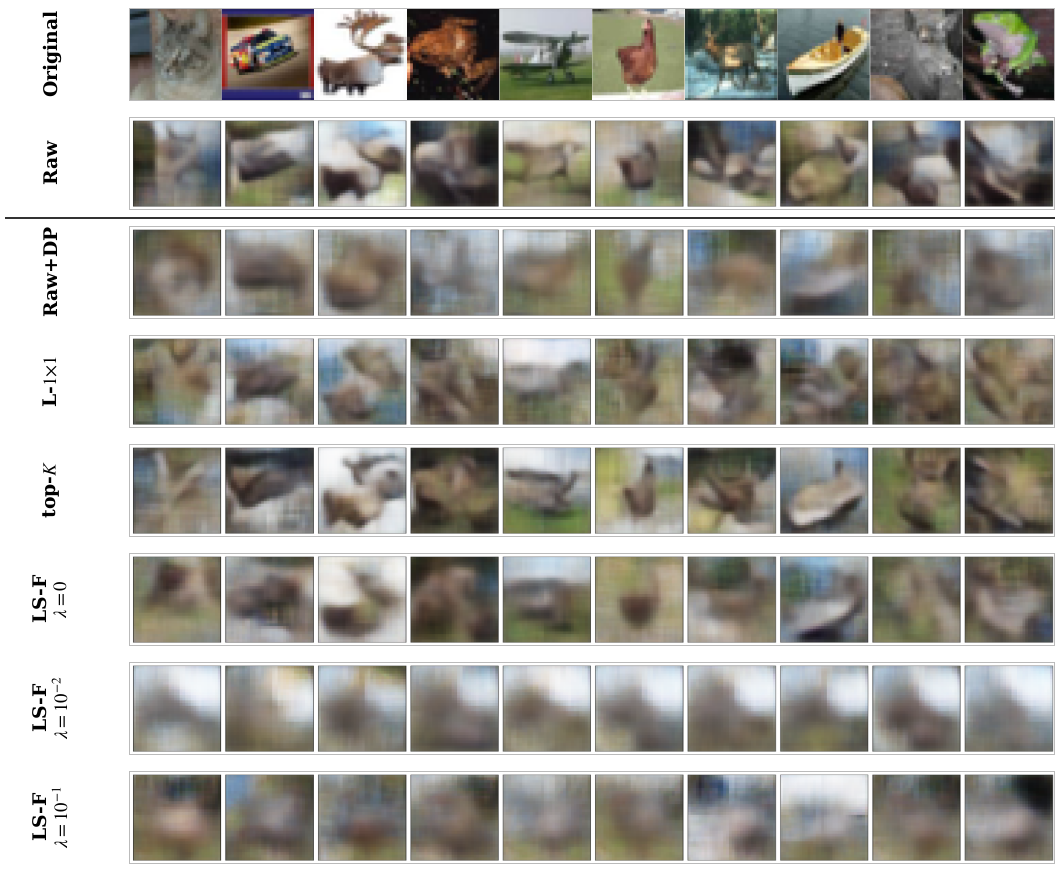} &
\includegraphics[width=0.235\textwidth]{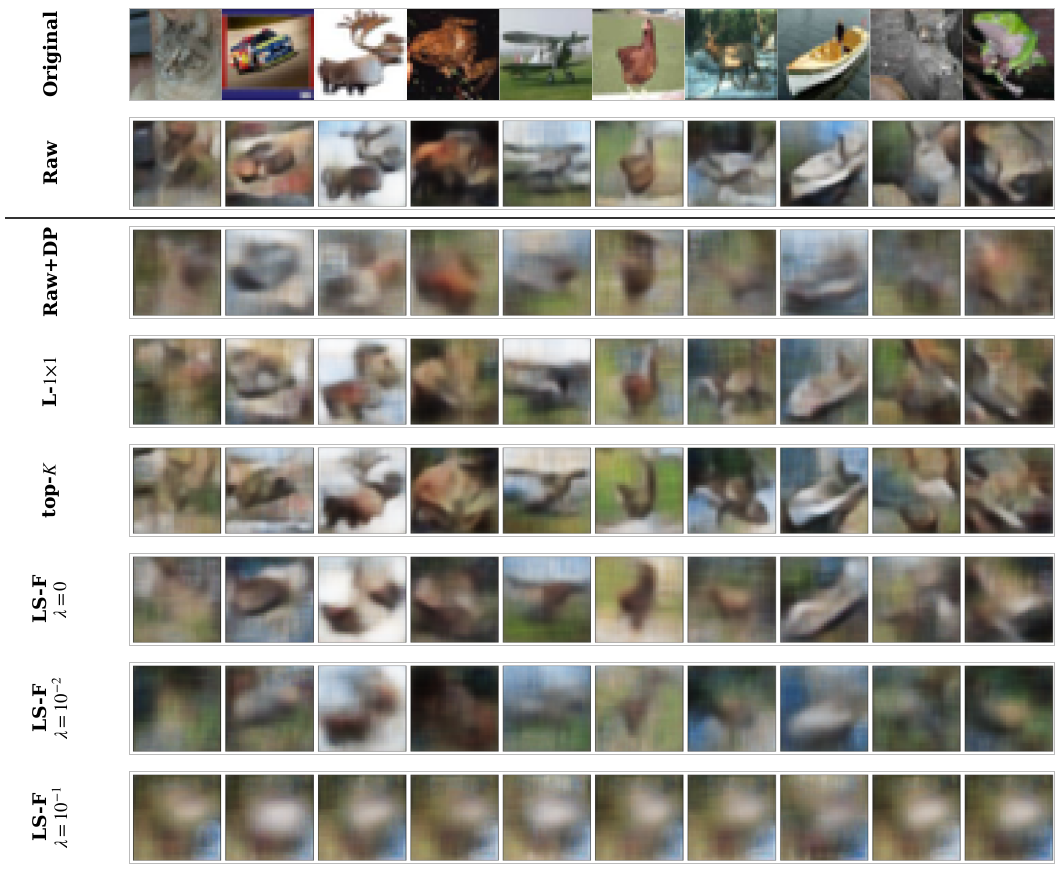} \\
\rotatebox{90}{\scriptsize $\CR{=}16\times$} &
\includegraphics[width=0.235\textwidth]{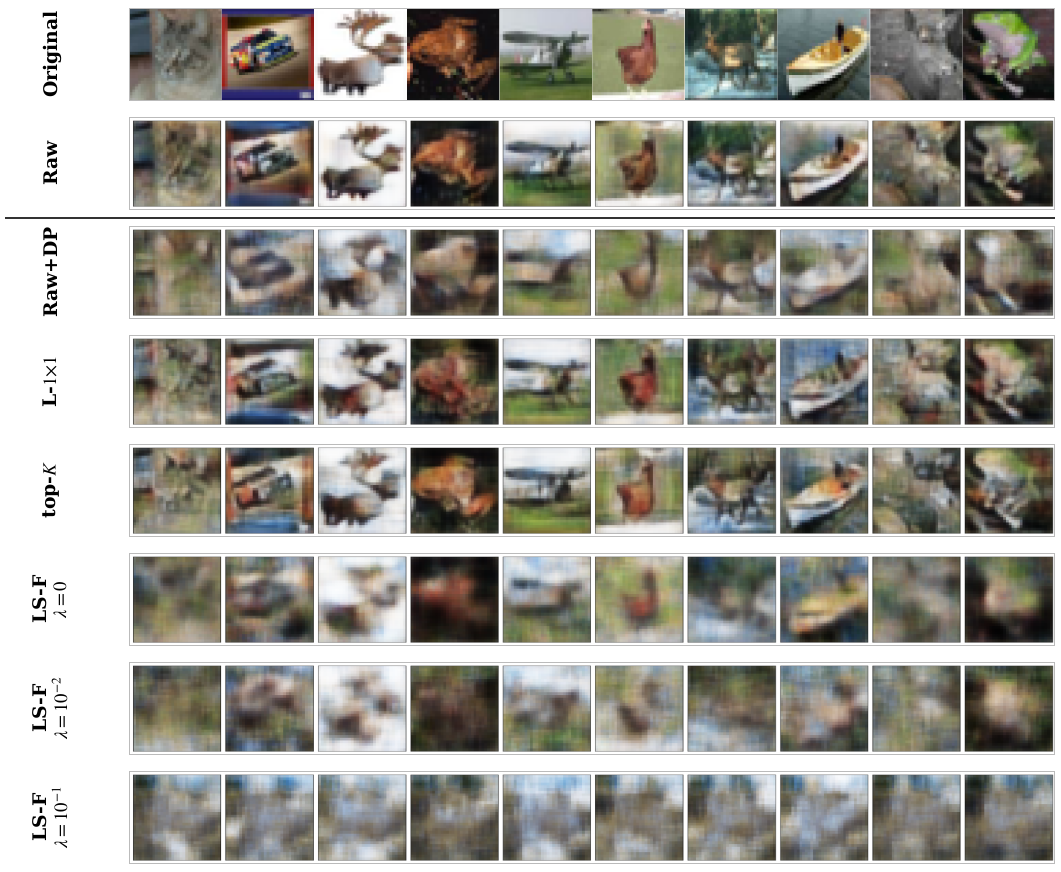} &
\includegraphics[width=0.235\textwidth]{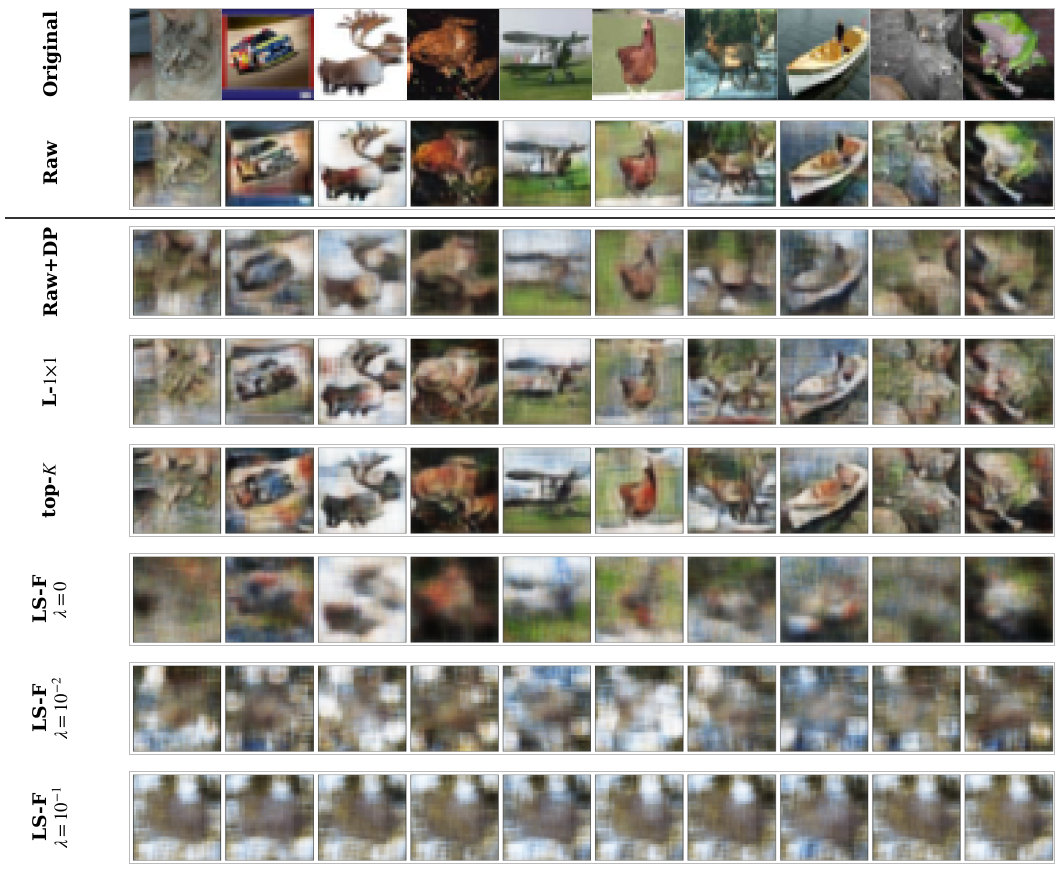} &
\includegraphics[width=0.235\textwidth]{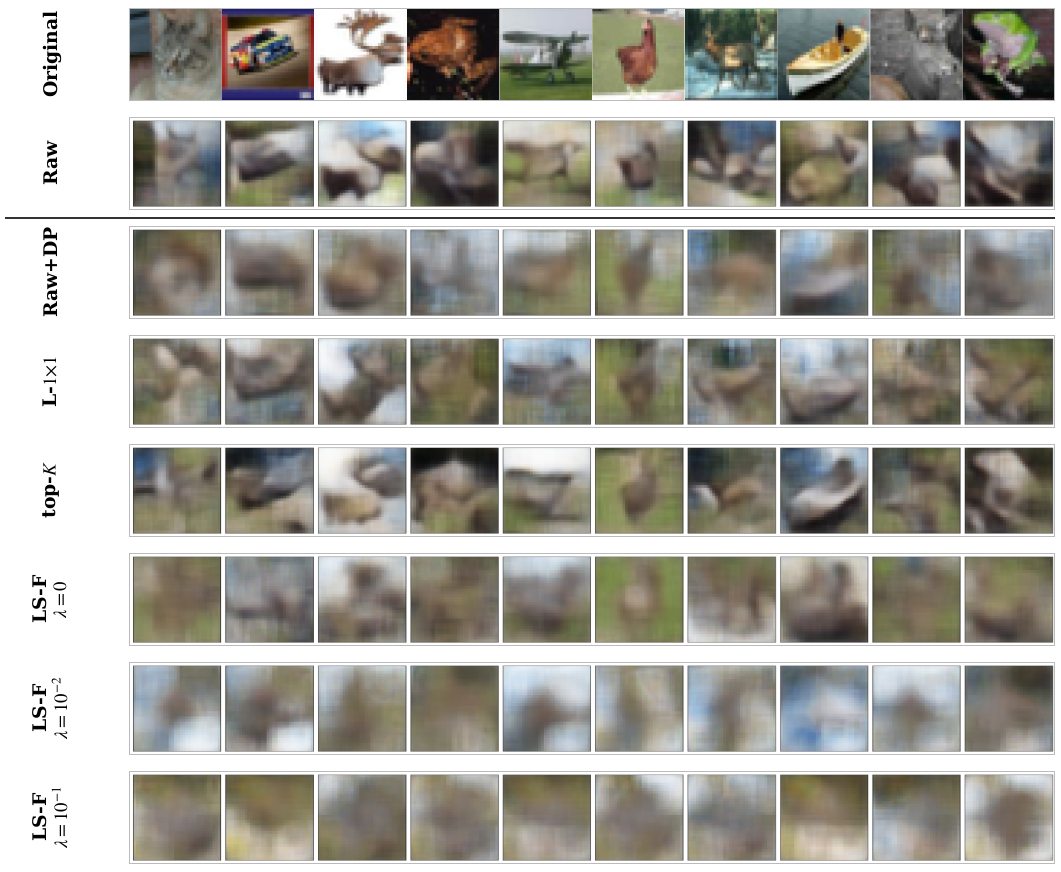} &
\includegraphics[width=0.235\textwidth]{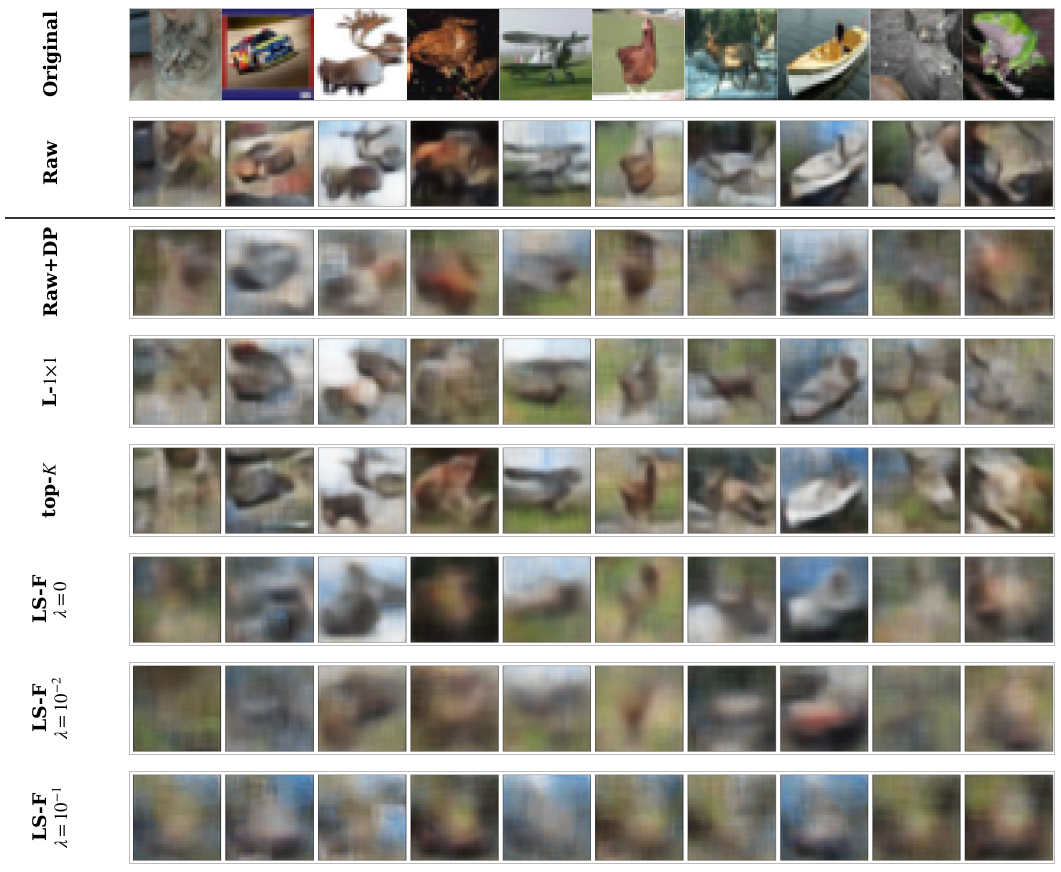} \\
\rotatebox{90}{\scriptsize $\CR{=}32\times$} &
\includegraphics[width=0.235\textwidth]{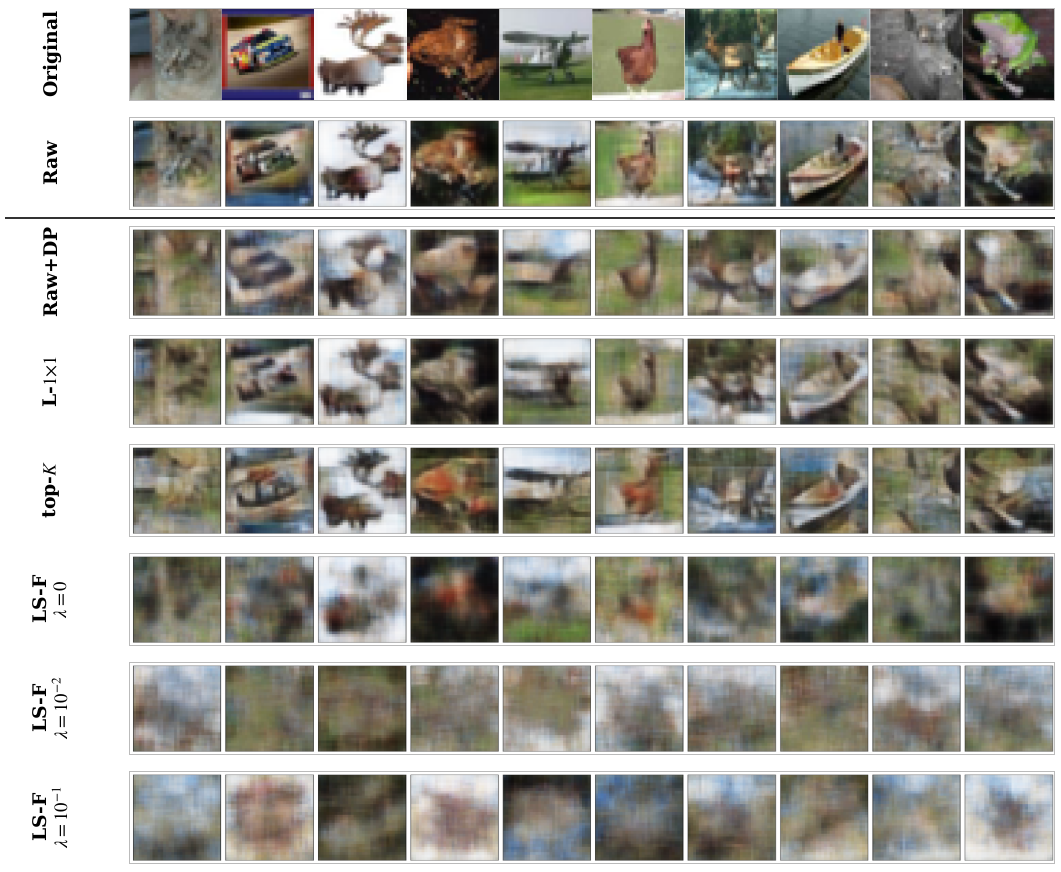} &
\includegraphics[width=0.235\textwidth]{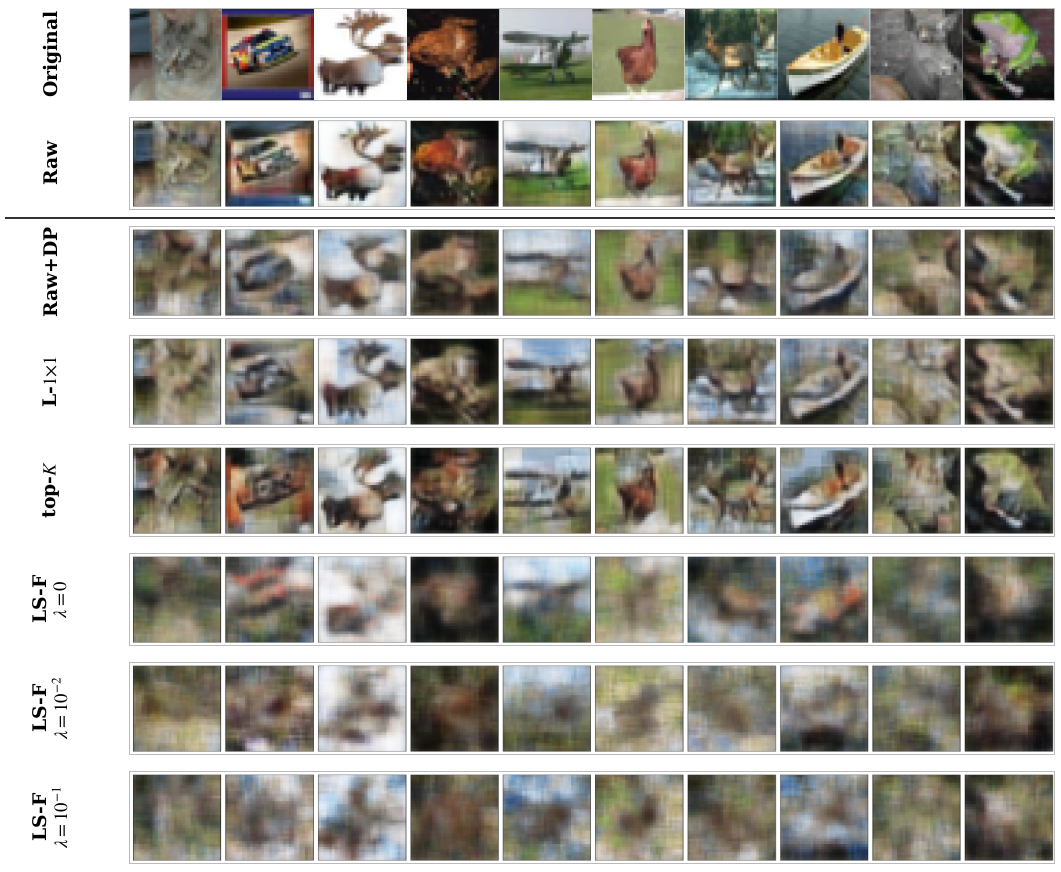} &
\includegraphics[width=0.235\textwidth]{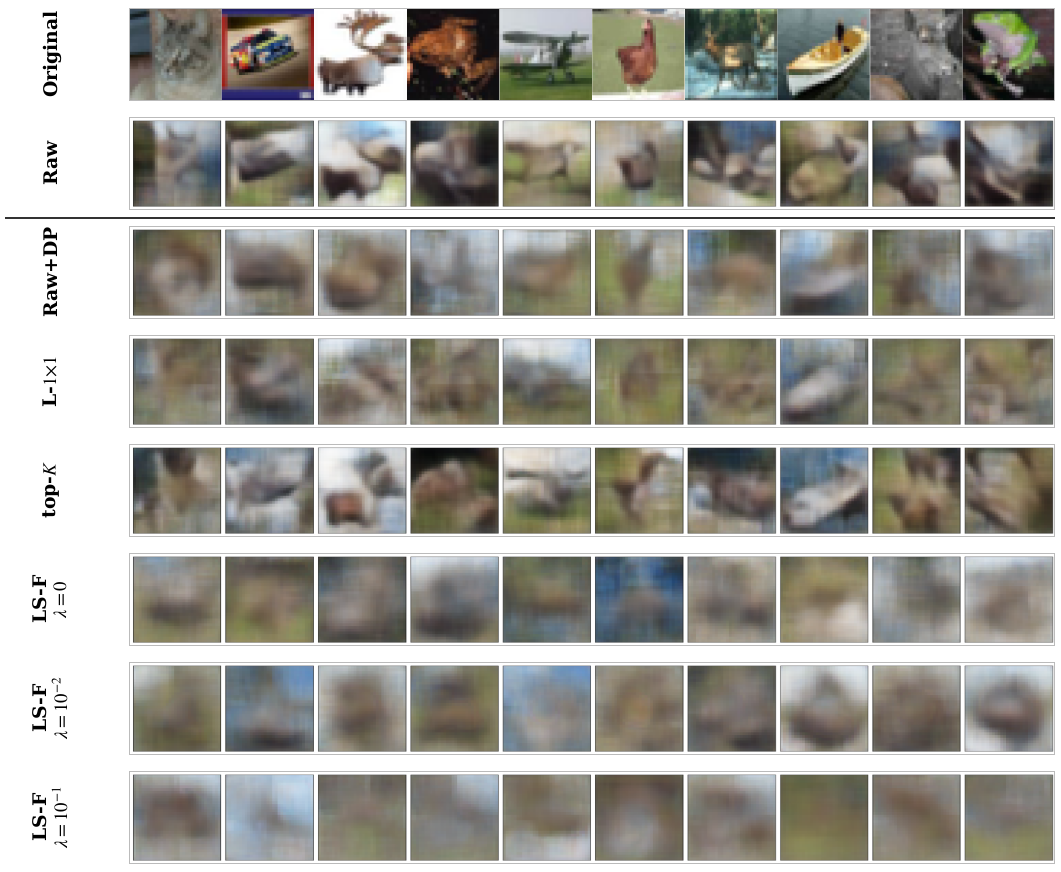} &
\includegraphics[width=0.235\textwidth]{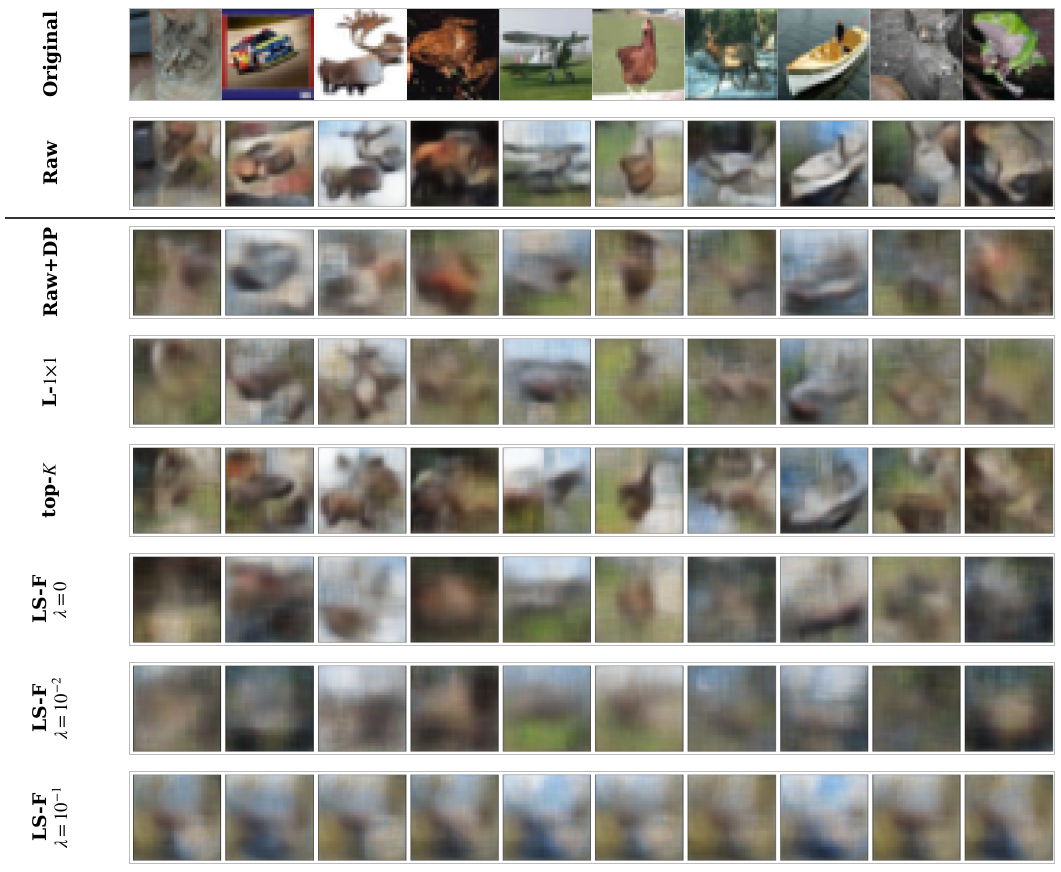} \\
\end{tabular}
\caption{SDAR reconstruction grids. Columns: split level and training mode; rows: compression ratio. Within a panel, rows run from the client images through \textsc{Raw}, \textsc{Raw}\,$+$\,DP, \textsc{Learned $1{\times}1$}, top-$K$ sparse, and \ournameFixed at $\lambda_{\mathrm{WCC}}\in\{0,10^{-2},10^{-1}\}$.}
\label{fig:sdar_recon_grids}
\end{figure*}

Table~\ref{tab:final-sdar-usl-all-split-points} reports the full SDAR grid under both protocols, at every split point, compression ratio, and $\lambda_{\mathrm{WCC}}$ (App.~\ref{app:sdar_setup}). The \textsc{VSL} columns show that \ournameGen privacy gain does not depend on the protocol. With the server holding the labels, \ournameFixed lowers SSIM at CR\,$8\times$ from $0.565$ to $0.342$ at $L{=}7$ and from $0.794$ to $0.495$ at $L{=}4$ without WCC, close to the $0.360$ and $0.524$ under \textsc{USL}, and WCC lowers both further, to $0.100$ and $0.081$ at $\lambda_{\mathrm{WCC}}{=}10^{-1}$; the \textsc{VSL} panels of Fig.~\ref{fig:sdar_recon_grids} show the matching reconstructions.

 Each panel fixes one combination of split level $\{4,7\}$, training mode $\{\textsc{USL},\textsc{VSL}\}$, and compression ratio $\CR{\in}\{8,16,32\}$, and shows reconstructions from \textsc{Raw}, \textsc{Raw}\,$+$\,DP, \textsc{Learned $1{\times}1$}, \topksparse, and \textsc{\ournameFixed} at $\lambda_{\mathrm{WCC}}\in\{0,10^{-2},10^{-1}\}$ on the same ten client images. Reading down a column of any panel shows the effect of WCC at fixed $\CR$; comparing panels at fixed method shows the effect of $\CR$ at fixed level and training mode.

\label{app:sdar_metrics}
\textbf{Metric agreement across the $\lambda_{\mathrm{WCC}}$ sweep.} Table~\mbox{\ref{tab:final-sdar-usl-all-split-points}} sweeps $\lambda_{\mathrm{WCC}}$ at twelve (protocol, cut, CR) cells. In nine of them the pixel metrics (SSIM, PSNR) move toward stronger privacy as $\lambda_{\mathrm{WCC}}$ grows, LPIPS does so in nine, and all three agree in seven. Neither family is monotone everywhere, so we report every metric at every cell. The exceptions are small: two sit at the shallow $L{=}4$ cut at CR$\,8\times$, where the projection removes the least, and the third at $L{=}7$, $32\times$, where SSIM is already down to $0.08$ without WCC and the sweep shifts it by at most $0.04$.

The clearest case is USL $L{=}4$, CR$\,8\times$. SSIM falls from $0.524$ to $0.374$ at $\lambda_{\mathrm{WCC}}{=}10^{-2}$ and returns to $0.527$ at $10^{-1}$, while LPIPS keeps rising ($0.468/0.497/0.533$). WCC pulls same-class activations toward a common center, so the decoder returns an image typical of the class rather than of the sample. The L4 USL, $\CR{=}8\times$ column of Fig.~\mbox{\ref{fig:sdar_recon_grids}} shows what that means: background and dominant color come back, which the pixel metrics reward, and the object stays a blur, which LPIPS penalizes~\mbox{\cite{Zhang_2018_CVPR}}. Every setting stays far from the undefended $0.832$. Additional five converged attacker seeds at this cell show the same behavior: in two of them SSIM at $\lambda_{\mathrm{WCC}}{=}10^{-1}$ rises above every $10^{-2}$ seed ($0.53$ and $0.48$ against $0.30$--$0.42$), in the other three it keeps falling ($0.11$--$0.25$), while LPIPS stays at $0.51$--$0.60$ against $0.50$--$0.54$ in every seed; the pixel metrics swing with the mode the decoder lands in.

\textbf{Block-local projection.} Table~\ref{tab:block_privacy_grid} runs the SDAR black-box attack over the same block sizes at both the $4096$-d cut and the deeper $8192$-d cut, sweeping CR and $\lambda_{\mathrm{WCC}}$. At the $4096$-d cut the projection alone ($\lambda_{\mathrm{WCC}}{=}0$) holds every tile at $16$--$56\%$ of \textsc{Raw} and tile size does not order the reconstruction, so there the block variant inherits \ourname's reconstruction defense unchanged and $b$ is a utility knob at fixed privacy. The $8192$-d cut is more invertible, and there tile size matters. At $\lambda_{\mathrm{WCC}}{=}0$ the global projection $b{=}8$ sits at $42$--$57\%$ of \textsc{Raw}, $b{=}4$ at $46$--$72\%$, and the small tiles at $71$--$86\%$, with the per-position $b{=}1$ leaking most. WCC lowers the wider tiles further ($b{=}8$ falls to $40/31/20\%$ at $10^{-2}$ across the three ratios) but never brings $b{=}1$ below $64\%$. At a deep cut a small tile therefore gives part of the privacy back. DP at $8\times$ (the $+$DP rows of Table~\ref{tab:block_privacy_unsplit}) brings every tile size at the $4096$-d cut to the same floor, $23$--$26\%$ of \textsc{Raw} against $29$--$56\%$ without it; at the $8192$-d cut DP lowers every tile ($86/83/72\%\to69/63/46\%$) and the global projection keeps its lead at $36\%$ (App.~Table~\mbox{\ref{tab:dp_privacy}}).

\subsubsection{FORA}
\label{app:fora}

\begin{table*}[!tbp]
\centering
\small
\setlength{\tabcolsep}{4pt}
\caption{FORA reconstruction on CIFAR-10 over CR and $\lambda_{\mathrm{WCC}}$ for the deployable \emph{pseudo} decoder and the white-box \emph{reference} decoder; $\Delta_\text{Raw}$ is the reference-head LPIPS ratio to \textsc{Raw}.}
\label{tab:fora-full}
\begin{tabular}{l c c c c c c c c c}
\toprule
 & & & \multicolumn{3}{c}{Pseudo decoder} & \multicolumn{4}{c}{Reference decoder (white-box ceiling)} \\
\cmidrule(lr){4-6}
\cmidrule(lr){7-10}
\textbf{Method} & \textbf{CR} & $\boldsymbol{\lambda_{\mathrm{WCC}}}$ & SSIM\,$\downarrow$ & PSNR\,$\downarrow$ & LPIPS\,$\uparrow$ & SSIM\,$\downarrow$ & PSNR\,$\downarrow$ & LPIPS\,$\uparrow$ & $\Delta_\text{Raw}$\,$\uparrow$ \\
\midrule
\rowcolor{rawgray} \textsc{Raw} & $1\times$ & -- & 0.81 & 21.84 & 0.206 & 0.922 & 27.15 & 0.100 & $\times1.0$ \\
\rowcolor{rawgray} \textsc{Raw}\,$+$\,DP & $1\times$ & -- & 0.04 & 9.52 & 0.645 & 0.441 & 15.76 & 0.475 & $\times4.7$ \\
\midrule
\textsc{Learned $1{\times}1$} & $8\times$ & -- & 0.56 & 14.35 & 0.408 & 0.841 & 23.14 & 0.190 & $\times1.9$ \\
\textsc{Learned $1{\times}1$} & $16\times$ & -- & 0.51 & 13.78 & 0.443 & 0.772 & 21.01 & 0.245 & $\times2.5$ \\
\textsc{Learned $1{\times}1$} & $32\times$ & -- & 0.34 & 10.84 & 0.547 & 0.667 & 18.45 & 0.345 & $\times3.4$ \\
\midrule
\textsc{\ournameFixed} & $8\times$ & $0$ & 0.03 & 10.24 & 0.656 & 0.652 & 20.63 & 0.381 & $\times3.8$ \\
\textsc{\ournameFixed} & $16\times$ & $0$ & 0.06 & 10.37 & 0.646 & 0.564 & 19.43 & 0.438 & $\times4.4$ \\
\textsc{\ournameFixed} & $32\times$ & $0$ & 0.05 & 10.16 & 0.644 & 0.489 & 18.62 & 0.481 & $\times4.8$ \\
\addlinespace[1pt]
\textsc{\ournameFixed} & $8\times$ & $10^{-3}$ & 0.12 & 11.83 & 0.610 & 0.607 & 19.99 & 0.405 & $\times4.1$ \\
\addlinespace[1pt]
\textsc{\ournameFixed} & $8\times$ & $10^{-2}$ & 0.08 & 10.93 & 0.643 & 0.555 & 18.75 & 0.445 & $\times4.4$ \\
\textsc{\ournameFixed} & $16\times$ & $10^{-2}$ & 0.05 & 10.41 & 0.644 & 0.468 & 18.03 & 0.477 & $\times4.8$ \\
\textsc{\ournameFixed} & $32\times$ & $10^{-2}$ & 0.04 & 10.48 & 0.627 & 0.434 & 17.68 & 0.500 & $\times5.0$ \\
\addlinespace[1pt]
\textsc{\ournameFixed} & $8\times$ & $10^{-1}$ & $-0.02$ & 10.39 & 0.663 & 0.424 & 13.92 & 0.489 & $\times4.9$ \\
\textsc{\ournameFixed} & $16\times$ & $10^{-1}$ & 0.01 & 10.62 & 0.652 & 0.244 & 10.36 & 0.544 & $\times5.4$ \\
\textsc{\ournameFixed} & $32\times$ & $10^{-1}$ & 0.04 & 11.35 & 0.652 & 0.273 & 9.52 & 0.563 & $\times5.6$ \\
\midrule
\topksparse & $8\times$ & -- & 0.57 & 18.60 & 0.369 & 0.836 & 23.67 & 0.205 & $\times2.0$ \\
\topksparse & $16\times$ & -- & 0.66 & 18.44 & 0.340 & 0.735 & 20.21 & 0.300 & $\times3.0$ \\
\topksparse & $32\times$ & -- & 0.53 & 15.62 & 0.416 & 0.642 & 17.35 & 0.355 & $\times3.6$ \\
\addlinespace[1pt]
\textsc{\ournameFixed}\,$+$\,DP & $8\times$ & $0$ & 0.02 & 9.61 & 0.644 & 0.391 & 15.72 & 0.500 & $\times5.0$ \\
\textsc{\ournameFixed}\,$+$\,DP & $16\times$ & $0$ & 0.05 & 10.10 & 0.652 & 0.324 & 15.76 & 0.528 & $\times5.3$ \\
\textsc{\ournameFixed}\,$+$\,DP & $32\times$ & $0$ & 0.03 & 10.51 & 0.654 & 0.316 & 16.06 & 0.536 & $\times5.4$ \\
\bottomrule
\end{tabular}
\\[2pt]
\parbox{0.95\textwidth}{\footnotesize\raggedright At each CR the payload is $k{=}2048/1024/512$ floats for the projection, $k_\mathrm{ch}{=}8/4/2$ channels for the learned $1{\times}1$, and $2k/3$ coordinates for \topksparse, which spends the remainder on the $2$-byte index of each kept coordinate. Each cell is the mean over the attacker runs trained for it. The pseudo decoder does not converge in every run; the pseudo columns average the converged runs only. The reference decoder converged in every run. Pseudo cells read against the \textsc{Raw} pseudo row; $\Delta_\text{Raw}$ uses the reference head only.}
\end{table*}

Table~\ref{tab:fora-full} reports the full FORA grid for both decoders at every compression ratio and $\lambda_{\mathrm{WCC}}$ (App.~\ref{app:fora_setup}).

\begin{figure*}[!tbp]
\centering
\includegraphics[width=0.92\textwidth]{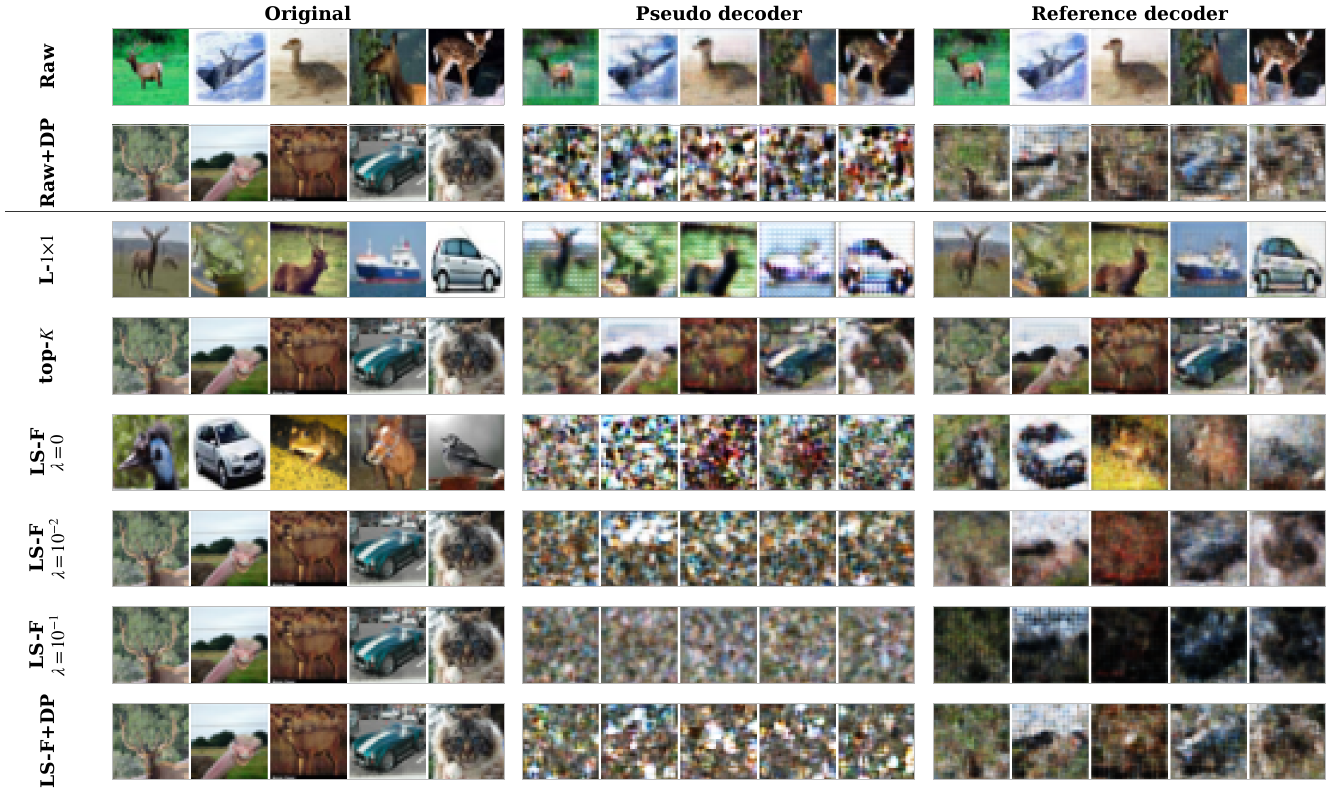}
\caption{FORA reconstructions on CIFAR-10 at the $8\times$ tier (\textsc{Raw} and \textsc{Raw}\,$+$\,DP at $1\times$). Each row shows one method's evaluation images beside the deployable pseudo decoder and the white-box reference decoder, from the strongest attacker run of its Table~\ref{tab:fora-full} cell.}
\label{fig:fora_recon}
\end{figure*}

The full grid adds the $16\times$ and $32\times$ ratios and the DP rows to the CR\,$8\times$ slice of the main text. The pseudo decoder fails at every ratio: \ournameFixed holds its SSIM at or below $0.12$ whatever the CR or $\lambda_{\mathrm{WCC}}$, while \textsc{Learned $1{\times}1$} and \topksparse remain recognisable even at $32\times$ (SSIM $0.34$ and $0.53$). On the reference decoder, the leakage ceiling of App.~\ref{app:fora_setup}, compression itself helps every method, but the ordering holds at each ratio: at $\lambda_{\mathrm{WCC}}{=}0$ \ournameFixed reaches $\Delta_{\text{Raw}}$ $\times3.8$/$\times4.4$/$\times4.8$ over the three ratios against $\times1.9$/$\times2.5$/$\times3.4$ for \textsc{Learned $1{\times}1$} and $\times2.0$/$\times3.0$/$\times3.6$ for \topksparse, and $\lambda_{\mathrm{WCC}}{=}10^{-1}$ lifts it to $\times4.9$/$\times5.4$/$\times5.6$. DP composes here as it does under SDAR: \ournameFixed$+$DP at $\lambda_{\mathrm{WCC}}{=}0$ sits at $\times5.0$--$5.4$, above \textsc{Raw}$+$DP at $\times4.7$. Fig.~\ref{fig:fora_recon} shows the two decoders side by side at $8\times$.

\textbf{Block-local projection.} The FORA columns of Table~\mbox{\ref{tab:block_privacy_unsplit}} run the same attack against the block variant at the matched payloads $k{=}2048/1024/512$. The pseudo decoder fails on the tiles as it does on the global projection: every cell reads SSIM $0.27$ or below against $0.81$ on \textsc{Raw}, except the per-position tile $b{=}1$ at $16\times$ ($0.58$). On the white-box reference decoder the tiles stay closer to \textsc{Raw}'s $0.92$ than the global projection does, $0.57$ to $0.87$ against $0.49$ to $0.65$ (Table~\mbox{\ref{tab:fora-full}}), and the compression ratio, not the tile size, drives the decline.

\subsubsection{UnSplit}
\label{app:unsplit}

\begin{table*}[!tbp]
\centering
\small
\setlength{\tabcolsep}{3.5pt}
\caption{Full UnSplit sweep on \texttt{MnistNet}: victim accuracy, the foreground-masked reconstruction error relative to \textsc{Raw} ($\mathrm{MSE}_{\mathrm{fg}}/\textsc{Raw}$, larger $=$ more resistant) and the whole-image metrics over all $28{\times}28$ pixels; arrows point toward more privacy.}
\label{tab:unsplit_attack_full}\label{tab:unsplit_defenses_whole}
\scaleTable{%
\begin{tabular}{l c c r r r r r r r r r r r r}
\toprule
 & & & \multicolumn{6}{c}{\textbf{MNIST}} & \multicolumn{6}{c}{\textbf{Fashion-MNIST}} \\
\cmidrule(lr){4-9}\cmidrule(l){10-15}
 & & & & & \multicolumn{4}{c}{whole image} & & & \multicolumn{4}{c}{whole image} \\
\cmidrule(lr){6-9}\cmidrule(l){12-15}
\textbf{Method} & $\boldsymbol{\lambda_{\mathrm{WCC}}}$ & \textbf{CR} & \shortstack{Acc.\\(\%)} & \shortstack{$\mathrm{MSE}_{\mathrm{fg}}$\\$/\textsc{Raw}\,\uparrow$} & MSE\,$\uparrow$ & SSIM\,$\downarrow$ & PSNR\,$\downarrow$ & LPIPS\,$\uparrow$ & \shortstack{Acc.\\(\%)} & \shortstack{$\mathrm{MSE}_{\mathrm{fg}}$\\$/\textsc{Raw}\,\uparrow$} & MSE\,$\uparrow$ & SSIM\,$\downarrow$ & PSNR\,$\downarrow$ & LPIPS\,$\uparrow$ \\
\midrule
\multicolumn{15}{l}{\emph{Cut at split layer 2 \,(\texttt{MnistNet} post-pool$_1$, $d{=}1152$, $k{\in}\{144,72,36\}$).}} \\
\midrule
\rowcolor{rawgray} \textsc{Raw} & -- & $1\times$ & 99.02 & $\times 1.00$ & 0.1323 & 0.1798 & 8.85 & 0.4540 & 88.64 & $\times 1.00$ & 0.3010 & 0.0599 & 5.26 & 0.7112 \\
\rowcolor{rawgray} \textsc{Raw}\,$+$\,DP & -- & $1\times$ & 98.9 & $\times 1.23$ & 0.1618 & 0.0957 & 7.97 & 0.6832 & 89.5 & $\times 1.07$ & 0.3244 & 0.0375 & 4.98 & 0.7814 \\
\addlinespace[2pt]
\textsc{Learned $1{\times}1$} & -- & $8\times$ & 98.65 & $\times 1.03$ & 0.1352 & 0.0763 & 8.69 & 0.4792 & 87.90 & $\times 1.06$ & 0.3232 & 0.0086 & 4.98 & 0.7102 \\
\textsc{Learned $1{\times}1$} & -- & $16\times$ & 98.60 & $\times 0.97$ & 0.1262 & 0.1819 & 8.99 & 0.4450 & 88.60 & $\times 1.08$ & 0.3355 & 0.0280 & 4.92 & 0.7635 \\
\textsc{Learned $1{\times}1$} & -- & $32\times$ & 98.95 & $\times 1.08$ & 0.1406 & 0.1546 & 8.52 & 0.5309 & 88.45 & $\times 0.93$ & 0.2790 & 0.0532 & 5.56 & 0.6053 \\
\addlinespace[2pt]
\textsc{\ournameFixed} & $0$ & $8\times$ & 97.95 & $\times 1.27$ & 0.1660 & 0.1281 & 7.82 & 0.6811 & 89.00 & $\times 1.09$ & 0.3243 & 0.0434 & 4.90 & 0.7683 \\
\textsc{\ournameFixed} & $0$ & $16\times$ & 97.66 & $\times 1.27$ & 0.1669 & 0.1278 & 7.81 & 0.6815 & 87.01 & $\times 1.09$ & 0.3240 & 0.0434 & 4.90 & 0.7683 \\
\textsc{\ournameFixed} & $0$ & $32\times$ & 97.56 & $\times 1.27$ & 0.1759 & 0.1045 & 7.83 & 0.6954 & 84.03 & $\times 1.08$ & 0.3205 & 0.0451 & 4.94 & 0.7697 \\
\addlinespace[1pt]
\textsc{\ournameFixed} & $10^{-2}$ & $8\times$ & 97.50 & $\times 1.27$ & 0.1656 & 0.1278 & 7.81 & 0.6813 & 86.05 & $\times 1.09$ & 0.3240 & 0.0434 & 4.90 & 0.7683 \\
\textsc{\ournameFixed} & $10^{-2}$ & $16\times$ & 97.46 & $\times 1.27$ & 0.1656 & 0.1278 & 7.81 & 0.6813 & 85.84 & $\times 1.09$ & 0.3240 & 0.0434 & 4.90 & 0.7683 \\
\textsc{\ournameFixed} & $10^{-2}$ & $32\times$ & 95.90 & $\times 1.27$ & 0.1656 & 0.1278 & 7.81 & 0.6813 & 85.99 & $\times 1.09$ & 0.3240 & 0.0434 & 4.90 & 0.7683 \\
\addlinespace[1pt]
\textsc{\ournameFixed} & $10^{-1}$ & $8\times$ & 97.20 & $\times 1.27$ & 0.1656 & 0.1278 & 7.81 & 0.6813 & 84.70 & $\times 1.09$ & 0.3240 & 0.0434 & 4.90 & 0.7683 \\
\textsc{\ournameFixed} & $10^{-1}$ & $16\times$ & 96.95 & $\times 1.27$ & 0.1656 & 0.1278 & 7.81 & 0.6813 & 85.75 & $\times 1.09$ & 0.3240 & 0.0434 & 4.90 & 0.7683 \\
\textsc{\ournameFixed} & $10^{-1}$ & $32\times$ & 96.65 & $\times 1.27$ & 0.1656 & 0.1278 & 7.81 & 0.6813 & 84.05 & $\times 1.09$ & 0.3240 & 0.0434 & 4.90 & 0.7683 \\
\addlinespace[2pt]
\topksparse & -- & $8\times$ & 98.6 & $\times 1.12$ & 0.1478 & 0.1436 & 8.36 & 0.4898 & 87.8 & $\times 1.05$ & 0.3176 & 0.0355 & 5.05 & 0.6138 \\
\topksparse & -- & $16\times$ & 98.4 & $\times 1.21$ & 0.1594 & 0.1297 & 8.03 & 0.5756 & 83.8 & $\times 1.09$ & 0.3240 & 0.0434 & 4.90 & 0.7683 \\
\topksparse & -- & $32\times$ & 97.8 & $\times 1.24$ & 0.1622 & 0.1296 & 7.92 & 0.6304 & 79.8 & $\times 1.08$ & 0.3242 & 0.0437 & 4.91 & 0.7640 \\
\addlinespace[1pt]
\textsc{\ournameFixed}\,$+$\,DP & $0$ & $8\times$ & 97.8 & $\times 1.27$ & 0.1656 & 0.1278 & 7.81 & 0.6813 & 87.3 & $\times 1.11$ & 0.3240 & 0.0434 & 4.90 & 0.7683 \\
\midrule
\multicolumn{15}{l}{\emph{Cut at split layer 4 \,(\texttt{MnistNet} post-ReLU$_2$, $d{=}1024$, $k{\in}\{128,64,32\}$).}} \\
\midrule
\rowcolor{rawgray} \textsc{Raw} & -- & $1\times$ & 98.96 & $\times 1.00$ & 0.1771 & 0.0892 & 7.81 & 0.5702 & 88.82 & $\times 1.00$ & 0.3224 & 0.0450 & 4.95 & 0.7451 \\
\rowcolor{rawgray} \textsc{Raw}\,$+$\,DP & -- & $1\times$ & 99.1 & $\times 1.00$ & 0.1646 & 0.1288 & 7.84 & 0.6813 & 89.5 & $\times 1.01$ & 0.3236 & 0.0440 & 4.91 & 0.7692 \\
\addlinespace[2pt]
\textsc{Learned $1{\times}1$} & -- & $8\times$ & 98.60 & $\times 0.82$ & 0.1466 & 0.0448 & 8.50 & 0.4397 & 88.35 & $\times 0.97$ & 0.3277 & 0.0439 & 5.07 & 0.6227 \\
\textsc{Learned $1{\times}1$} & -- & $16\times$ & 97.45 & $\times 0.91$ & 0.1621 & 0.0657 & 8.18 & 0.5142 & 88.05 & $\times 0.98$ & 0.3493 & 0.0438 & 5.05 & 0.6290 \\
\textsc{Learned $1{\times}1$} & -- & $32\times$ & 98.25 & $\times 0.85$ & 0.1529 & 0.0263 & 8.39 & 0.4734 & 88.70 & $\times 0.94$ & 0.3248 & 0.0094 & 5.18 & 0.6519 \\
\addlinespace[2pt]
\textsc{\ournameFixed} & $0$ & $8\times$ & 98.30 & $\times 0.89$ & 0.1628 & 0.0600 & 8.26 & 0.6175 & 89.00 & $\times 1.01$ & 0.3239 & 0.0434 & 4.90 & 0.7683 \\
\textsc{\ournameFixed} & $0$ & $16\times$ & 98.19 & $\times 0.97$ & 0.1688 & 0.1178 & 7.95 & 0.6463 & 87.11 & $\times 1.01$ & 0.3378 & 0.0436 & 4.90 & 0.7702 \\
\textsc{\ournameFixed} & $0$ & $32\times$ & 98.58 & $\times 1.00$ & 0.1920 & 0.1252 & 7.82 & 0.6825 & 86.96 & $\times 1.01$ & 0.3253 & 0.0437 & 4.90 & 0.7692 \\
\addlinespace[1pt]
\textsc{\ournameFixed} & $10^{-2}$ & $8\times$ & 98.25 & $\times 1.00$ & 0.1656 & 0.1278 & 7.81 & 0.6813 & 87.75 & $\times 1.01$ & 0.3240 & 0.0434 & 4.90 & 0.7683 \\
\textsc{\ournameFixed} & $10^{-2}$ & $16\times$ & 98.68 & $\times 1.00$ & 0.1656 & 0.1278 & 7.81 & 0.6813 & 86.91 & $\times 1.01$ & 0.3240 & 0.0434 & 4.90 & 0.7683 \\
\textsc{\ournameFixed} & $10^{-2}$ & $32\times$ & 96.97 & $\times 1.00$ & 0.1656 & 0.1278 & 7.81 & 0.6813 & 84.77 & $\times 1.01$ & 0.3240 & 0.0434 & 4.90 & 0.7683 \\
\addlinespace[1pt]
\textsc{\ournameFixed} & $10^{-1}$ & $8\times$ & 96.95 & $\times 1.00$ & 0.1656 & 0.1278 & 7.81 & 0.6813 & 86.10 & $\times 1.01$ & 0.3240 & 0.0434 & 4.90 & 0.7683 \\
\textsc{\ournameFixed} & $10^{-1}$ & $16\times$ & $\dagger$ & -- & -- & -- & -- & -- & 84.45 & $\times 1.01$ & 0.3240 & 0.0434 & 4.90 & 0.7683 \\
\textsc{\ournameFixed} & $10^{-1}$ & $32\times$ & $\dagger$ & -- & -- & -- & -- & -- & 85.10 & $\times 1.01$ & 0.3240 & 0.0434 & 4.90 & 0.7683 \\
\addlinespace[2pt]
\topksparse & -- & $8\times$ & 99.0 & $\times 0.96$ & 0.1626 & 0.1052 & 8.01 & 0.5102 & 89.3 & $\times 1.00$ & 0.3222 & 0.0449 & 4.96 & 0.7667 \\
\topksparse & -- & $16\times$ & 98.3 & $\times 0.98$ & 0.1649 & 0.1286 & 7.90 & 0.6188 & 88.5 & $\times 1.01$ & 0.3243 & 0.0438 & 4.92 & 0.7684 \\
\topksparse & -- & $32\times$ & 96.9 & $\times 0.99$ & 0.1676 & 0.1140 & 7.86 & 0.6736 & 85.2 & $\times 1.01$ & 0.3248 & 0.0430 & 4.91 & 0.7580 \\
\addlinespace[1pt]
\textsc{\ournameFixed}\,$+$\,DP & $0$ & $8\times$ & 97.8 & $\times 1.00$ & 0.1656 & 0.1278 & 7.81 & 0.6813 & 87.9 & $\times 1.01$ & 0.3240 & 0.0434 & 4.90 & 0.7683 \\
\bottomrule
\end{tabular}%
}\\[2pt]
\parbox{0.95\textwidth}{\footnotesize\raggedright $^{\dagger}$: at this setting the victim collapses to chance accuracy in every run we made, the WCC term pulling the activations to a point before the classifier learns; fresh seeds collapse the same way, so we leave the cells empty. Rows with identical whole-image metrics sit on a plateau: once the projection removes the recoverable detail, the inversion converges to the same output.}
\end{table*}

Table~\ref{tab:unsplit_attack_full} reports the full UnSplit grid at both cuts (App.~\ref{app:unsplit_setup}).

\textbf{Block-local projection.} The UnSplit columns of Table~\ref{tab:block_privacy_unsplit} run the same attack against the block-local variant on \texttt{MnistNet} at the split-2 cut. Whole-image reconstruction SSIM stays low at every tile size: at most $0.229$ on MNIST and $0.073$ on Fashion-MNIST. Most tiles land in the same band as the global projection $b{=}8$ ($0.105$ to $0.128$ on MNIST, $0.038$ to $0.053$ on Fashion-MNIST); the two cells that sit above it are $b{=}1$ at $8\times$ ($0.224$) and $b{=}2$ at $32\times$ ($0.229$), both on MNIST. Adding DP at $8\times$ removes that gap: every tile then reads $0.114$ to $0.127$ on MNIST and $0.043$ to $0.046$ on Fashion-MNIST, the level of the global projection.

\subsubsection{Tabular probes}
\label{app:tabular_privacy}
\begin{figure}[!htbp]
    \centering
    \includegraphics[width=\columnwidth]{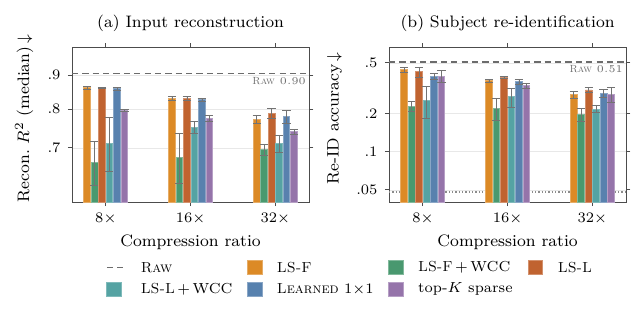}
    \caption{Privacy probes on tabular data (UCI HAR) over the compression ratio, lower $=$ more private: (a)~median per-feature $R^2$ of input reconstruction; (b)~subject re-identification accuracy. The dashed line is \textsc{Raw}, the dotted line is chance ($0.048$), error bars span the seeds.}
    \label{fig:har_privacy}
\end{figure}

Fig.~\ref{fig:har_privacy} sweeps both HAR probes of App.~\ref{app:tabular_attack}, feature reconstruction and subject re-identification, over the compression ratios for every method, with and without WCC. Both curves fall as compression tightens, and WCC lowers them further at every ratio; \sect\ref{subsec:tabular-privacy} reports the values. The low $R^2$ of \topksparse comes from the zeroed coordinates, which the regressor cannot recover, and its re-identification accuracy matches the other arms, so the dip is not a privacy gain.

\subsubsection{Active attack: FSHA}
\label{app:fsha_active}
\begin{table}[!htbp]
\centering
\caption{FSHA reconstruction at CR\,${=}\,8\times$ on MNIST and Fashion-MNIST: clean accuracy of the victim and attack quality, $\Delta_{\text{Raw}}$ the LPIPS ratio to \textsc{Raw}; arrows point toward more privacy.}
\label{tab:fsha_active}
\footnotesize
\setlength{\tabcolsep}{4pt}
\resizebox{\columnwidth}{!}{%
\begin{tabular}{lccccc}
\toprule
\textbf{Method} & Acc.\ (\%)\,$\uparrow$ & SSIM\,$\downarrow$ & MSE\,$\uparrow$ & LPIPS\,$\uparrow$ & $\Delta_{\text{Raw}}$\,$\uparrow$ \\
\midrule
\multicolumn{6}{@{}l}{\textit{MNIST}}\\
\rowcolor{rawgray}
\textsc{Raw} & $98.9$ & $0.679$ & $0.004$ & $0.010$ & $\times1.0$ \\
\rowcolor{rawgray}
\textsc{Raw}\,$+$\,DP & $99.2$ & $0.605$ & $0.008$ & $0.019$ & $\times1.8$ \\
\textsc{Learned $1{\times}1$} & $98.6$ & $0.726$ & $0.005$ & $0.016$ & $\times1.5$ \\
\ournameFixed & $98.9$ & $0.634$ & $0.038$ & $0.086$ & $\times8.4$ \\
\ournameFixed\,$+$WCC ($\lambda{=}0.1$) & $98.5$ & $0.483$ & $0.063$ & $0.182$ & $\times17.7$ \\
\ournameFixed\,$+$WCC ($\lambda{=}0.3$) & $97.8$ & $0.433$ & $0.061$ & $0.187$ & $\times18.3$ \\
\topksparse & $98.6$ & $0.885$ & $0.005$ & $0.012$ & $\times1.2$ \\
\ournameFixed\,$+$\,DP & $99.3$ & $0.342$ & $0.084$ & $0.172$ & $\times16.7$ \\
\midrule
\multicolumn{6}{@{}l}{\textit{Fashion-MNIST}}\\
\rowcolor{rawgray}
\textsc{Raw} & $90.6$ & $0.606$ & $0.007$ & $0.039$ & $\times1.0$ \\
\rowcolor{rawgray}
\textsc{Raw}\,$+$\,DP & $91.8$ & $0.500$ & $0.015$ & $0.071$ & $\times1.8$ \\
\textsc{Learned $1{\times}1$} & $90.6$ & $0.583$ & $0.008$ & $0.052$ & $\times1.3$ \\
\ournameFixed & $91.3$ & $0.507$ & $0.024$ & $0.097$ & $\times2.5$ \\
\ournameFixed\,$+$WCC ($\lambda{=}0.1$) & $87.4$ & $0.295$ & $0.059$ & $0.210$ & $\times5.4$ \\
\ournameFixed\,$+$WCC ($\lambda{=}0.3$) & $87.0$ & $0.233$ & $0.070$ & $0.215$ & $\times5.5$ \\
\topksparse & $91.0$ & $0.691$ & $0.008$ & $0.050$ & $\times1.3$ \\
\ournameFixed\,$+$\,DP & $91.8$ & $0.393$ & $0.059$ & $0.131$ & $\times3.4$ \\
\bottomrule
\end{tabular}}
\end{table}

\begin{figure*}[!tbp]
    \centering
    \includegraphics[width=\textwidth]{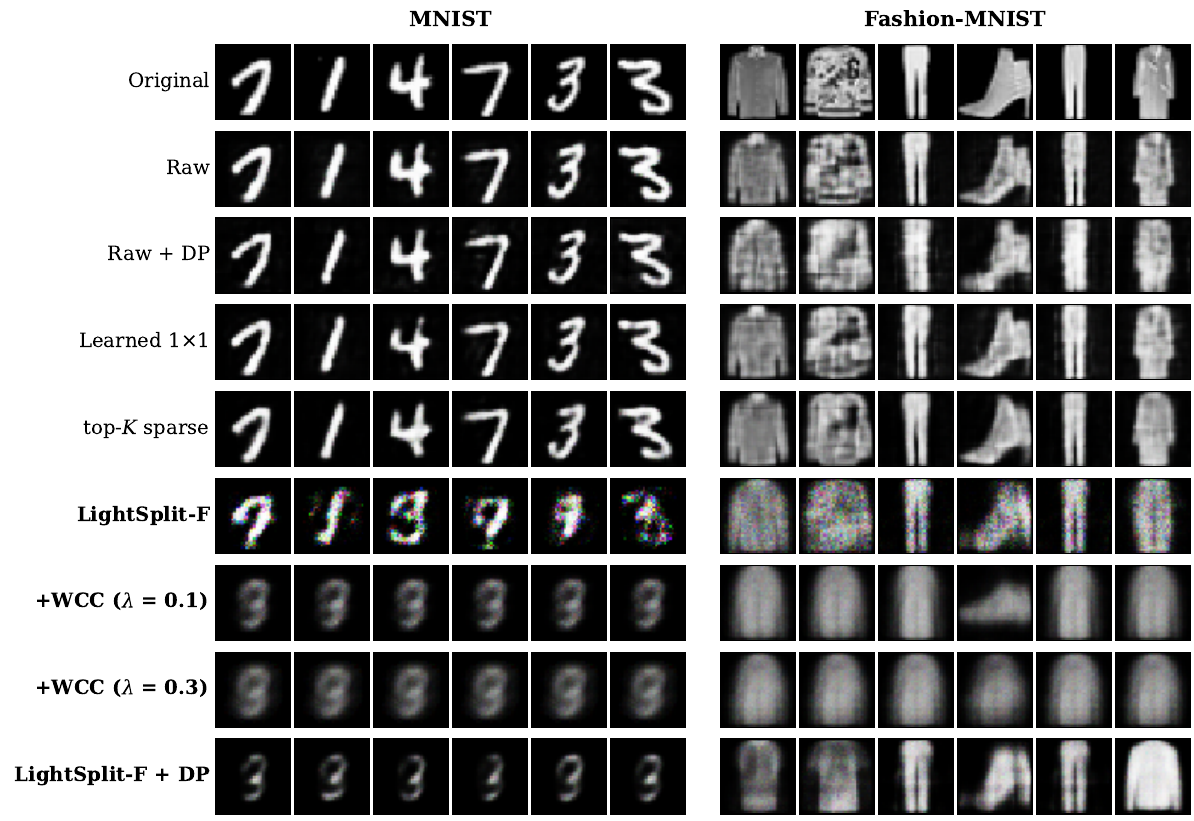}
    \caption{FSHA reconstructions at CR$\,{=}\,8\times$ on MNIST and Fashion-MNIST, one row per defense; the source samples are the same in every row.}
    \label{fig:fsha_recon}
\end{figure*}

\ourname assumes an honest-but-curious server (\sect\ref{sec:problem-threat}). This subsection steps outside that model with a \emph{fully-malicious} server that actively reshapes the cut representation. Table~\ref{tab:fsha_active} reports reconstruction quality and clean accuracy for every defense at CR\,${=}\,8\times$, and Fig.~\ref{fig:fsha_recon} shows the reconstructions: the digit or garment stays legible under \textsc{Raw}, the learned $1{\times}1$ conv and \topksparse, and becomes unrecognizable under \ournameFixed. WCC widens the gap, to ${\times}18$ on MNIST and ${\times}5.5$ on Fashion-MNIST at $\lambda_{\mathrm{WCC}}{=}0.3$. Adding DP to \ournameFixed lifts $\Delta_{\text{Raw}}$ above both \ournameFixed alone and \textsc{Raw}$+$DP (${\times}16.7$/${\times}3.4$) and gives the lowest SSIM on MNIST ($0.342$), with accuracy unchanged. The block-local variant (FSHA columns of Table~\ref{tab:block_privacy_fsha}) sits between \textsc{Raw} and the global projection. The $b{=}1$ tile at $16\times$ and $32\times$ reconstructs almost as well as \textsc{Raw} (SSIM $0.65$/$0.64$ against $0.68$ on MNIST, $0.56$/$0.54$ against $0.61$ on Fashion-MNIST); $b{=}2$ and $b{=}4$ land between $0.42$ and $0.66$ on MNIST and between $0.45$ and $0.55$ on Fashion-MNIST. The DP rows pull $b{=}2$ and $b{=}4$ lower ($0.51$/$0.44$ on MNIST, $0.46$/$0.30$ on Fashion-MNIST), while $b{=}1$ with DP rises above \textsc{Raw} ($0.81$/$0.62$).

\subsubsection{Projection-optimized attackers}
\label{app:proj_opt}
\begin{table}[!htbp]
\centering
\caption{Passive attackers that know the public projection $R$ (CIFAR-10): reconstruction SSIM by the signals the attacker fuses, the payload $z$, the lift-back $zR^{\top}$ and the returned cut-layer gradient $g$; higher $=$ stronger attack. The first row denies the attacker $R$.}
\label{tab:proj_opt}
\footnotesize
\setlength{\tabcolsep}{4pt}
\resizebox{\columnwidth}{!}{%
\begin{tabular}{lccc}
\toprule
\textbf{Attacker (signals fused)} & CR $8\times$ & CR $16\times$ & CR $32\times$ \\
\midrule
$z$ alone \ ($R$ denied, ablation) & $0.592$ & $0.549$ & $0.505$ \\
$z + zR^{\top}$ \ (activations $+$ lift-back) & $0.624$ & $0.569$ & $0.520$ \\
$z + g$ \ (activations $+$ gradient) & $0.557$ & $0.504$ & $0.469$ \\
$z + g + zR^{\top}$ \ (all signals) & $0.588$ & $0.536$ & $0.493$ \\
\bottomrule
\end{tabular}}
\end{table}

SDAR, FORA, and UnSplit already use the public $R$, but only through the lift-back: the server lifts every payload to $zR^{\top}$, the substitute encoder of SDAR and FORA passes through the same projection and lift-back, and each attack then inverts the lifted $d$-dimensional activation as if it were an ordinary smashed activation. None of them reads the $k$-dimensional payload $z$ itself or the gradient the server returns. The attackers here stay within the honest-but-curious model but are built around the public projection: a supervised decoder, trained on paired images and activations from an auxiliary set against the converged split model, reads $z$ directly and, row by row, also the analytic lift-back $zR^{\top}$ and the returned cut-layer gradient $g$ (App.~\ref{app:projopt}). Table~\ref{tab:proj_opt} and Fig.~\ref{fig:proj_opt}(a) report reconstruction SSIM for the four attackers at each compression ratio. The lift-back adds $+0.032$, $+0.020$, and $+0.015$ SSIM at CR $8/16/32\times$ over a decoder that reads $z$ alone. The returned cut-layer gradient $g$ adds nothing: fused with $z$ it lowers SSIM at every ratio, and the maximal attacker that concatenates $z$, $g$, and $zR^{\top}$ into one decoder stays below the lift-back decoder, and even below $z$ alone, at every ratio. The gradient is a function of $z$ in the same $k$-dimensional space; it carries task information but no new input coordinates.

\subsubsection{Energy discarded by the projection}
\label{app:energy_discarded}
\begin{figure}[!htbp]
    \centering
    \includegraphics[width=\columnwidth]{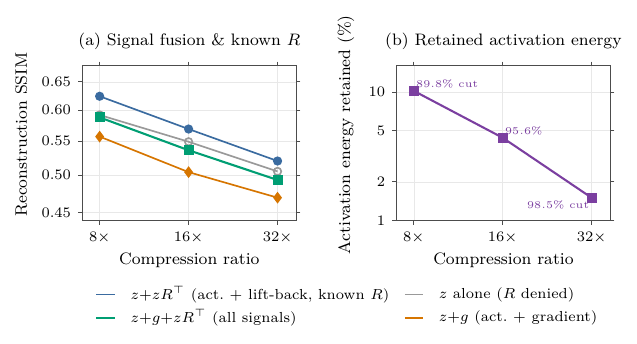}
    \caption{Attackers that know the projection $R$ (CIFAR-10, L7 cut) over the compression ratio: (a)~reconstruction SSIM, higher $=$ stronger attack, for the payload $z$ alone, $+$ lift-back $zR^{\top}$, $+$ cut-layer gradient $g$, and all three fused; (b)~share of the activation energy the projection retains, the labels give the discarded share.}
    \label{fig:proj_opt}
\end{figure}

Fig.~\ref{fig:proj_opt}(b) involves no attacker. For each activation $h$ at the cut we project and lift back, $\hat{h}=hRR^{\top}$, and measure the share of $\|h\|^2$ that does not come back, $\|h-\hat{h}\|^2/\|h\|^2$. That share grows with the compression ratio. It is the part of the activation that never leaves the client, so it stays out of reach of every attacker in Table~\ref{tab:proj_opt}, whatever signals the decoder reads.

\subsection{Differential Privacy}
\label{app:dp}

\begin{table*}[!tbp]
\centering
\scriptsize
\setlength{\tabcolsep}{3pt}
\caption{Utility cost of DP ($\sigma{=}0.02$): $\Delta$ best test accuracy (points) of each $+$DP model against its matching no-DP model, by dataset, distribution, head depth and head ownership; global \ourname cells pair the lift-backs (\ournameFixed\,/\,\ournameLearned); \textsc{n/a}: the tile does not divide the activation.}
\label{tab:dp_utility}\label{tab:dp_block}
\scaleTable{%
\begin{tabular}{l c c c cc cc cc cc}
\toprule
 & & & & \multicolumn{2}{c}{\textbf{CIFAR-10}} & \multicolumn{2}{c}{\textbf{CIFAR-100}} & \multicolumn{2}{c}{\textbf{FashionMNIST}} & \multicolumn{2}{c}{\textbf{GTSRB}} \\
\cmidrule(lr){5-6} \cmidrule(lr){7-8} \cmidrule(lr){9-10} \cmidrule(lr){11-12}
\textbf{Method} & \textbf{CR} & \textbf{Depth} & \textbf{Head} & $\alpha{=}0.1$ & $\alpha{\to}\infty$ & $\alpha{=}0.5$ & $\alpha{\to}\infty$ & $\alpha{=}0.1$ & $\alpha{\to}\infty$ & $\alpha{=}0.1$ & $\alpha{\to}\infty$ \\
\midrule
\rowcolor{rawgray} \textsc{Raw}$+$DP & $1\times$ & L1H & SCH & $-4.7$ & $-5.3$ & $-16.2$ & $-11.1$ & $-5.7$ & $-1.2$ & $-11.7$ & $-5.7$ \\
\rowcolor{rawgray}  &  & L1H & PCH & $+2.7$ & $-2.5$ & $-0.7$ & $-4.7$ & $+5.8$ & $+0.5$ & $-2.1$ & $-5.4$ \\
\rowcolor{rawgray}  &  & L2H & SCH & $-8.5$ & $-1.3$ & $-9.8$ & $-5.3$ & $-2.2$ & $+0.3$ & $-2.3$ & $-1.4$ \\
\rowcolor{rawgray}  &  & L2H & PCH & $-4.0$ & $-0.4$ & $-0.7$ & $-2.8$ & $+1.1$ & $+1.1$ & $+2.7$ & $-2.1$ \\
\midrule
\ourname$+$DP (F\,/\,L) & 8$\times$ & L1H & SCH & $-8.4$\,/\,$-5.2$ & $-0.9$\,/\,$-2.2$ & $-10.0$\,/\,$-10.4$ & $-1.1$\,/\,$-2.0$ & $-3.0$\,/\,$-4.6$ & $-0.6$\,/\,$-0.2$ & $-6.3$\,/\,$-8.2$ & $-1.8$\,/\,$-1.7$ \\
 &  & L1H & PCH & $+4.2$\,/\,$-1.3$ & $+3.3$\,/\,$-0.4$ & $+3.3$\,/\,$+2.2$ & $+1.9$\,/\,$-1.3$ & $-0.0$\,/\,$-3.7$ & $+1.1$\,/\,$+2.9$ & $+14.9$\,/\,$+6.1$ & $-2.8$\,/\,$-1.8$ \\
 &  & L2H & SCH & $-2.2$\,/\,$-2.2$ & $+0.4$\,/\,$-0.8$ & $-11.1$\,/\,$-10.1$ & $-1.4$\,/\,$-2.4$ & $-1.6$\,/\,$-2.4$ & $-0.2$\,/\,$-0.1$ & $-2.8$\,/\,$-4.9$ & $-0.8$\,/\,$-0.7$ \\
 &  & L2H & PCH & $+1.9$\,/\,$-2.2$ & $-1.6$\,/\,$+0.2$ & $+1.9$\,/\,$+0.8$ & $+1.2$\,/\,$+0.8$ & $-0.6$\,/\,$-0.9$ & $+0.3$\,/\,$+1.7$ & $+4.4$\,/\,$-3.7$ & $-1.5$\,/\,$-3.3$ \\
\addlinespace[1pt]
 & 16$\times$ & L1H & SCH & $-3.5$\,/\,$-12.0$ & $-2.5$\,/\,$-3.3$ & $-9.0$\,/\,$-13.2$ & $-3.8$\,/\,$-0.6$ & $-1.9$\,/\,$+0.3$ & $-0.1$\,/\,$-0.2$ & $-11.1$\,/\,$-4.9$ & $-3.4$\,/\,$-0.9$ \\
 &  & L1H & PCH & $+10.5$\,/\,$-3.2$ & $+2.6$\,/\,$-1.1$ & $+1.1$\,/\,$+1.4$ & $+0.4$\,/\,$-0.8$ & $-3.1$\,/\,$+0.2$ & $+1.2$\,/\,$+4.2$ & $+5.2$\,/\,$+15.7$ & $-3.2$\,/\,$-3.1$ \\
 &  & L2H & SCH & $-2.3$\,/\,$+0.9$ & $+0.4$\,/\,$-1.6$ & $-11.0$\,/\,$-8.1$ & $-1.6$\,/\,$-2.0$ & $-1.6$\,/\,$-4.1$ & $-0.6$\,/\,$-0.4$ & $-5.1$\,/\,$-1.3$ & $-0.8$\,/\,$-0.6$ \\
 &  & L2H & PCH & $-6.3$\,/\,$-0.1$ & $-0.5$\,/\,$+0.5$ & $+2.2$\,/\,$-0.3$ & $-0.0$\,/\,$+1.4$ & $-0.5$\,/\,$+13.0$ & $+0.8$\,/\,$+1.7$ & $-9.3$\,/\,$-9.8$ & $-2.4$\,/\,$-1.6$ \\
\addlinespace[1pt]
 & 32$\times$ & L1H & SCH & $-1.6$\,/\,$-7.5$ & $-2.5$\,/\,$-2.4$ & $-8.9$\,/\,$-10.0$ & $-1.7$\,/\,$-1.0$ & $+1.6$\,/\,$+0.2$ & $-1.0$\,/\,$-0.5$ & $-6.0$\,/\,$-5.1$ & $-5.2$\,/\,$-1.5$ \\
 &  & L1H & PCH & $+4.3$\,/\,$-0.7$ & $+1.3$\,/\,$-0.3$ & $+2.7$\,/\,$+1.0$ & $+1.3$\,/\,$+0.8$ & $-2.9$\,/\,$+15.2$ & $+1.0$\,/\,$+5.0$ & $-3.6$\,/\,$+11.5$ & $+0.2$\,/\,$+2.7$ \\
 &  & L2H & SCH & $-1.4$\,/\,$-3.0$ & $+0.2$\,/\,$-0.0$ & $-9.8$\,/\,$-10.3$ & $-1.6$\,/\,$-0.7$ & $+0.5$\,/\,$-4.5$ & $-0.0$\,/\,$-0.3$ & $-5.0$\,/\,$-3.0$ & $-2.5$\,/\,$-1.4$ \\
 &  & L2H & PCH & $-1.0$\,/\,$+1.8$ & $+0.2$\,/\,$+1.3$ & $+4.9$\,/\,$+1.1$ & $+0.7$\,/\,$+2.1$ & $-2.2$\,/\,$+8.1$ & $-0.2$\,/\,$+2.0$ & $-0.5$\,/\,$+0.5$ & $-5.3$\,/\,$-3.4$ \\
\midrule
$b{=}1$$+$DP & 8$\times$ & L1H & SCH & $-4.6$ & $-2.7$ & $-4.2$ & $-1.7$ & $-1.1$ & $-0.4$ & $-5.7$ & $-1.5$ \\
 &  & L1H & PCH & $+1.7$ & $-1.6$ & $+1.1$ & $-1.6$ & $+1.0$ & $-0.2$ & $+12.3$ & $-1.0$ \\
 &  & L2H & SCH & $-0.2$ & $-0.7$ & $+0.1$ & $-1.1$ & $+0.9$ & $+0.2$ & $-1.7$ & $-0.6$ \\
 &  & L2H & PCH & $-1.6$ & $-1.0$ & $+1.4$ & $+0.2$ & $-3.1$ & $+0.2$ & $+18.1$ & $-1.0$ \\
\addlinespace[1pt]
 & 16$\times$ & L1H & SCH & $+0.7$ & $-3.1$ & $-2.7$ & $-3.0$ & $+0.9$ & $-0.9$ & $-3.6$ & $-1.5$ \\
 &  & L1H & PCH & $+3.9$ & $-2.1$ & $-0.3$ & $-2.5$ & $+1.3$ & $+0.3$ & $+31.8$ & $-1.4$ \\
 &  & L2H & SCH & $-5.8$ & $-1.2$ & $-1.2$ & $+0.9$ & $-2.5$ & $-0.2$ & $-1.9$ & $-0.6$ \\
 &  & L2H & PCH & $-6.6$ & $-1.7$ & $+1.3$ & $-0.4$ & $+3.1$ & $-0.1$ & $+14.8$ & $-1.6$ \\
\addlinespace[1pt]
 & 32$\times$ & L1H & SCH & $+1.6$ & $-2.4$ & $-1.0$ & $-1.8$ & $-1.4$ & $-0.3$ & $-2.5$ & $-2.0$ \\
 &  & L1H & PCH & $-3.0$ & $-0.7$ & $-2.0$ & $-3.1$ & $+1.7$ & $-0.5$ & $+2.3$ & $-3.4$ \\
 &  & L2H & SCH & $+1.4$ & $-0.8$ & $+0.6$ & $-0.2$ & $+4.0$ & $-0.4$ & $-1.3$ & $-0.8$ \\
 &  & L2H & PCH & $-0.8$ & $-1.2$ & $-1.3$ & $-1.3$ & $-6.0$ & $-0.4$ & $+5.4$ & $-1.7$ \\
\midrule
$b{=}2$$+$DP & 8$\times$ & L1H & SCH & $-3.8$ & $-2.4$ & $-2.2$ & $-2.9$ & $-2.5$ & $-0.6$ & $-7.7$ & $-2.7$ \\
 &  & L1H & PCH & $+15.5$ & $+0.3$ & $+1.4$ & $-0.5$ & $+1.3$ & $+0.6$ & $+23.9$ & $-2.1$ \\
 &  & L2H & SCH & $-2.5$ & $-0.5$ & $-0.1$ & $-0.6$ & $-0.2$ & $-0.0$ & $-2.2$ & $-0.8$ \\
 &  & L2H & PCH & $-4.8$ & $-0.5$ & $-1.0$ & $+0.8$ & $-2.1$ & $+0.0$ & $+7.4$ & $-1.2$ \\
\addlinespace[1pt]
 & 16$\times$ & L1H & SCH & $+2.0$ & $-2.5$ & $-2.3$ & $-2.1$ & $-2.1$ & $-0.6$ & $-3.0$ & $-3.0$ \\
 &  & L1H & PCH & $+3.6$ & $-0.3$ & $+0.7$ & $-1.1$ & $-2.6$ & $-0.2$ & $+6.2$ & $-1.3$ \\
 &  & L2H & SCH & $+0.2$ & $-0.2$ & $-1.2$ & $-0.8$ & $+4.0$ & $+0.0$ & $-1.3$ & $-0.4$ \\
 &  & L2H & PCH & $+8.3$ & $-0.5$ & $+1.4$ & $+1.1$ & $-0.8$ & $-1.1$ & $+11.5$ & $-1.0$ \\
\addlinespace[1pt]
 & 32$\times$ & L1H & SCH & $-0.0$ & $-1.0$ & $-1.7$ & $-1.4$ & $+0.6$ & $-0.6$ & $-1.4$ & $-2.5$ \\
 &  & L1H & PCH & $-5.1$ & $-1.3$ & $+1.1$ & $-0.2$ & $-4.9$ & $-0.4$ & $-10.1$ & $-2.5$ \\
 &  & L2H & SCH & $-4.2$ & $-0.1$ & $+0.6$ & $-0.5$ & $+2.0$ & $-0.2$ & $-1.1$ & $-0.7$ \\
 &  & L2H & PCH & $+2.3$ & $-1.3$ & $-2.1$ & $+1.1$ & $-0.6$ & $+0.2$ & $+9.1$ & $-1.5$ \\
\midrule
$b{=}4$$+$DP & 8$\times$ & L1H & SCH & $-2.6$ & $-2.6$ & $-1.1$ & $-1.5$ & $+0.9$ & $-0.8$ & \textsc{n/a} & \textsc{n/a} \\
 &  & L1H & PCH & $+6.5$ & $-2.7$ & $+4.1$ & $+0.1$ & $-0.7$ & $+0.5$ & \textsc{n/a} & \textsc{n/a} \\
 &  & L2H & SCH & $-3.5$ & $-0.1$ & $-1.7$ & $+0.1$ & $-0.4$ & $-0.1$ & \textsc{n/a} & \textsc{n/a} \\
 &  & L2H & PCH & $-10.0$ & $+0.1$ & $+1.8$ & $+0.5$ & $+0.6$ & $+0.5$ & \textsc{n/a} & \textsc{n/a} \\
\addlinespace[1pt]
 & 16$\times$ & L1H & SCH & $-0.7$ & $-0.4$ & $-2.4$ & $-2.1$ & $-0.6$ & $-0.8$ & \textsc{n/a} & \textsc{n/a} \\
 &  & L1H & PCH & $-7.0$ & $-0.7$ & $+1.1$ & $+0.3$ & $+3.1$ & $+0.9$ & \textsc{n/a} & \textsc{n/a} \\
 &  & L2H & SCH & $-2.9$ & $-0.3$ & $+1.6$ & $-0.5$ & $+0.9$ & $-0.4$ & \textsc{n/a} & \textsc{n/a} \\
 &  & L2H & PCH & $+10.5$ & $-0.8$ & $+5.9$ & $+0.5$ & $-0.7$ & $-0.1$ & \textsc{n/a} & \textsc{n/a} \\
\addlinespace[1pt]
 & 32$\times$ & L1H & SCH & $+1.5$ & $-0.9$ & $-0.3$ & $-1.8$ & $-0.2$ & $-0.5$ & \textsc{n/a} & \textsc{n/a} \\
 &  & L1H & PCH & $-4.4$ & $+0.2$ & $+3.0$ & $+0.5$ & $+1.6$ & $+0.1$ & \textsc{n/a} & \textsc{n/a} \\
 &  & L2H & SCH & $+2.3$ & $-0.4$ & $+0.2$ & $+1.1$ & $+2.3$ & $+0.0$ & \textsc{n/a} & \textsc{n/a} \\
 &  & L2H & PCH & $-7.6$ & $+0.2$ & $+3.9$ & $+1.8$ & $+3.6$ & $-0.4$ & \textsc{n/a} & \textsc{n/a} \\
\bottomrule
\end{tabular}%
}\\[2pt]
\parbox{0.95\textwidth}{\footnotesize\raggedright The no-DP references are the \textsc{Raw}, \ournameFixed and \ournameLearned rows of Table~\ref{tab:full_sweep_baseline_appendix} for the global rows and the matching cells of Table~\ref{tab:block_util_F} for the block-local rows; the absolute $+$DP accuracy is the reference cell plus $\Delta$.}
\end{table*}

\ourname composes with DP at the cut (mechanism and operating point in App.~\ref{app:setup_dp}). The server's lift-back is post-processing of the privatized payload, so it cannot weaken the guarantee. This subsection reports the utility cost; the DP rows of each attack grid in App.~\ref{app:attacks} report the reconstruction side.

Table~\ref{tab:dp_utility} extends the noise sweep of App.~\ref{app:setup_dp} to every dataset, head, and distribution: \ournameFixed$+$DP and \ournameLearned$+$DP at CR\,$8/16/32\times$ beside \textsc{Raw}$+$DP, with the DP cost $\Delta$ against the matching no-DP model. On the deep shared head the cost stays within ${\sim}2$ points on the stable datasets and does not grow with compression (\ournameFixed on GTSRB IID $95.6/94.0/90.6\%$ at $8/16/32\times$); the large penalties fall mainly on the non-IID CIFAR-100 cells, which are fragile without DP already. The block-local projection (block rows of Table~\ref{tab:dp_block}) behaves the same: across $b\,{\in}\,\{1,2,4\}$ the cost stays within ${\sim}1$ point. Fig.~\ref{fig:dp_lowcr}(b) then raises $\sigma$: accuracy degrades gracefully, and at the same noise level the noise costs the compressed cut less accuracy than it costs the raw cut. The noise energy grows with the dimension as $\sqrt{\text{dim}}$, so the $512$-d payload at $8\times$ absorbs ${\approx}2.8\times$ less noise than the $4096$-d raw cut at the same $\sigma$. Below $\sigma{\approx}0.1$ the raw cut wins because it pays no compression loss; above it the projection holds up better, $68.0\%$ against $52.6\%$ at $\sigma{=}0.25$. The operating point $\sigma{=}0.02$ sits well inside the regime where the noise is nearly free.

\subsection{Compression range and cut width}
\label{app:lowcr_results}
This subsection separates the two knobs that set the payload (App.~\ref{app:setup_lowcr}): the compression ratio, swept over its full range on utility, and the cut width, swept at a fixed depth on both utility and reconstruction privacy.

\begin{figure}[!htbp]
    \centering
    \includegraphics[width=\columnwidth]{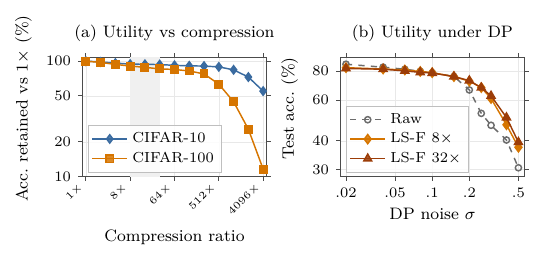}
    \caption{(a)~\ournameFixed test accuracy as \% of the uncompressed model over CR\,$1$--$4096\times$, the $8$--$32\times$ band shaded; (b)~test accuracy under DP (CIFAR-10, deep head) over the noise $\sigma$ for \textsc{Raw} and \ournameFixed at $8\times$ and $32\times$; $\sigma{=}0.02$ is the leftmost point.}
    \label{fig:dp_lowcr}
\end{figure}

\paragraph{Lower compression ratios}
Fig.~\ref{fig:dp_lowcr}(a) plots \ournameFixed test accuracy as a share of the uncompressed model over the full range CR\,$1$--$4096\times$, with the $8$--$32\times$ band shaded. Compression costs more on CIFAR-100 than on CIFAR-10 at every ratio, so the gentle end of the curve ($2\times$, $4\times$) is where a label-rich, sample-poor dataset buys accuracy back at a proportionally larger payload. The band itself sits on the flat part of both curves: inside it each doubling of the ratio costs CIFAR-10 under a point, and the curves only fall away once the payload shrinks to a handful of coordinates. CIFAR-100 stays above half of its uncompressed accuracy down to $k{=}8$ ($512\times$), and CIFAR-10 does so at every ratio.

\begin{figure}[!htbp]
    \centering
    \includegraphics[width=\columnwidth]{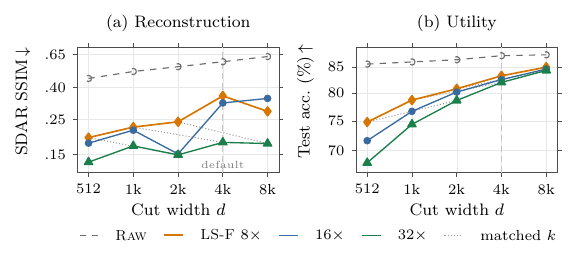}
    \caption{Cut width $d$ at a fixed cut depth (CIFAR-10, default head, IID); the dashed line is \textsc{Raw}, the shaded band the default width $d{=}4096$. (a)~SDAR reconstruction SSIM, lower $=$ more private, strongest attacker run per cell; (b)~test accuracy, mean over seeds. Dotted connectors join matched-$k$ cells, the same payload crossing a $4\times$ wider cut (CR\,$8\times$ at $d$ and CR\,$32\times$ at $4d$).}
    \label{fig:cutwidth}
\end{figure}

\paragraph{Cut width}
Fig.~\ref{fig:cutwidth} varies the cut width $d$ at a fixed cut depth (App.~\ref{app:setup_lowcr}). Each privacy cell is the strongest attacker run over the seeds whose victim trained normally. A narrower cut leaks less even without a defense: \textsc{Raw} SSIM rises from $0.457$ at $d{=}512$ to $0.630$ at $d{=}8192$, so each width has its own baseline. The default width $d{=}4096$ reproduces the default cell: $87.30\%/83.30\%$ \textsc{Raw}/CR\,$8\times$ accuracy against $87.24\%/83.50\%$ in Table~\ref{tab:full_sweep_baseline_appendix}. The projection removes information at every width. The strongest projected cell in the sweep ($0.354$, $d{=}4096$, CR\,$8\times$) stays below the weakest \textsc{Raw} cell ($0.457$, $d{=}512$). At the default width, SSIM falls with the ratio: $0.354$, $0.319$ and $0.179$ at CR\,$8/16/32\times$. We do not rank the projected cells of different widths against each other: at the narrow cuts the SDAR attacker converged in a minority of runs, and the differences there are within that noise. Utility favors a wider cut. The dotted connectors in Fig.~\ref{fig:cutwidth} join cells with the same payload $k$ across a $4\times$ wider cut: every such pair gains accuracy, $+3.9$ points at $k{=}64$, $+3.3$ at $k{=}128$ and $+3.5$ at $k{=}256$, and the gap to that width's own \textsc{Raw} shrinks as well. A wider cut costs the client. The last stage of its head grows with $d$: at $d{=}8192$ it holds about three times the parameters of the default, for the same payload and the same bandwidth. We keep $d{=}4096$ for that reason.

\subsection{Backdoor detection: signal sweeps and F\textsubscript{1} table}
\label{app:backdoor}

\begin{table}[!htbp]
\centering
\caption{Label-free backdoor detection F$_1$ on the backbone output $\bar{\mathbf{r}}$ (CR\,${=}\,16\times$); one column covers \ournameFixed and \ournameLearned, whose $3\sigma$ crossings coincide; $0.00$: no client crosses $3\sigma$.}
\label{tab:bd-ours}
\scaleTable{%
\begin{tabular}{l l c}
\toprule
\textbf{Setting} & \textbf{Value(s)} & \textbf{F$_1$} \\
\midrule
Baseline & $\alpha{\to}\infty$ (IID)$,\,n{=}10,\,\mu{=}0.1,\,p{=}0.3$ & $0.89$ \\
\midrule
$\alpha$ (Dirichlet) & $0.1$ / $1$ / $10$        & $0.00$ / $0.00$ / $1.00$ \\
$n$ (clients)        & $5$ / $20$ / $50$         & $1.00$ / $0.67$ / $1.00$ \\
$\mu$ (mal. ratio)   & $0.2$ / $0.3$             & $1.00$ / $1.00$ \\
$p$ (poison rate)    & $0.1$ / $0.2$ / $0.5$ / $0.8$ & $1.00$ / $1.00$ / $0.67$ / $0.67$ \\
\bottomrule
\end{tabular}%
}\\[2pt]
\parbox{0.95\columnwidth}{\footnotesize\raggedright The baseline row is the mean over seeds; each ablation row changes one parameter from the baseline and is a single run. In the $0.00$ cells the malicious client still ranks most anomalous.}
\end{table}

\begin{figure}[!htbp]
    \centering
    \includegraphics[width=0.8\columnwidth]{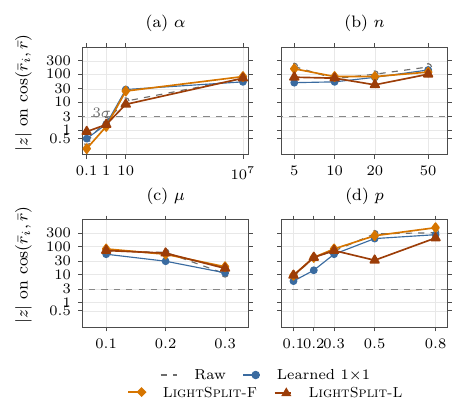}
    \caption{Backbone-output cosine signal $\cos(\bar{r}_i,\bar{r})$. Except varied parameters, the remaining parameters are set to the defaults of App.~\mbox{\ref{app:backdoor_setup}}.}%
    \label{fig:backdoor_r}
\end{figure}

\begin{figure}[!htbp]
    \centering
    \includegraphics[width=0.8\columnwidth]{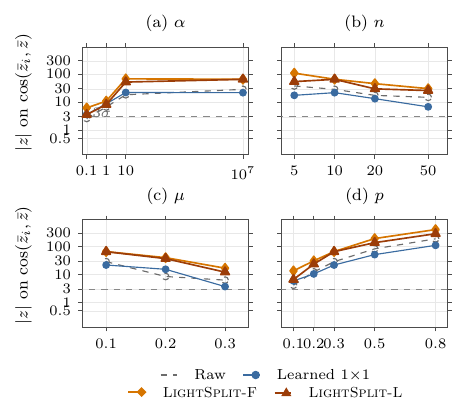}
    \caption{\textbf{Server-input cosine signal $\cos(\bar{z}_i,\bar{z})$.}
    Malicious-client $|$MAD-z-score$|$ vs.\ each ablation axis; in every panel the swept parameter is varied while the rest are held at the baseline of Table~\mbox{\ref{tab:bd-ours}}; the dashed line at $3$ is the flagging threshold.}
    \label{fig:backdoor_z}
\end{figure}

Fig.~\ref{fig:backdoor_r} reports the cosine-to-consensus anomaly score on the \emph{server-output} backbone activation $\bar{\mathbf{r}}_i$. For completeness, Fig.~\ref{fig:backdoor_z} shows the same sweep on the \emph{server-input} smashed payload $\bar{\mathbf{z}}_i$. Tab.~\ref{tab:bd-ours} reports the detection F$_1$ at the conventional $3\sigma$ flagging threshold on the backbone-output scores of Fig.~\ref{fig:backdoor_r}. The results are consistent with the main-text claim: both \textsc{\ournameFixed} and \textsc{\ournameLearned} preserve, and on $\bar{\mathbf{z}}_i$ amplify, the malicious-client signature relative to \textsc{Raw}. Across the operative regime ($\alpha\,{\geq}\,10$, $p\,{\leq}\,0.3$, every $\mu$, and every $n$ we test) threshold F$_1$ stays in $[0.67,\,1.00]$, reaching $1.0$ in most cells. The cells where F$_1\,{=}\,0$ are runs in which the malicious client is still the most-anomalous in the field but its $|$MAD-$z|$ does not cross the strict $3\sigma$ line; the underlying signal, plotted in the figures, remains intact.

\subsection{Ablations}
\label{app:ablations}
\label{app:additional_ablations}
\label{app:mlp_hidden_ablation}

\begin{table}[!tbp]
\centering
\small
\setlength{\tabcolsep}{3pt}
\caption{MLP-liftback hidden-dimension ablation for \textsc{\ournameLearned} across the L1H/L2H $\times$ SCH/PCH cross at CR$\,{=}\,8\times$, IID, $n{=}10$ (App.~\ref{app:mlp_hidden_ablation}). Acc is best test accuracy (\%), $\Delta$ is against matched \textsc{Raw}; a dash marks a width not run for that dataset.}
\label{tab:mlp-hidden-ablation}
\resizebox{\columnwidth}{!}{%
\begin{tabular}{l l r *{6}{r}}
\toprule
\textbf{Head} & & \textbf{Raw} & \multicolumn{6}{c}{\textbf{Hidden width} $\boldsymbol{h}$} \\
\cmidrule(lr){4-9}
 & & & $64$ & $128$ & $256$ & $512$ & $1024$ & $2048$ \\
\midrule
\multicolumn{9}{c}{\textbf{CIFAR-10}} \\
\midrule
 L2H+SCH  & Acc & 87.24 & -- & -- & 85.76 & 85.91 & 84.36 & 83.93 \\
          & $\Delta$ &  & -- & -- & -1.48 & -1.33 & -2.88 & -3.31 \\
\addlinespace[1pt]
 L1H+SCH  & Acc & 86.86 & -- & -- & 76.69 & 78.68 & 78.07 & 77.89 \\
          & $\Delta$ &  & -- & -- & -10.17 & -8.18 & -8.79 & -8.97 \\
\addlinespace[1pt]
 L2H+PCH  & Acc & 78.31 & -- & -- & 70.34 & 70.59 & 70.90 & 70.70 \\
          & $\Delta$ &  & -- & -- & -7.97 & -7.72 & -7.41 & -7.61 \\
\addlinespace[1pt]
 L1H+PCH  & Acc & 81.28 & -- & -- & 71.82 & 73.00 & 71.77 & 72.25 \\
          & $\Delta$ &  & -- & -- & -9.46 & -8.28 & -9.51 & -9.03 \\
\midrule
\multicolumn{9}{c}{\textbf{CIFAR-100}} \\
\midrule
 L2H+SCH  & Acc & 59.47 & -- & -- & 52.23 & 54.10 & 54.38 & 55.34 \\
          & $\Delta$ &  & -- & -- & -7.24 & -5.37 & -5.09 & -4.13 \\
\addlinespace[1pt]
 L1H+SCH  & Acc & 59.56 & -- & -- & 44.45 & 44.82 & 44.82 & 45.52 \\
          & $\Delta$ &  & -- & -- & -15.11 & -14.74 & -14.74 & -14.04 \\
\addlinespace[1pt]
 L2H+PCH  & Acc & 47.09 & -- & -- & 38.24 & 38.15 & 38.44 & 37.58 \\
          & $\Delta$ &  & -- & -- & -8.85 & -8.94 & -8.65 & -9.51 \\
\addlinespace[1pt]
 L1H+PCH  & Acc & 49.95 & -- & -- & 39.72 & 40.96 & 41.04 & 40.20 \\
          & $\Delta$ &  & -- & -- & -10.23 & -8.99 & -8.91 & -9.75 \\
\midrule
\multicolumn{9}{c}{\textbf{FashionMNIST}} \\
\midrule
 L2H+SCH  & Acc & 91.91 & 90.91 & 91.22 & 91.16 & 91.21 & 91.16 & -- \\
          & $\Delta$ &  & -1.00 & -0.69 & -0.75 & -0.70 & -0.75 & -- \\
\addlinespace[1pt]
 L1H+SCH  & Acc & 91.75 & 90.29 & 90.14 & 90.10 & 90.29 & 90.09 & -- \\
          & $\Delta$ &  & -1.46 & -1.61 & -1.65 & -1.46 & -1.66 & -- \\
\addlinespace[1pt]
 L2H+PCH  & Acc & 88.75 & 85.34 & 84.79 & 85.58 & 85.98 & 86.00 & -- \\
          & $\Delta$ &  & -3.41 & -3.96 & -3.17 & -2.77 & -2.75 & -- \\
\addlinespace[1pt]
 L1H+PCH  & Acc & 88.81 & 84.99 & 84.56 & 84.85 & 84.36 & 84.25 & -- \\
          & $\Delta$ &  & -3.82 & -4.25 & -3.96 & -4.45 & -4.56 & -- \\
\midrule
\multicolumn{9}{c}{\textbf{GTSRB}} \\
\midrule
 L2H+SCH  & Acc & 98.82 & 95.67 & 95.88 & 95.59 & 95.49 & 95.27 & -- \\
          & $\Delta$ &  & -3.15 & -2.94 & -3.23 & -3.33 & -3.55 & -- \\
\addlinespace[1pt]
 L1H+SCH  & Acc & 98.05 & 89.11 & 89.35 & 89.92 & 89.63 & 89.15 & -- \\
          & $\Delta$ &  & -8.94 & -8.70 & -8.13 & -8.42 & -8.90 & -- \\
\addlinespace[1pt]
 L2H+PCH  & Acc & 97.52 & 93.52 & 93.00 & 92.76 & 92.68 & 92.72 & -- \\
          & $\Delta$ &  & -4.00 & -4.52 & -4.77 & -4.84 & -4.80 & -- \\
\addlinespace[1pt]
 L1H+PCH  & Acc & 96.12 & 87.36 & 86.96 & 86.60 & 86.06 & 87.60 & -- \\
          & $\Delta$ &  & -8.76 & -9.16 & -9.52 & -10.06 & -8.52 & -- \\
\bottomrule
\end{tabular}}
\end{table}

\begin{table}[!htbp]
\centering
\small
\caption{Client-count scaling: best test accuracy (\%) at CR$\,{=}\,8\times$, L2H-SCH, IID ($\alpha{\to}\infty$).}
\label{tab:client_scaling_ablation}
\scaleTable{%
\begin{tabular}{@{}llccccc@{}}
\toprule
\textbf{Dataset} & \textbf{Method}
& \textbf{$n{=}10$}
& \textbf{$n{=}25$}
& \textbf{$n{=}50$}
& \textbf{$n{=}100$}
& \textbf{$\Delta_{10\to100}$} \\
\midrule
CIFAR-10 & \textsc{Raw}        & 87.24 & 85.89 & 88.31 & 88.02 & $+0.78$ \\
         & \textsc{\ournameFixed} & 83.50 & 84.69 & 83.76 & 83.31 & $-0.19$ \\
         & \textsc{\ournameLearned} & 83.42 & 83.80 & 83.82 & 83.40 & $-0.02$ \\
\midrule
CIFAR-100 & \textsc{Raw}        & 59.47 & 54.85 & 54.21 & 54.39 & $-5.08$ \\
          & \textsc{\ournameFixed} & 52.83 & 50.56 & 49.52 & 49.50 & $-3.33$ \\
          & \textsc{\ournameLearned} & 56.49 & 55.51 & 54.61 & 53.89 & $-2.60$ \\

\bottomrule
\end{tabular}%
}
\end{table}

\begin{table}[!tbp]
\centering
\caption{Effect of WCC regularization on accuracy across SCH, PCH and cut layer levels (CIFAR-10 IID, $\alpha{\to}\infty$, $n{=}10$, CR$\,{=}\,8\times$). Each cell reports the relative accuracy loss vs the matching \textsc{Raw} baseline. The $b{=}2$ rows add the block-local projection (App.~\ref{app:block}).}
\label{tab:wcc_ablation}
\begin{adjustbox}{max width=\columnwidth,center}
\begin{tabular}{l c c cccc}
\toprule
& & & \multicolumn{4}{c}{$\lambda_{\mathrm{WCC}}$} \\
\cmidrule(lr){4-7}
\textbf{Method} & \textbf{Configuration} & \textbf{Proj.}
  & $0$ & $10^{-3}$ & $10^{-2}$ & $10^{-1}$ \\
\midrule
\ournameFixed & L1H-SCH & global & $-9.5\%$  & $-13.1\%$ & $-10.9\%$ & $-16.8\%$ \\
 &  & $b{=}2$ & $-3.8\%$ & $-4.4\%$ & $-7.5\%$ & $-12.7\%$ \\
\addlinespace[2pt]
\ournameFixed & L2H-SCH & global & $-5.0\%$ & $-6.0\%$ & $-7.5\%$ & $-5.3\%$ \\
 &  & $b{=}2$ & $-0.7\%$ & $-2.4\%$ & $-3.5\%$ & $-10.5\%$ \\
\addlinespace[2pt]
\ournameFixed & L1H-PCH & global & $-6.4\%$  & $-7.9\%$  & $-16.7\%$ & $-22.5\%$ \\
 &  & $b{=}2$ & $-4.0\%$ & $-6.5\%$ & $-28.9\%$ & $-57.6\%$ \\
\addlinespace[2pt]
\ournameFixed & L2H-PCH & global & $-9.1\%$  & $-16.4\%$ & $-33.0\%$ & $-46.9\%$ \\
 &  & $b{=}2$ & $+0.2\%$ & $-11.5\%$ & $-58.2\%$ & $-73.0\%$ \\
\midrule
\ournameLearned & L1H-SCH & global & $-10.9\%$ & $-12.7\%$ & $-14.9\%$ & $-18.3\%$ \\
 &  & $b{=}2$ & $-8.1\%$ & $-8.7\%$ & $-10.6\%$ & $-12.5\%$ \\
\addlinespace[2pt]
\ournameLearned & L2H-SCH & global & $-4.6\%$ & $-5.8\%$ & $-5.7\%$ & $-5.5\%$ \\
 &  & $b{=}2$ & $-2.1\%$ & $-1.7\%$ & $-2.7\%$ & $-8.7\%$ \\
\addlinespace[2pt]
\ournameLearned & L1H-PCH & global & $-11.7\%$ & $-16.1\%$ & $-26.3\%$ & $-48.2\%$ \\
 &  & $b{=}2$ & $-13.0\%$ & $-39.5\%$ & $-67.2\%$ & $-75.8\%$ \\
\addlinespace[2pt]
\ournameLearned & L2H-PCH & global & $-10.4\%$ & $-43.2\%$ & $-70.0\%$ & $-75.2\%$ \\
 &  & $b{=}2$ & $-5.2\%$ & $-33.7\%$ & $-70.2\%$ & $-79.5\%$ \\
\bottomrule
\end{tabular}
\end{adjustbox}\\[2pt]
\parbox{0.95\columnwidth}{\footnotesize\raggedright
The $b{=}2$ rows come from a separate training grid with different seeds, so its \textsc{Raw} means differ from the global rows' by up to $2.5$ points; each row reads against the \textsc{Raw} of its own grid.}
\end{table}

\begin{figure}[!htbp]
    \centering
    \includegraphics[width=0.6\columnwidth]{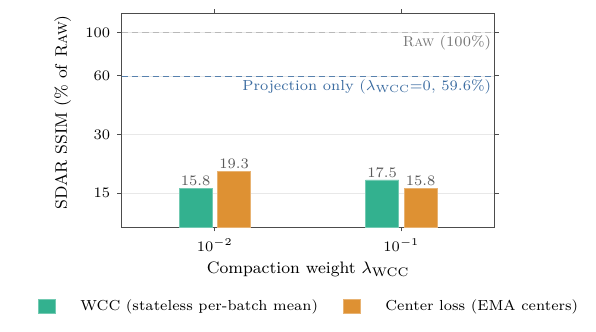}
    \caption{Center loss and WCC at matched compaction weight (CIFAR-10, L7 cut, CR$\,8\times$): SDAR SSIM as \% of \textsc{Raw}, lower $=$ more private; the dashed lines mark \textsc{Raw} and the projection alone.}
    \label{fig:center_wcc}
\end{figure}

\begin{figure}[!htbp]
    \centering
    \includegraphics[width=0.65\columnwidth]{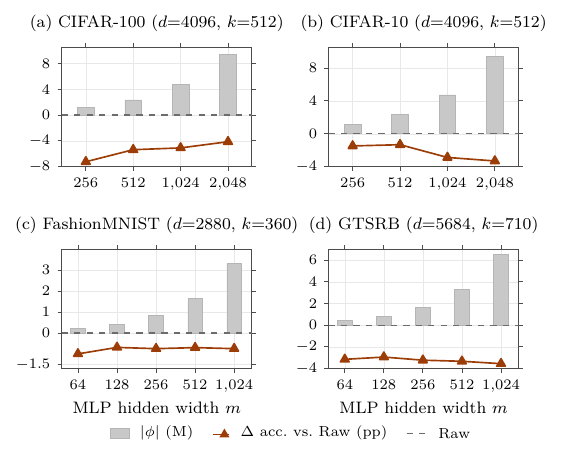}
    \caption{Impact of MLP lift-back size on accuracy gap (L2H-SCH, CR$=8$). Bars: $|\phi|$ in M; line: $\Delta$ acc.\ vs.\ Raw (dashed).}
    \label{fig:mlp_ablation}
\end{figure}

We run four ablation studies: the effect of the WCC regularization strength $\lambda_{\mathrm{WCC}}$ on task accuracy (Table~\ref{tab:wcc_ablation}), the center loss in place of WCC (Fig.~\ref{fig:center_wcc}), client-count scaling (Table~\ref{tab:client_scaling_ablation}), and the \textsc{\ournameLearned} lift-back hidden width $m$ (Fig.~\ref{fig:mlp_ablation}; full per-head-config table in Table~\ref{tab:mlp-hidden-ablation}).

\paragraph{WCC strength}
The results in Tab.~\ref{tab:wcc_ablation} shows the impact of varying $\lambda_{\mathrm{WCC}}$. We choose values for $\lambda_{\mathrm{WCC}}$ from $0$ up to $10^{-1}$ on CIFAR-10 IID at CR$\,{=}\,8\times$, across all four head configurations
(parameter counts in Table~\ref{tab:model_partitions}), measuring relative
accuracy loss against the matching \textsc{Raw} baseline.

Only L2H-SCH achieves consistent accuracy values between roughly $5$ to $8\%$ of \textsc{Raw}, independent of the choice for $\lambda_{\mathrm{WCC}}$, for both \textsc{\ournameFixed}
and \textsc{\ournameLearned}.

The other configuration L1H-SCH degrades steadily,
reaching nearly a fifth below \textsc{Raw} at the strongest setting. Also, PCH
variants reduce the accuracy. \textsc{\ournameFixed} L2H-PCH starts at a $9\%$ loss with no regularization and falls to nearly half of \textsc{Raw}
once $\lambda_{\mathrm{WCC}}{=}10^{-1}$. Notably, for L2H-PCH, accuracy significantly drops compared to the same setting for L2H-SCH, showing the benefit of the inter-client collaboration.

\begin{figure}[!htbp]
    \centering
    \includegraphics[width=0.85\columnwidth]{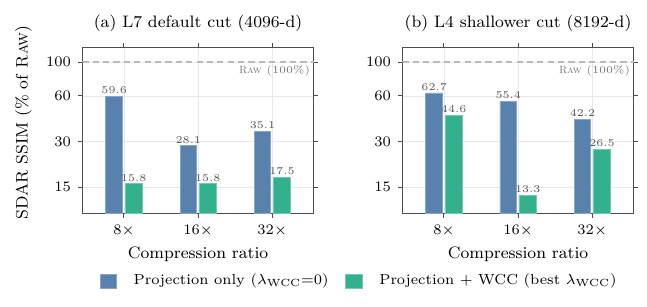}
    \caption{Privacy-gain decomposition (CIFAR-10): SDAR SSIM as \% of \textsc{Raw} (dashed, $100\%$), lower $=$ more private, at the (a)~L7 and (b)~L4 cuts, for the projection alone ($\lambda_{\mathrm{WCC}}{=}0$) and with the privacy-optimal WCC weight.}
    \label{fig:wcc_decomp}
\end{figure}

Our takeaway matches the operating point of \sect\ref{sec:method-wcc} and the guidance of \sect\ref{sec:deployment}: with a shared head, $\lambda_{\mathrm{WCC}}{\approx}10^{-3}$ is a safe default, and L2H-SCH holds its accuracy at every setting we sweep.
\emph{Projection does the bulk; WCC adds further} (Fig.~\ref{fig:wcc_decomp}). With $\lambda_{\mathrm{WCC}}{=}0$ the projection alone lowers SDAR SSIM from $0.57$ to $0.34/0.16/0.20$ at the L7 cut and from
$0.83$ to $0.52/0.46/0.35$ at L4 (CR $8/16/32\times$). This gain is information removal. Whatever the attacker trains, it sees only the $k$ coordinates of $z$. The best lift-back a server that knows $R$ can form is $RR^{\top}z$; the $d-k$ discarded coordinates lie in the null space of $R^{\top}$, out of reach of any decoder. App.~\ref{app:proj_opt} confirms this
empirically: a supervised decoder trained on paired images and payloads gains at most $+0.032/{+}0.020/{+}0.015$ SSIM at CR $8/16/32\times$ from the exact lift-back, and the returned gradient adds nothing. WCC lowers the attack further: with any $\lambda_{\mathrm{WCC}}\geq10^{-3}$ the L7 attacker stays at $0.09$--$0.13$ SSIM at every ratio.

\begin{figure*}[!tbp]
    \centering
    \includegraphics[width=0.7\textwidth]{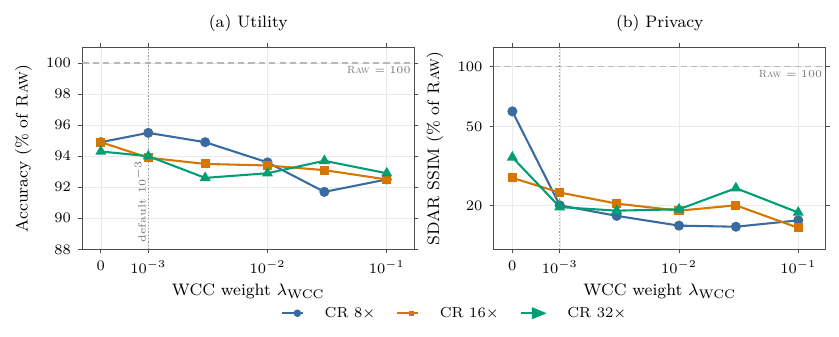}
    \caption{WCC weight per compression ratio (CIFAR-10, L2H-SCH head), both axes relative to \textsc{Raw} (dashed, $100\%$): (a)~converged accuracy; (b)~median SDAR SSIM, lower $=$ more private; the dotted line marks the default $\lambda_{\mathrm{WCC}}{=}10^{-3}$.}
    \label{fig:wcc_op}
\end{figure*}

The coefficient saturates early (Fig.~\ref{fig:wcc_op}). Reconstruction is at its floor already at $10^{-3}$, and converged accuracy barely moves: $82.8\%$ at $\lambda_{\mathrm{WCC}}{=}0$, $83.3\%$ at $10^{-3}$, and $81.6$/$80.7\%$ at $10^{-2}$/$10^{-1}$. We therefore recommend $\lambda_{\mathrm{WCC}}\!\approx\!10^{-3}$. Table~\ref{tab:privacy_all}
reports this operating point against every attack, and the per-attack grids (Tables~\ref{tab:final-sdar-usl-all-split-points}, \ref{tab:fora-full} and~\ref{tab:unsplit_attack_full}) bracket it with $\lambda_{\mathrm{WCC}}\in\{0,10^{-2},10^{-1}\}$.
The $\lambda_{\mathrm{WCC}}$ recommendation above holds for the L2H-SCH configuration, which stays within $8\%$ of \textsc{Raw} at every weight (Table~\ref{tab:wcc_ablation}). With per-client heads, WCC costs accuracy at every weight, and more so for the deeper head: \ournameFixed L2H-PCH drops $16.4\%$ below \textsc{Raw} at $10^{-3}$ and $46.9\%$ at $10^{-1}$, against $7.9\%$ and $22.5\%$ for L1H-PCH. We attribute this to the head configuration. WCC minimizes each client's within-class variance around the class means of its own batches. Per-client heads share no parameters, so each head contracts its classes toward a different set of means, and the shared backbone is trained on as many mutually inconsistent embeddings as there are clients; the deeper the per-client head, the further the embeddings can drift apart.

For per-client heads we therefore use a different operating point: no WCC, the block-local projection for utility, and DP noise for privacy. Neither knob couples the heads. On the utility side, at $8\times$ on CIFAR-10, L2H-PCH stays within $2.7$ points of \textsc{Raw} with $b{=}2$ tiles, against $5.7$ points for the global projection (Table~\ref{tab:block_util_F}), and
the SDAR attack moves from $29\%$ to $36\%$ of \textsc{Raw} (Table~\ref{tab:block_privacy_grid}). On the privacy side, DP cuts the SDAR attack from $0.360$ to $0.139$ SSIM at the L7 cut (Table~\ref{tab:dp_privacy}) at almost no cost to a per-client head, $+0.7$ points on average over the \ournameFixed PCH cells of Table~\ref{tab:dp_utility}.

\paragraph{Center loss vs.\ WCC}
At matched strength the two are indistinguishable (Fig.~\ref{fig:center_wcc}): median SDAR SSIM over the healthy seeds is $0.09$ for center loss and $0.10$ for WCC at $\lambda{=}0.1$, and $0.11$ versus $0.09$ at $\lambda{=}0.01$, both far below the projection-only $0.34$ and inside each other's seed spread. The stateful centers therefore add no measurable reconstruction-privacy gain over the stateless per-batch penalty, which needs no persistent buffer, no center learning rate, and no cross-round center synchronization, so we keep WCC as the default. Cells are median SDAR SSIM over multiple seeds.

\paragraph{Lift-back hidden-dim ablation}
The MLP lift-back of \textsc{\ournameLearned} introduces one server-side hyperparameter not exposed in the main eval, the hidden width $m$ in Eq.~\eqref{eq:frll_liftback}. Table~\ref{tab:mlp-hidden-ablation} sweeps $m$ across the four head configurations to motivate the per-dataset values used in the main runs. Returns are dataset-specific and saturate quickly. The L2H + SCH cell reaches its best at $h{=}512$ on CIFAR-10 ($85.91$, $-1.33$ vs.\ \textsc{Raw}), $h{=}2048$ on CIFAR-100 ($55.34$), $h{=}128$ on Fashion-MNIST ($91.22$), and $h{=}128$ on GTSRB ($95.88$); larger $h$ stops helping or actively hurts on the easier datasets. The per-dataset hidden dims used in Table~\ref{tab:full_sweep_baseline_appendix} (cifar10$=$512, cifar100$=$2048, fmnist$=$128, gtsrb$=$128) are read off this ablation, so the main \textsc{\ournameLearned} accuracy is the per-dataset best of this sweep rather than a single fixed $h$.

\paragraph{MLP lift-back size}
Fig.~\ref{fig:mlp_ablation} varies $m\!\in\!\{64,\dots,2048\}$ at L2H-SCH, CR$\,{=}\,8\times$. On FashionMNIST, GTSRB, and CIFAR-10 the gap is flat from $|\phi|\!<\!1$\,M (best-to-worst spread $\leq 1.98$\,pp); on CIFAR-100, scaling $\phi$ from $1.2$M to $9.4$M tightens the gap by $3.11$\,pp ($-7.24\!\to\!-4.13$). The lift-back's required capacity scales with task difficulty, matching the motivation in \sect\ref{sec:method-liftback}.

\paragraph{Client-count scaling}
Table~\ref{tab:client_scaling_ablation} varies the client count $n\in\{10,25,50,100\}$ at the L2H-SCH, CR$\,{=}\,8\times$ default across both datasets, and reports the best test accuracy at each $n$ along with the $n{=}10\!\to\!n{=}100$ swing $\Delta_{10\to100}$.
\textsc{\ournameFixed} loses at most $0.67$\,pp from $n{=}10$ to $n{=}100$ on CIFAR-10 and $3.33$\,pp on CIFAR-100; \textsc{\ournameLearned} is even flatter, losing $\leq 0.25$\,pp on CIFAR-10 and $2.60$\,pp on CIFAR-100.
On CIFAR-100 the gap to \textsc{Raw} actually reduces with $n$, as the performance also for \textsc{Raw} drops $5.08$\,pp from $n{=}10$, while \textsc{\ournameLearned} drops only $2.60$\,pp over the same interval.
Thus, \ourname achieves consistent results with increasing number of clients.

\subsection{Runtime Evaluation on Representative Hardware}
\label{app:runtime}
We evaluated the runtime overhead of \ourname on representative edge hardware using a Raspberry Pi with 4 GB RAM and our reported server. Without compression, the average processing time from the input through the client head and server backbone was $0.189$ s per batch. With $8\times$, $16\times$, and $32\times$ compression, the average runtime increased to $0.247$ s, $0.216$ s, and $0.204$ s, corresponding to overheads of approximately $30.8\%$, $14.3\%$, and $8.1\%$, respectively. Despite the additional compression and decompression operations, the $32\times$ configuration introduces only a minimal increase in runtime, with the per-sample amortized runtime being similarly low, increasing from $1.48$ ms to $1.60$ ms. The client stores $\mathbf{R}$ as $d{\times}k$ values, $4.2$/$2.1$/$1.0$\,MB in fp16 at $8/16/32\times$ for $d{=}4096$, once at setup, or regenerates it from its seed.

\section{Training Analysis}
\label{app:training}

\begin{table}[!htbp]
\centering
\caption{Per-step messages on the cut interface.}
\label{tab:wcc_wire_summary}
\small
\begin{tabular}{ll}
\toprule
\textbf{Direction} & \textbf{Payload} \\
\midrule
Forward, $\text{C}\!\to\!\text{S}$ &
    $\tilde{\mathbf{z}}_i \in \mathbb{R}^{k}$ \\
Forward, $\text{S}\!\to\!\text{C}$ &
    $\mathbf{u}_i$ (server-backbone output) \\
Backward, $\text{C}\!\to\!\text{S}$ &
    $\partial \mathcal{L}_{\mathrm{CE}}/\partial \mathbf{u}_i$ \\
Backward, $\text{S}\!\to\!\text{C}$ &
    $\partial \mathcal{L}_{\mathrm{CE}}/\partial \tilde{\mathbf{z}}_i$ \\
\bottomrule
\end{tabular}
\end{table}

\subsection{Computational complexity}
\label{app:exp:complexity}
Let $b$ denote the mini-batch size, let the cut activation have shape $C{\times}H{\times}W$ and flattened dimension $d{=}CHW$, let $k$ be the transmitted width ($\CR{=}d/k$), and let $m$ be the hidden width of the learned lift-back. We count only the bottleneck and lift-back overhead beyond the common split model $h\circ g\circ f$.

\paragraph{Cost of \ourname}
Both \textsc{\ourname} variants send exactly $bk$ scalars per batch. In \textsc{\ournameFixed}, the client computes the fixed projection $\mathbf{R}^{\top}\mathbf{z}$ and the server applies the fixed lift-back $\mathbf{R}\tilde{\mathbf{z}}$, so the extra work is $O(bdk)$ on each side and the bottleneck adds no trainable parameters. \textsc{\ournameLearned} keeps the same client computation and the same $bk$ communication, but replaces the fixed server lift-back with the MLP $k\!\to\!m\!\to\!d$. Its extra adaptation cost is therefore server-side: $O(m(k+d))$ trainable parameters and $O(bm(k+d))$ work. WCC, when enabled, is only a training-time client regularizer on $\tilde{\mathbf{z}}$; it costs $O(bk)$ arithmetic, sends no extra values, and disappears at inference.

\subsection{Training steps and WCC backprop}
\label{app:wcc_backprop}

This appendix walks through one mini-batch of \ourname{} training end to end, using the two-optimizer deployment that mirrors how U-shaped split learning is actually run in practice: the client process holds its own optimizer over the head $f$ and the tail $h$, and the server process holds its own optimizer over the backbone $g$ and, in \textsc{\ournameLearned}, the learned lift-back $\phi$. The projection matrix $\mathbf{R}$ is fixed at provisioning time and is never touched by either optimizer. We provide this walkthrough as a self-contained reference for the forward and backward computations, the messages exchanged across the trust boundary, and the parameter updates each side applies. A useful by-product of laying everything out in this form is that it makes immediate the claim, stated in the methodology, that adding $\mathcal{L}_{\mathrm{WCC}}$ leaves the client-server protocol of plain U-shaped split learning unchanged: the forward and backward payloads exchanged at the cut, and the server's parameter update, are bit-identical with or without $\mathcal{L}_{\mathrm{WCC}}$.

\subsubsection{Forward pass}

\paragraph{Step 1 (client, head)}
On a mini-batch $\{(\mathbf{x}_i, y_i)\}_{i=1}^{b}$, the client computes the cut-layer activation $\mathbf{z}_i = f(\mathbf{x}_i)\in\mathbb{R}^{d}$.

\paragraph{Step 2 (client, projection)}
The client applies the fixed projection $\tilde{\mathbf{z}}_i = \mathbf{R}^{\top}\mathbf{z}_i \in \mathbb{R}^{k}$, following Eq.~\eqref{eq:projection}. The resulting tensor is recorded in the client's autograd graph as a function of $f$'s parameters.

\paragraph{Step 3 (network, $\text{client}\to\text{server}$).}
The client transmits $\{\tilde{\mathbf{z}}_i\}_{i=1}^{b}$ to the server. This is the only forward payload that crosses the trust boundary; the labels $\{y_i\}$ remain on the client.

\paragraph{Step 4 (server, lift-back)}
The server computes $\hat{\mathbf{z}}_i = \phi(\tilde{\mathbf{z}}_i)$, where $\phi$ is either the fixed map of \textsc{\ournameFixed} or the learned MLP of \textsc{\ournameLearned}. The result is recorded in the \emph{server's} autograd graph, which is a graph distinct from the client's.

\paragraph{Step 5 (server, backbone)}
The server computes $\mathbf{u}_i = g(\hat{\mathbf{z}}_i)$ and retains the server-side autograd graph rooted at $\tilde{\mathbf{z}}_i$.

\paragraph{Step 6 (network, $\text{server}\to\text{client}$)}
The server transmits $\{\mathbf{u}_i\}_{i=1}^{b}$ to the client. On reception, the client treats $\mathbf{u}_i$ as a fresh leaf with \texttt{requires\_grad=True}; the client's autograd graph resumes from $\mathbf{u}_i$ onward.

\paragraph{Step 7 (client, tail).}
The client computes $\hat{y}_i = h(\mathbf{u}_i)$.

\paragraph{Step 8 (client, cross-entropy loss)}
The client evaluates the per-batch cross-entropy $\mathcal{L}_{\mathrm{CE}} = b^{-1}\sum_i \ell_{\mathrm{CE}}(\hat{y}_i,y_i)$. By construction, $\mathcal{L}_{\mathrm{CE}}$ depends on $h$, $\mathbf{u}$, and $y$ in the client's graph, but not on $\tilde{\mathbf{z}}$ directly.

\paragraph{Step 9 (client, within-class compaction)}
Reusing the $\tilde{\mathbf{z}}_i$ tensors from Step~2, which remain alive in the client's autograd graph, the client computes $\mathcal{L}_{\mathrm{WCC}}$ from Eq.~\eqref{eq:wcc}. The labels $y$ are used only to partition the batch into per-class index sets $S_c$, in keeping with the U-shaped topology in which labels never leave the client.

\paragraph{Step 10 (client, total loss)}
The client forms the scalar objective $\mathcal{L} = \mathcal{L}_{\mathrm{CE}} + \lambda_{\mathrm{WCC}}\,\mathcal{L}_{\mathrm{WCC}}$ as in Eq.~\eqref{eq:total_loss}.

\subsubsection{Backward pass: client (Phase~1)}

\paragraph{Step 11}
The client zeros its optimizer state with \texttt{client\_optimizer.zero\_grad()}.

\paragraph{Step 12.}
The client invokes \texttt{loss.backward()} on its local graph. Autograd traverses the two branches simultaneously:
\begin{itemize}
    \item The CE branch reaches $h$ and accumulates $\partial \mathcal{L}_{\mathrm{CE}} / \partial \theta_h$ into $h$'s \texttt{.grad}, then propagates back to $\mathbf{u}$, leaving $\partial \mathcal{L}_{\mathrm{CE}} / \partial \mathbf{u}$ on $\mathbf{u}$'s \texttt{.grad}. Because $\mathbf{u}$ is a leaf in the client's graph, traversal stops there.
    \item The WCC branch propagates from $\mathcal{L}_{\mathrm{WCC}}$ through $\tilde{\mathbf{z}}$ and the fixed factor $\mathbf{R}^{\top}$ to $f$, accumulating $\lambda_{\mathrm{WCC}}\,\partial \mathcal{L}_{\mathrm{WCC}} / \partial \theta_f$ into $f$'s \texttt{.grad}, and leaving $\lambda_{\mathrm{WCC}}\,\partial \mathcal{L}_{\mathrm{WCC}} / \partial \tilde{\mathbf{z}}$ on $\tilde{\mathbf{z}}$'s \texttt{.grad}.
\end{itemize}
At this point $f$'s \texttt{.grad} contains only the contribution from the client-local term; the contribution from $\mathcal{L}_{\mathrm{CE}}$ requires the server's backward signal, which has not yet arrived.

\paragraph{Step 13 (network, $\text{client}\to\text{server}$).}
The client transmits $\partial \mathcal{L}_{\mathrm{CE}} / \partial \mathbf{u}$ to the server. This is the standard backward payload of U-shaped split learning.

\subsubsection{Backward pass: server}

\paragraph{Step 14}
The server zeros its optimizer state with \texttt{server\_optimizer.zero\_grad()}.

\paragraph{Step 15}
The server uses the incoming $\partial \mathcal{L}_{\mathrm{CE}} / \partial \mathbf{u}$ as the upstream seed and calls \texttt{u.backward(grad\_output)} on its server-side graph. Autograd traverses $\mathbf{u}\to g \to \hat{\mathbf{z}} \to \phi \to \tilde{\mathbf{z}}$, accumulating $\partial \mathcal{L}_{\mathrm{CE}} / \partial \theta_g$ and (in \textsc{\ournameLearned}) $\partial \mathcal{L}_{\mathrm{CE}} / \partial \theta_\phi$ into the corresponding \texttt{.grad} buffers, and leaving $\partial \mathcal{L}_{\mathrm{CE}} / \partial \tilde{\mathbf{z}}$ on the server-side $\tilde{\mathbf{z}}$.

\paragraph{Step 16}
The server applies \texttt{server\_optimizer.step()}. The update of $\theta_g$ and $\theta_\phi$ is determined entirely by $\mathcal{L}_{\mathrm{CE}}$.

\paragraph{Step 17 (network, $\text{server}\to\text{client}$)}
The server transmits $\partial \mathcal{L}_{\mathrm{CE}} / \partial \tilde{\mathbf{z}}$ back to the client.

\subsubsection{Backward pass: client (Phase~2)}

\paragraph{Step 18.}
The client receives $\partial \mathcal{L}_{\mathrm{CE}} / \partial \tilde{\mathbf{z}}$ and \emph{adds} it to the contribution already sitting on $\tilde{\mathbf{z}}$'s \texttt{.grad} from Step~12, yielding
\begin{equation}
    \frac{\partial \mathcal{L}}{\partial \tilde{\mathbf{z}}}
    \;=\;
    \frac{\partial \mathcal{L}_{\mathrm{CE}}}{\partial \tilde{\mathbf{z}}}
    \;+\;
    \lambda_{\mathrm{WCC}}\,
    \frac{\partial \mathcal{L}_{\mathrm{WCC}}}{\partial \tilde{\mathbf{z}}}.
    \label{eq:zgrad_fuse}
\end{equation}
This client-local addition is the only point in the entire training step at which the two gradient pathways meet.

\paragraph{Step 19.}
The client backpropagates Eq.~\eqref{eq:zgrad_fuse} through the fixed projection (chain-rule by $\mathbf{R}$, which has no parameters to update) and through $f$. The CE contribution to $f$'s gradient is now accumulated on top of the WCC contribution from Step~12, giving
\begin{equation}
    \frac{\partial \mathcal{L}}{\partial \theta_f}
    \;=\;
    \frac{\partial \mathcal{L}_{\mathrm{CE}}}{\partial \theta_f}
    \;+\;
    \lambda_{\mathrm{WCC}}\,
    \frac{\partial \mathcal{L}_{\mathrm{WCC}}}{\partial \theta_f}.
    \label{eq:fgrad_fuse}
\end{equation}

\paragraph{Step 20.}
The client applies \texttt{client\_optimizer.step()}, updating $h$ from $\partial \mathcal{L}_{\mathrm{CE}} / \partial \theta_h$ and $f$ from Eq.~\eqref{eq:fgrad_fuse}. The current mini-batch step is complete.

\subsubsection{Wire-level and optimizer-level summary}

The four messages exchanged at the cut interface during a single mini-batch step are summarized in Table~\ref{tab:wcc_wire_summary}: the forward payload $\tilde{\mathbf{z}}$, the server's return signal $\mathbf{u}$, and the two backward gradient signals $\partial \mathcal{L}_{\mathrm{CE}}/\partial \mathbf{u}$ and $\partial \mathcal{L}_{\mathrm{CE}}/\partial \tilde{\mathbf{z}}$. Each direction transmits exactly one tensor per sample. The server's update of $(\theta_g, \theta_\phi)$ depends on $\mathcal{L}_{\mathrm{CE}}$ alone, because $\mathcal{L}_{\mathrm{WCC}}$ has no functional dependency on any server-side variable. The client's update of $\theta_h$ depends on $\mathcal{L}_{\mathrm{CE}}$ alone for the same reason, and the client's update of $\theta_f$ is the sum of contributions from $\mathcal{L}_{\mathrm{CE}}$ and $\mathcal{L}_{\mathrm{WCC}}$ (the latter being a client-local quantity). $\mathbf{R}$ is fixed and never updated.

A direct consequence is that the protocol described in Steps~1 to 20 reduces to the standard U-shaped split-learning protocol when $\lambda_{\mathrm{WCC}}=0$, with neither the cut-layer messages nor the server-side update changing form. Setting $\lambda_{\mathrm{WCC}}>0$ introduces the additive client-local term in Eq.~\eqref{eq:fgrad_fuse} and nothing else.

\subsubsection{Equivalence between single- and two-optimizer implementations}

For any first-order optimizer (SGD, Adam, AdamW, etc.) applied parameter-wise, splitting a single \texttt{loss.backward()} / \texttt{optimizer.step()} call into two independent \texttt{(zero\_grad, backward, step)} sequences, one on each side of the cut, produces identical per-parameter updates. The reason is that parameter-wise optimizers depend only on each parameter's accumulated gradient, which is determined by the autograd graph rather than by the process boundary. Our reference implementations in Alg.~\ref{alg:lightsplit-client} and Alg.~\ref{alg:lightsplit-server} express the procedure with one optimizer per side; the two-optimizer deployment described above produces mathematically identical $f$, $h$, $g$, and $\phi$ updates.

\subsection{Compact derivation in equations}
\begingroup\setlength{\abovedisplayskip}{2pt}\setlength{\belowdisplayskip}{2pt}\setlength{\abovedisplayshortskip}{1pt}\setlength{\belowdisplayshortskip}{1pt}\setlength{\jot}{1pt}%
\label{app:wcc_backprop_eqs}

For a reader who prefers the gradient bookkeeping in equation form, this subsection restates the per-step computation as a sequence of derivatives. Let $\theta_f, \theta_h, \theta_g, \theta_\phi$ denote the trainable parameters of $f, h, g, \phi$, respectively, and let $J_F$ denote the Jacobian of a map $F$ at the current point.

\paragraph{Forward composition}
Per sample $i$ in a batch of size $b$:
\begin{align}
    \mathbf{z}_i        &= f(\mathbf{x}_i;\theta_f),
    \label{eq:fwd_f}\\
    \tilde{\mathbf{z}}_i &= \mathbf{R}^{\top}\mathbf{z}_i,
    \label{eq:fwd_proj}\\
    \hat{\mathbf{z}}_i  &= \phi(\tilde{\mathbf{z}}_i;\theta_\phi),
    \label{eq:fwd_phi}\\
    \mathbf{u}_i        &= g(\hat{\mathbf{z}}_i;\theta_g),
    \label{eq:fwd_g}\\
    \hat{y}_i           &= h(\mathbf{u}_i;\theta_h).
    \label{eq:fwd_h}
\end{align}

\paragraph{Losses}
\begin{align}
    \mathcal{L}_{\mathrm{CE}}
        &= \frac{1}{b}\sum_{i=1}^{b}\ell_{\mathrm{CE}}(\hat{y}_i, y_i),
    \label{eq:ce_def_app}\\
    \mathcal{L}_{\mathrm{WCC}}
        &= \sum_{c\in\mathcal{Y}_b}\frac{1}{|S_c|}\sum_{i\in S_c}
           \big\|\tilde{\mathbf{z}}_i-\boldsymbol{\mu}_c\big\|_2^{2},
    \label{eq:wcc_def_app}\\
    \mathcal{L}
        &= \mathcal{L}_{\mathrm{CE}}
         + \lambda_{\mathrm{WCC}}\,\mathcal{L}_{\mathrm{WCC}}.
    \label{eq:total_def_app}
\end{align}

\paragraph{Functional dependencies}
$\mathcal{L}_{\mathrm{CE}}$ depends on $\tilde{\mathbf{z}}_i$ only through the round-trip $\tilde{\mathbf{z}}_i\!\to\!\hat{\mathbf{z}}_i\!\to\!\mathbf{u}_i\!\to\!\hat{y}_i$. $\mathcal{L}_{\mathrm{WCC}}$ depends on $\{\tilde{\mathbf{z}}_i, y_i\}$ alone, so its partials with respect to all server-side variables vanish:
\begin{multline}
    \frac{\partial \mathcal{L}_{\mathrm{WCC}}}{\partial \mathbf{u}_i}
    \;=\;
    \frac{\partial \mathcal{L}_{\mathrm{WCC}}}{\partial \hat{\mathbf{z}}_i}
    \;=\;
    \frac{\partial \mathcal{L}_{\mathrm{WCC}}}{\partial \theta_h}
    \\[2pt]
    \;=\;
    \frac{\partial \mathcal{L}_{\mathrm{WCC}}}{\partial \theta_g}
    \;=\;
    \frac{\partial \mathcal{L}_{\mathrm{WCC}}}{\partial \theta_\phi}
    \;=\; \mathbf{0}.
    \label{eq:wcc_zero_partials}
\end{multline}

\paragraph{Backward, client phase~1}
Starting from $\mathcal{L}$ and walking back through the client's autograd graph until the cut interface:
\begin{align}
    \frac{\partial \mathcal{L}}{\partial \hat{y}_i}
        &= \frac{\partial \mathcal{L}_{\mathrm{CE}}}{\partial \hat{y}_i},
    \label{eq:bwd_yhat}\\
    \frac{\partial \mathcal{L}}{\partial \theta_h}
        &= \sum_i \frac{\partial \mathcal{L}_{\mathrm{CE}}}{\partial \hat{y}_i}\,
           J_h(\mathbf{u}_i;\theta_h)_{\theta_h},
    \label{eq:bwd_thetah}\\
    \frac{\partial \mathcal{L}}{\partial \mathbf{u}_i}
        &= \frac{\partial \mathcal{L}_{\mathrm{CE}}}{\partial \hat{y}_i}\,
           J_h(\mathbf{u}_i;\theta_h)_{\mathbf{u}_i}.
    \label{eq:bwd_u}
\end{align}
The CE gradient at $\mathbf{u}_i$ in Eq.~\eqref{eq:bwd_u} is the only quantity transmitted from client to server in the backward direction. By Eq.~\eqref{eq:wcc_zero_partials}, this quantity is independent of $\mathcal{L}_{\mathrm{WCC}}$.

In parallel, autograd accumulates the WCC contribution along the head-side chain:
\begin{align}
    \frac{\partial \mathcal{L}_{\mathrm{WCC}}}{\partial \tilde{\mathbf{z}}_i}
    &\quad\text{(closed form from Eq.~\eqref{eq:wcc_def_app})},
    \label{eq:bwd_wcc_zt}\\
    \frac{\partial \mathcal{L}_{\mathrm{WCC}}}{\partial \mathbf{z}_i}
        &= \mathbf{R}\,
           \frac{\partial \mathcal{L}_{\mathrm{WCC}}}{\partial \tilde{\mathbf{z}}_i},
    \label{eq:bwd_wcc_z}\\
    \frac{\partial \mathcal{L}_{\mathrm{WCC}}}{\partial \theta_f}
        &= \sum_i \frac{\partial \mathcal{L}_{\mathrm{WCC}}}{\partial \mathbf{z}_i}\,
           J_f(\mathbf{x}_i;\theta_f)_{\theta_f}.
    \label{eq:bwd_wcc_thetaf}
\end{align}

\paragraph{Backward, server}
The server seeds its backward with the incoming $\partial \mathcal{L}_{\mathrm{CE}}/\partial \mathbf{u}_i$ from Eq.~\eqref{eq:bwd_u} and produces:
\begin{align}
    \frac{\partial \mathcal{L}_{\mathrm{CE}}}{\partial \theta_g}
        &= \sum_i \frac{\partial \mathcal{L}_{\mathrm{CE}}}{\partial \mathbf{u}_i}\,
           J_g(\hat{\mathbf{z}}_i;\theta_g)_{\theta_g},
    \label{eq:bwd_thetag}\\
    \frac{\partial \mathcal{L}_{\mathrm{CE}}}{\partial \hat{\mathbf{z}}_i}
        &= \frac{\partial \mathcal{L}_{\mathrm{CE}}}{\partial \mathbf{u}_i}\,
           J_g(\hat{\mathbf{z}}_i;\theta_g)_{\hat{\mathbf{z}}_i},
    \label{eq:bwd_zhat}\\
    \frac{\partial \mathcal{L}_{\mathrm{CE}}}{\partial \theta_\phi}
        &= \sum_i \frac{\partial \mathcal{L}_{\mathrm{CE}}}{\partial \hat{\mathbf{z}}_i}\,
           J_\phi(\tilde{\mathbf{z}}_i;\theta_\phi)_{\theta_\phi},
    \label{eq:bwd_thetaphi}\\
    \frac{\partial \mathcal{L}_{\mathrm{CE}}}{\partial \tilde{\mathbf{z}}_i}
        &= \frac{\partial \mathcal{L}_{\mathrm{CE}}}{\partial \hat{\mathbf{z}}_i}\,
           J_\phi(\tilde{\mathbf{z}}_i;\theta_\phi)_{\tilde{\mathbf{z}}_i}.
    \label{eq:bwd_zt_server}
\end{align}

By Eq.~\eqref{eq:wcc_zero_partials}, Eqs.~\eqref{eq:bwd_thetag} to \eqref{eq:bwd_thetaphi} are also the gradients of $\mathcal{L}$ with respect to the server-side parameters, not just of $\mathcal{L}_{\mathrm{CE}}$. Eq.~\eqref{eq:bwd_zt_server} is the only quantity transmitted from server to client in the backward direction.

\paragraph{Backward, client phase~2 (gradient fusion)}
The client adds the server's contribution from Eq.~\eqref{eq:bwd_zt_server} to the WCC contribution from Eq.~\eqref{eq:bwd_wcc_zt}:
\begin{equation}
    \frac{\partial \mathcal{L}}{\partial \tilde{\mathbf{z}}_i}
        \;=\;
        \frac{\partial \mathcal{L}_{\mathrm{CE}}}{\partial \tilde{\mathbf{z}}_i}
        \;+\;
        \lambda_{\mathrm{WCC}}\,
        \frac{\partial \mathcal{L}_{\mathrm{WCC}}}{\partial \tilde{\mathbf{z}}_i}.
    \label{eq:fuse_zt}
\end{equation}
This is the unique addition step at which CE and WCC information meet, and it occurs entirely on the client. Continuing through the projection and the head:
\begin{align}
    \frac{\partial \mathcal{L}}{\partial \mathbf{z}_i}
        &= \mathbf{R}\,\frac{\partial \mathcal{L}}{\partial \tilde{\mathbf{z}}_i},
    \label{eq:fuse_z}\\
    \frac{\partial \mathcal{L}}{\partial \theta_f}
        &= \sum_i \frac{\partial \mathcal{L}}{\partial \mathbf{z}_i}\,
           J_f(\mathbf{x}_i;\theta_f)_{\theta_f}
        \;=\;
        \frac{\partial \mathcal{L}_{\mathrm{CE}}}{\partial \theta_f}
        +
        \lambda_{\mathrm{WCC}}\,
        \frac{\partial \mathcal{L}_{\mathrm{WCC}}}{\partial \theta_f}.
    \label{eq:fuse_thetaf}
\end{align}

\paragraph{Optimizer updates}
Writing $\eta$ for the learning rate and using SGD for brevity (any parameter-wise first-order rule applies analogously, with its own state):
\begin{align}
    \text{(server)}\quad
    \theta_g     &\leftarrow \theta_g
        - \eta\,\frac{\partial \mathcal{L}_{\mathrm{CE}}}{\partial \theta_g},
    \label{eq:upd_thetag}\\
    \theta_\phi  &\leftarrow \theta_\phi
        - \eta\,\frac{\partial \mathcal{L}_{\mathrm{CE}}}{\partial \theta_\phi}
        \quad(\text{\textsc{\ournameLearned} only}),
    \label{eq:upd_thetaphi}\\[4pt]
    \text{(client)}\quad
    \theta_h     &\leftarrow \theta_h
        - \eta\,\frac{\partial \mathcal{L}_{\mathrm{CE}}}{\partial \theta_h},
    \label{eq:upd_thetah}\\
    \theta_f     &\leftarrow \theta_f
        - \eta\,\Big(
            \frac{\partial \mathcal{L}_{\mathrm{CE}}}{\partial \theta_f}
            +
            \lambda_{\mathrm{WCC}}\,
            \frac{\partial \mathcal{L}_{\mathrm{WCC}}}{\partial \theta_f}
        \Big).
    \label{eq:upd_thetaf}
\end{align}
The matrix $\mathbf{R}$ is treated as a fixed constant and receives no update.

\paragraph{Reduction to plain U-shaped split learning}
Setting $\lambda_{\mathrm{WCC}}{=}0$ in Eq.~\eqref{eq:upd_thetaf} recovers plain U-shaped split learning exactly: the cut-interface payloads (Eqs.~\eqref{eq:fwd_proj}, \eqref{eq:fwd_g}, \eqref{eq:bwd_u} and~\eqref{eq:bwd_zt_server}) and the update rules (Eqs.~\eqref{eq:upd_thetag}--\eqref{eq:upd_thetah}) depend on $\mathcal{L}_{\mathrm{CE}}$ and the model alone, and the WCC term in Eq.~\eqref{eq:upd_thetaf} is computed and applied on the client.

\endgroup

\end{document}